\documentclass[11pt]{article}

\usepackage[preprint]{acl}

\usepackage{times}
\usepackage{latexsym}
\usepackage{amsmath}
\usepackage{amssymb}
\usepackage{multirow}
\usepackage{booktabs}
\usepackage{enumitem}
\usepackage{xcolor}
\usepackage{arydshln}

\usepackage{tcolorbox}
\tcbuselibrary{listings, breakable, skins}

\lstdefinelanguage{Prompt}{
  moredelim=**[is][\color{blue}]{@}{@},
}

\usepackage[T1]{fontenc}

\usepackage[utf8]{inputenc}

\usepackage{microtype}

\usepackage{inconsolata}

\usepackage{graphicx}
\usepackage{listings}

\definecolor{aihl1}{HTML}{FBE4E2}  
\definecolor{aihl2}{HTML}{F9D2CF}
\definecolor{aihl3}{HTML}{F1A8A2}
\definecolor{huhl1}{HTML}{E2ECFA}  
\definecolor{huhl2}{HTML}{CFE0F7}
\definecolor{huhl3}{HTML}{A9C7EE}
\newcommand{\hlai}[2][2]{\colorbox{aihl#1}{\strut #2}}   
\newcommand{\hlhu}[2][2]{\colorbox{huhl#1}{\strut #2}}
\newcommand{\fk}[1]{\textsuperscript{\tiny #1}}   
\title{Building Auditable AI Text Detectors with Sparse Autoencoders}

\newcommand{\capIcon}[1]{\includegraphics[height=\fontcharht\font`X]{#1}}
\newcommand{\pennIcon}{\capIcon{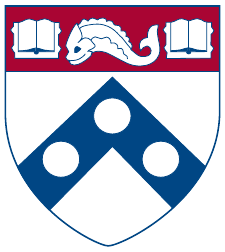}}
\newcommand{\gdmIcon}{\capIcon{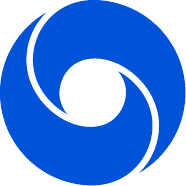}}

\author{\normalfont
  \textbf{Liam Dugan\textsuperscript{1}}\thanks{Work done as a Student Researcher at Google DeepMind. Code and trained models available at \url{https://github.com/google-deepmind/detectorscope}} \quad
  \textbf{Callum McDougall\textsuperscript{2}} \quad
  \textbf{Matthew Denton\textsuperscript{2}} \quad
  \textbf{Philip Pham\textsuperscript{2}} \\[0.25em]
  \textbf{Christine Kaeser-Chen\textsuperscript{2}} \quad
  \textbf{Chris Callison-Burch\textsuperscript{1}} \\[0.3em]
  \textsuperscript{1}\,\pennIcon\ University of Pennsylvania \qquad
  \textsuperscript{2}\,\gdmIcon\ Google DeepMind \\[0.2em]
  {\normalsize\texttt{\{ldugan,ccb\}@engineering.upenn.edu}} \\[0.1em]
  {\normalsize\texttt{\{mcdougallc,myuzaki,phillypham,christinech\}@google.com}}
}

\begin{document}
\maketitle
\begin{abstract}
AI-generated text detectors are increasingly deployed in consequential settings—peer review, education, admissions, journalism, and content moderation—where their outputs directly affect individuals. Yet the strongest detectors are fine-tuned neural classifiers whose decisions are opaque, while transparent alternatives (stylometric and linguistic-feature classifiers) underperform them by wide margins. We present a pipeline that converts a fine-tuned neural detector into a classifier that operates on human-readable concepts. We accomplish this by training sparse autoencoders (SAEs) directly on a detector's residual stream, assigning auto-interpretation labels to each SAE feature, and retaining only features whose generated descriptions pass a global fidelity threshold. We then train a lightweight classifier on just the interpretable features. We find that our SAEs recover latent concepts that correspond to highly specific and predictive linguistic ``tells'' of AI-generated text. Classifiers built on these extracted features are not only highly interpretable but exhibit stronger performance on out-of-distribution texts than either neural or stylometric alternatives.
\end{abstract}

\section{Introduction}

AI-generated text detection is no longer a purely academic problem. Detectors are used to flag student essays, vet peer-review submissions, screen job applications, and moderate online content. Their outputs feed into decisions that affect individuals, and in practice they are often treated as black boxes by users who cannot inspect or contest them.

\begin{figure}
    \centering
    \includegraphics[width={\columnwidth}]{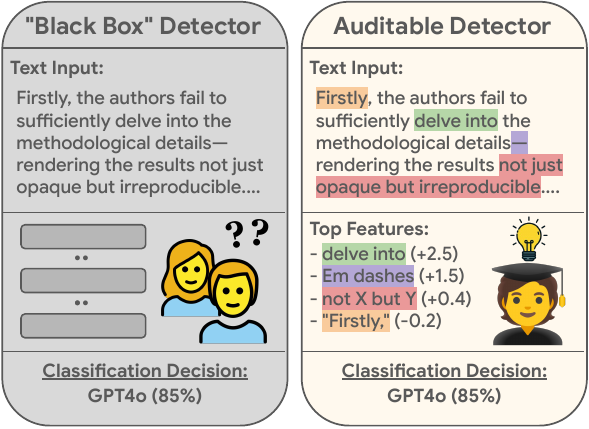}
    \caption{In our work, we use Sparse Autoencoders to convert existing black box AI detectors into ones that operate on human-interpretable concepts---allowing for more auditable deployment.}
    \label{fig:pg1}
\end{figure}

In settings like these, accuracy is not the only thing that matters. What also matters is whether an auditor---a reviewer, an instructor, a compliance officer---can look inside a decision, understand what the detector is keying on, and stress-test it. Stylometric detectors \citep{Frohling2021,aityan2026lightweightapproachdetectionaigenerated} are interpretable by construction but underperform their neural counterparts. Fine-tuned neural models achieve high accuracy but operate as black boxes.
What real-world deployment needs is a performant AI detector whose decisions can be explained and grounded in specific characteristics and features of the input text.

Our work builds an interpretable AI detector in four stages. First, we fine-tune a model on the AI detection task. Then, we train sparse autoencoders on the model's residual stream at various depths. Then, we run an auto-interpretation pipeline on the SAE features \citep{bills2023language,paulo2024automatically}, and keep only those features whose explanations reach a certain global simulation score threshold. Finally, we train a lightweight $L_1$-regularized logistic regression classifier on the remaining features. The resulting classifier is built on automatically discovered features that each have an associated natural-language description that accurately describes their underlying activation patterns.

\section{Background}

\subsection{Sparse Autoencoders} 
Individual neurons in a neural network are often polysemantic \cite{olah2020zoom}---firing on seemingly unrelated inputs in ways that make interpretability challenging.
Sparse Autoencoders (SAEs) \cite{cunningham2023sparseautoencodershighlyinterpretable, bricken2023monosemanticity} undo this polysemanticity by converting dense neural network activations $\mathbf{x} \in \mathbb{R}^{d_{\text{model}}}$ into large sparsely-active vectors of monosemantic features $\mathbf{h} \in \mathbb{R}^{d_{\text{SAE}}}$. The encoder and decoder are parameterized as
\begin{align}
\mathbf{f}_{\text{enc}}(\mathbf{x}) &= \sigma(\mathbf{W}_{\text{enc}}\mathbf{x} + \mathbf{b}_{\text{enc}}) \label{eq:enc}\\
\mathbf{f}_{\text{dec}}(\mathbf{h}) &= \mathbf{W}_{\text{dec}}\mathbf{h} + \mathbf{b}_{\text{dec}}
\end{align}
where $\sigma$ is a non-linear function (often ReLU). These functions are trained to reconstruct the input $\mathbf{x}$ while minimizing a sparsity penalty on $\mathbf{h}$.
\begin{equation}
\mathcal{L} = ||\mathbf{x} - \mathbf{\hat{x}}||_2^2 + \lambda ||\mathbf{h}||_1
\end{equation}
where $\lambda$ is a coefficient set to balance the tradeoff between sparsity and reconstruction accuracy. The resulting features are highly monosemantic and are usable in place of neurons for interpretability analysis \cite{shu-etal-2025-survey,marks2025sparse}.

\subsection{Automatically Interpreting SAE Features}
\label{sec:autointerp}
Auto-interpretation \cite{bills2023language,paulo2024automatically} is the act of generating human-readable descriptions of SAE features given their activations on a large corpus. A typical pipeline for this is as follows: For each feature, sample a small representative set of documents from across the feature's activation distribution. Then, provide these documents to a large language model and ask it to generate a short human-readable description of that feature (typically 10-20 tokens).

Once a description is generated, the model is then asked to validate the accuracy of the description by ``simulating'' the neuron---predicting on which tokens of an arbitrary input text the feature is active---given only the generated description. The resulting correlation score between the true activations and simulated activations measures how faithfully the feature's description represents its true behavior.

\subsection{SAEs for Hypothesis Generation} 
Recent work has begun to investigate SAEs not just as ways of uncovering information about the internal activations of language models but also as ways of generating hypotheses about textual classification tasks in an unsupervised manner \cite{movva_sparse_2025}.

For example, say an SAE feature fires disproportionately on a particular class in a classification task. That feature may represent a semantically meaningful and predictive aspect of the class. In their work, \citet{movva_sparse_2025} found SAE features such as \textit{``mentions being surprised or shocked''} that were positively predictive of news engagement and \textit{``mentions issues related to food safety''} that were negatively predictive of restaurant rating.

Broadly speaking, this concept of using interpretability as a ``microscope'' \cite{Olah2015Visualizing} to illuminate interesting aspects of underlying data distributions is a strong undercurrent of our work. We follow this line of work by uncovering specific predictive linguistic cues of AI generated text and show that these tells create strong classifiers.

\begin{figure*}[t]
\centering
\includegraphics[width=\textwidth]{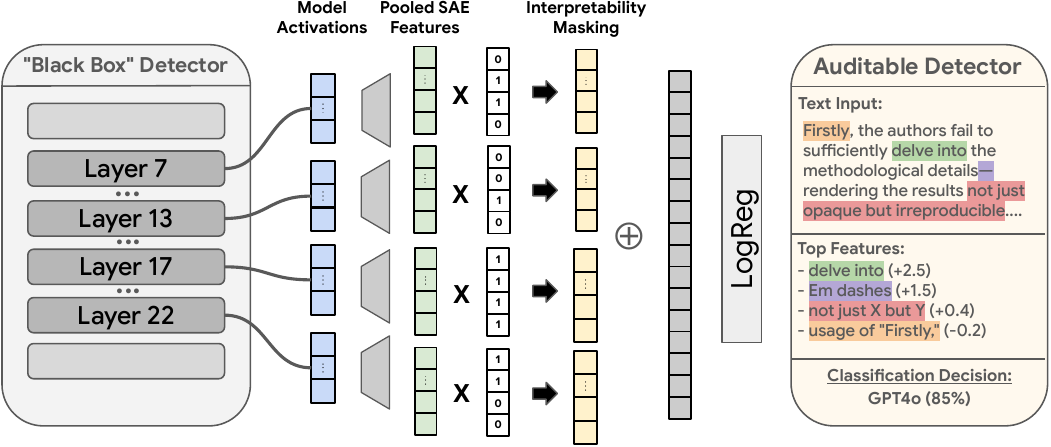}
\caption{Our pipeline. Starting from a fine-tuned AI-text detector, we train SAEs on its residual stream at four layers, label each latent with an auto-interp explanation, mask out all features that do not meet our global threshold and train a lightweight classifier over the remaining concepts. The result is a detector that retains the strength of the original neural model while being composed of auditable and intervenable human-understandable concepts.}
\label{fig:pipeline}
\end{figure*}

\section{Related Work}
\paragraph{Explainable AI Detection} Early work on explainable AI detectors primarily centered around distributional analysis of linguistic features \cite{Frohling2021, uchendu-etal-2020-authorship, zahid-etal-2024-probing, Zaitsu2025}. Follow-up work refined this to include discourse features \cite{kim-etal-2024-threads, russell2026storyscopeinvestigatingidiosyncrasiesai} dependency parses \cite{ahmed2026dependencyaidetectingaigenerated} ngram overlaps \cite{koike2026exagptexamplebasedmachinegeneratedtext} and ensembles \cite{aityan2026lightweightapproachdetectionaigenerated}. However, AI detectors built on these linguistic features generally underperform black-box counterparts and are limited in their ability to generalize to adversarial or out-of-distribution inputs \cite{schuster-etal-2020-limitations}.

On the other hand, fine-tuned neural classifiers \cite{solaiman2019releasestrategiessocialimpacts}, and other zero-shot likelihood-based metrics \cite{detectGPT,binoculars} exhibit surprisingly strong in-distribution performance and much better generalization than their stylometric counterparts \cite{dugan-etal-2025-genai,sandhan2026trustllmdetectors}. Previous work used techniques like LIME \cite{lime} and SHAP \cite{shap} to interpret these neural models \cite{Shah2023, schoenegger2026evaluationexplanationmethodsblackbox} but such explanations only give global token-level importance scores and fail to provide rationales as to \textit{why} particular tokens are predictive. Our work is the first to go beyond these token-based attribution scores and create a fully interpretable AI detection system that maintains the strong performance of neural models.

\paragraph{Interpretable Classifiers using SAE Features} 
$L_1$-regularized linear classifiers (``sparse probes'') trained on SAE features were first proposed as diagnostic tools for measuring monosemanticity \cite{gurnee2023findingneuronshaystackcase,gao2024scalingevaluatingsparseautoencoders}. Follow up work showed that such classifiers perform on par with (but no better than) dense linear probes \cite{bricken2024features, kantamneni2025sparseautoencodersusefulcase}. Our work follows \citet{peng2026position} in claiming that interpretability, not accuracy, is the primary benefit of these classifiers.

In this vein, recent work has shown that classifiers trained on SAE features are useful as interpretable reward models \cite{zhang2025interpretablerewardmodelsparse}, dense retrievers \cite{park-etal-2025-decoding}, hallucination detectors \cite{xiong2026faithfulretrievalaugmentedgenerationsparse}, and LLM-as-judge systems \cite{wedgwood2026automatedconceptdiscoveryllmasajudge}. However, these works all provide the full set of SAE features as input to the downstream classifiers without any explicit removal of high density or polysemantic SAE latents \cite{sun2025densesaelatentsfeatures,chanin2025absorptionstudyingfeaturesplitting}. The lack of such filtering undermines the faithfulness of the system-level explanations.

To combat this, \citet{kulkarni2026interpretablesteerableconceptbottleneck} proposed to remove the $M$ least interpretable SAE features from their classifier. However, this does not guarantee that the remaining features are themselves interpretable. Our work is the first to propose a global filtering threshold based on auto-interpretability score to ensure our explanations are faithful to the underlying dynamics of our classifier.

\paragraph{Using SAE Features for AI Detection}
Previous work by \citet{scher2024initial} noted the potential for SAEs as a means of interpreting AI detectors. They trained a lightweight classifier on SAE features of GPT-2 Small and found that they outperformed stylometric features in AI detection. Follow-up work by \citet{kuznetsov-etal-2025-feature} looked at how SAE features of Gemma-2-2B co-occurred with AI generated text. However, neither of these works train their own SAE models on AI detectors, nor do they evaluate the accuracy of the provided feature descriptions. Our work is the first to train SAEs directly on AI detectors and is the first to propose using these features as an interpretable AI detector.

\section{Methods}
\subsection{Pipeline Overview}
Figure \ref{fig:pipeline} shows an illustration of our full pipeline. Given a token sequence $(\mathbf{t}_0, \ldots, \mathbf{t}_n)$, we extract language model activations $(\mathbf{d}_0^{(\ell)}, \ldots, \mathbf{d}_n^{(\ell)})$ from the residual stream at layers $\ell \in L$ and pass them through the corresponding SAE encoder to get features $(\mathbf{h}_0^{(\ell)}, \ldots, \mathbf{h}_n^{(\ell)})$. We then max pool the resulting features along the sequence dimension and concatenate along the layer dimension to get the feature vector for our document $\mathbf{h} \in \mathbb{R}^{|L| \cdot d_{\text{SAE}}}$.\footnote{In our work we use $L=7,13,17,22$ following \citet{mcdougall2025gemma}, but this can be modified as needed.}

For our interpretability mask, we compute a simulation score $s_i \in [-1, 1]$ for each feature $h_i \in \mathbf{h}$ using our auto-interpretation pipeline (\S\ref{sec:autointerpretation}). This score measures how faithfully a generated natural-language description predicts that feature's activations on our training corpus. Given a global interpretability threshold $\tau$, we construct a binary mask $\mathbf{m} \in \{0,1\}^{|L| \cdot d_{\text{SAE}}}$ where $m_i = \mathbb{I}[s_i \geq \tau]$ and apply this mask element-wise to obtain our masked feature vector $\mathbf{h}_{\tau} = \mathbf{h} \odot \mathbf{m}$.

Finally, to obtain our prediction, we apply an $L_1$-regularized logistic regression to the features.\footnote{That is, given labeled data $\mathcal{D} = \{\mathbf{H}_\tau, \mathbf{y}\}$, we minimize
$\mathcal{L}(\boldsymbol{\beta}, \lambda) = \frac{1}{N}\,\mathrm{BCE}\big(\mathbf{y}, \, \sigma(\mathbf{H}_\tau \boldsymbol{\beta})\big) + \lambda \|\boldsymbol{\beta}\|_1$ where $\mathrm{BCE}$ is binary cross-entropy loss and $\sigma$ is sigmoid. The prediction for a given input is $\hat{y} = \sigma(\mathbf{h}_\tau^\top \boldsymbol{\beta}^\star)$, with $\boldsymbol{\beta}^\star = \operatorname*{arg\,min}_{\boldsymbol{\beta}} \mathcal{L}(\boldsymbol{\beta}, \lambda)$.} Since each feature $h_i \in \mathbf{h}_\tau$ has a corresponding natural-language description from our auto-interpretation pipeline, we can straightforwardly explain any given prediction by listing out the top-$k$ features with the largest net contribution to the prediction along with their descriptions.\footnote{Full training details such as preprocessing, hyperparameters, and grid searches for both our Logistic Regression model and trained SAEs are listed in Appendix \ref{app:training_details}.}

\subsection{Auto-Interpretation}
\label{sec:autointerpretation}
\paragraph{Description Generation} To generate a description for a feature $h_i$, we partition the training corpus into $k$ quantile bins $\mathcal{Q}_1, \ldots, \mathcal{Q}_k$ based on the empirical distribution of document-level max activations of the feature, plus a bin $\mathcal{Q}_0$ of inactive documents. We sample a stratified group of $N$ documents across bins, crop each to a short snippet centered on the highest-activating token, and add angled-bracket highlights (e.g., \texttt{{<}{<}word{>}{>}}) to any token in the snippet whose activations are more than 10\% of the feature's global maximum. We then prompt \texttt{gemini-3.1-flash-lite} with these quantile-grouped snippets to produce a concise (${\leq}20$ word) description of the feature's behavior.

\paragraph{Description Validation} To validate our generated descriptions we employ a quantile-based variant of the fuzzing procedure introduced in \citet{paulo2024automatically}. Given a generated description and a list of documents, our pipeline prompts \texttt{gemini-3.1-flash-lite} to predict which quantile bin (0-$k$) each document was sampled from. We compute the simulation score $s_i$ as the Spearman correlation between the predicted and true quantile assignments across all documents. In our experiments we use 10 documents for generation and 50 documents for validation. Full details including prompts, decoding strategies, and other hyperparameters are listed in Appendix~\ref{app:autointerp}.

\begin{figure}
\centering
\includegraphics[width=\columnwidth]{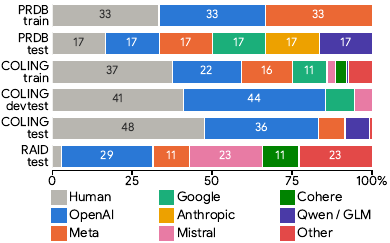}
\caption{Distribution of generative models for COLING, PRDB, and RAID datasets broken up by model family. We see that both our training and evaluation data span a large variety of model families.}
\label{fig:dataset_mix}
\end{figure}

\begin{figure*}
\centering
\includegraphics[width=\textwidth]{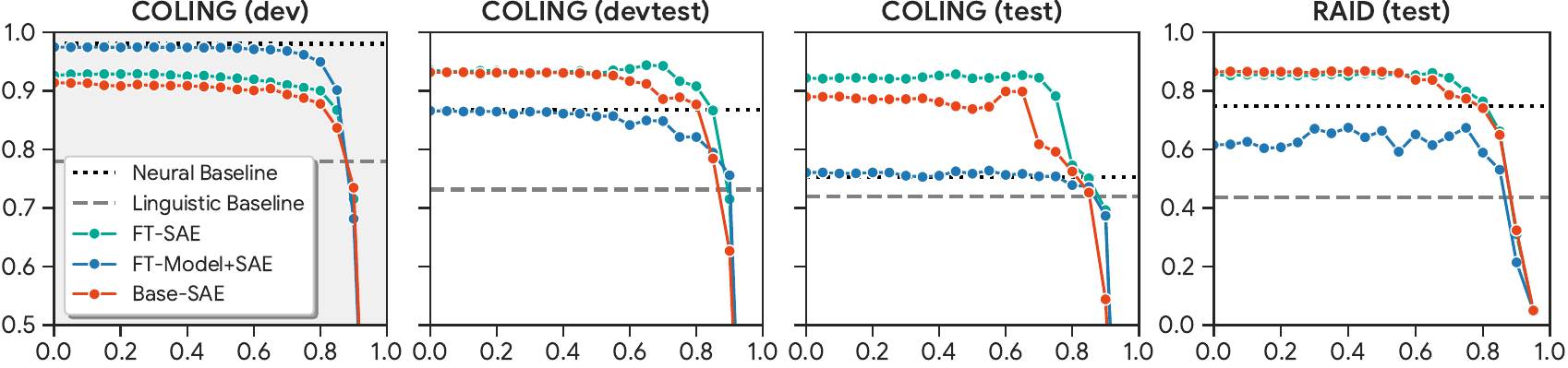}
\caption{F1 Score (COLING) and TPR@FPR=5\% (RAID) (y-axis) vs. Interpretability Threshold (x-axis) for classifiers trained on the COLING Shared Task 1 dataset evaluated on ID (grey) and OOD (white) data. We see that \textsc{FT-Model+SAE} matches the performance of the underlying neural detector even at high interpretability thresholds (0.7+) and that \textsc{Base-SAE} and \textsc{FT-SAE} models exceed the detector's performance on OOD data.}
\label{fig:four_panel}
\end{figure*}

\begin{table*}[t]
\centering
\small
\setlength{\tabcolsep}{5pt}
\begin{tabular}{lccccccccccccc}
\toprule
& \multicolumn{3}{c}{\textbf{COLING (dev)}} & \multicolumn{3}{c}{\textbf{COLING (devtest)}} & \multicolumn{3}{c}{\textbf{COLING (test)}} & \multicolumn{3}{c}{\textbf{RAID (test)}} \\
\cmidrule(lr){2-4} \cmidrule(lr){5-7} \cmidrule(lr){8-10} \cmidrule(lr){11-13}
 & AUC & F1 & TPR@$5$ & AUC & F1 & TPR@$5$ & AUC & F1 & TPR@$5$ & AUC & F1 & TPR@$5$ \\
\midrule
Binoculars$^\dagger$     & 75.8 & 73.1 & 56.0 & 88.6 & 81.6 & 67.2 & 86.7 & 81.4 & 59.9 & 85.0 & 87.6 & 79.1 \\
Fast-DetectGPT$^\dagger$ & 75.7 & 70.5 & 56.2 & 83.2 & 75.5 & 51.6 & 85.1 & 77.5 & 58.5 & 83.3 & 85.2 & 75.1 \\
ReMoDetect$^\dagger$     & 76.5 & 65.1 & 45.6 & 91.8 & 85.2 & 65.2 & 91.6 & 86.7 & 75.0 & 73.2 & 67.8 & 44.1 \\
SpecDetect$^\dagger$     & 80.2 & 74.4 & 57.2 & 81.3 & 74.2 & 48.9 & 83.6 & 74.6 & 56.4 & 91.6 & 88.1 & 77.3 \\
\midrule
\textbf{Baselines} && & & & & & & & & & & \\
gram2vec & 81.9 & 77.9 & 36.2 & 80.1 & 73.1 & 45.3 & 77.7 & 72.0 & 33.6 & 77.5 & 73.5 & 43.6 \\
gemma-2-2b & \textbf{99.8} & \textbf{98.0} & \textbf{98.9} & 92.5 & 86.8 & 77.9 & 85.6 & 75.2 & 53.2 & 93.6 & 89.3 & 74.9 \\
\midrule
\textbf{w/o Mask} && & & & & & & & & & & \\
\textsc{Base-SAE}     & 97.1 & 91.4 & 84.0 & 96.2 & 93.2 & 90.4 & 94.2 & 88.9 & 81.8 & 95.2 & 92.3 & \textbf{86.4} \\
\textsc{FT-SAE}       & 97.8 & 92.6 & 86.8 & 96.7 & 93.5 & 91.2 & \textbf{95.9} & 92.2 & 89.1 & \textbf{95.3} & 91.8 & 85.5 \\
\textsc{FT-Model+SAE} & 99.5 & 97.5 & 98.1 & 91.0 & 86.6 & 63.5 & 83.1 & 76.1 & 47.8 & 91.7 & 92.5 & 61.5 \\
\midrule
\textbf{w/ Mask ($\tau>0.7$)} && & & & & & & & & & & \\
\textsc{Base-SAE}     & 95.2 & 89.4 & 75.5 & 93.8 & 88.6 & 80.5 & 88.2 & 80.9 & 61.3 & 93.0 & 89.8 & 78.6 \\
\textsc{FT-SAE}     & 96.9 & 91.0 & 81.4 & \textbf{96.8} & \textbf{94.2} & \textbf{92.3} & 95.3 & \textbf{92.2} & \textbf{89.2} & 94.5 & 91.7 & 84.4 \\
\textsc{FT-Model+SAE}& 99.2 & 96.8 & 96.7 & 91.1 & 84.8 & 67.9 & 82.4 & 75.4 & 49.1 & 92.0 & \textbf{93.2} & 64.5 \\
\bottomrule
\end{tabular}
\caption{Model performance across held out splits of RAID and COLING datasets. Dagger $(\dagger)$ indicates off-the-shelf models not trained on the COLING train set. We see that \textsc{FT-Model+SAE} matches the performance of the underlying neural detector, that \textsc{Base-SAE} and \textsc{FT-SAE} models exhibit strong generalization performance, and that interpretability masking does not substantially reduce performance on OOD examples.}
\label{tab:results}
\end{table*}

\section{Experimental Setup}

\subsection{Data}
\paragraph{Open Domain} In this setting we train the model to detect AI generated text from an arbitrary AI model in an arbitrary domain. We train our models on the English split of the COLING shared task dataset \cite{wang-etal-2025-genai} a collection spanning 41 highly diverse language models paired with human texts. We use the official train and dev splits (610k and 261k documents) for training and in-distribution testing. For evaluating out-of-distribution generalization we use the official english devtest and test splits of the COLING dataset as well as the non-adversarial test split of the RAID leaderboard \cite{dugan-etal-2024-raid}.

\paragraph{Closed Domain (Paper Reviews)}
For our closed setting we fine-tune models to detect AI generated academic peer reviews using the Peer Review Detection Benchmark (PRDB) \cite{yu2026paperreviewedllmbenchmarking}. This contains 106,413 human-written peer reviews for the NeurIPS and ICLR conferences for years 2016-2024 along with paired AI generated texts. For training we use the extended split and separate it into two sets: Human vs. GPT-4o and Human vs. Llama-3.1-70B. For evaluation we use the combined calibration and test splits, which contain held out human texts and paired generations from GPT-4o, Llama 3.1 70B, Qwen 2.5 72B, Claude Sonnet 3.5, and Gemini 1.5 Pro.

\subsection{Baselines}
\label{sec:baselines}

\paragraph{Neural (gemma-2-2b)} For our neural baseline we fine-tune the gemma-2-2B model for the AI detection task. We replace the language modeling head with a classification head and train to minimize binary cross-entropy on the AI detection task. We verify that our neural model outperforms the baseline RoBERTa model from the COLING shared task \cite{wang-etal-2025-genai} on both in-distribution and out-of-distribution texts. Full training details can be found in Appendix \ref{app:training_details}.

\paragraph{Linguistic (gram2vec)} For our stylometric baseline we fine-tune a $L_1$-regularized Logistic Regression classifier on document embeddings produced by gram2vec \cite{zeng2025gram2vecinterpretabledocumentvectorizer}. These embeddings contain over 930 features from seven linguistic categories: punctuation marks, parts of speech, morphology tags, dependency labels, function words, emojis, and transition words. These features have been shown to outperform Biber features \cite{Biber1988} for both authorship attribution and AI detection.

\paragraph{Off-the-Shelf} In addition to our trained baselines, we compare against four off-the-shelf detectors: Binoculars \cite{binoculars}, Fast-DetectGPT \cite{bao2023fast}, ReMoDetect \cite{lee2024remodet}, and SpecDetect \cite{haitong2026spec}. These detectors are not trained on COLING; we apply them zero-shot using their publicly released implementations.

\subsection{Models}
Our models are built off of GemmaScope \cite{lieberum-etal-2024-gemma}, a collection of pre-trained SAEs trained to reconstruct activations of gemma-2-2b on arbitrary text. For our core comparisons we do the following:
\begin{enumerate}[itemsep=0.15em, topsep=0.6em]
\item \textbf{\textsc{Base-SAE}}: Use the pre-trained GemmaScope SAEs without any additional fine-tuning.
\item \textbf{\textsc{FT-SAE}}: Fine-tune the GemmaScope SAEs to better reconstruct base gemma-2-2b's model activations on in-domain examples from the AI detection task.
\item \textbf{\textsc{FT-Model+SAE}}: Fine-tune gemma-2-2b for AI detection, then fine-tune GemmaScope SAEs to reconstruct the fine-tuned model's activations on in-domain examples from the AI detection task.
\end{enumerate}
To train our SAEs we cache raw activations for all tokens in the training corpus and uniformly shuffle batches across all tokens in the dataset. For more details on our training procedure see Appendix \ref{app:train_saes}.

\section{Findings}
\label{sec:findings}
\paragraph{\textsc{FT-Model+SAE} recovers the performance of the underlying AI detector.} In Table \ref{tab:results} we report the results of our detector models on the in-distribution dev split and the three held out out-of-distribution test splits for the COLING dataset. We find that in all three cases, the \textsc{FT-Model+SAE} condition matches the AUC and F1 score performance of the full AI detector---only losing ground once we restrict to lower false positive rate. Furthermore, when restricting to highly interpretable features ($\tau>0.7$), our \textsc{FT-Model+SAE} system is still able to recover (and even improve) the performance of the full model despite only having a fraction of the features unmasked.

However, when the underlying detector fails to generalize, so too does our system. In Figure \ref{fig:four_panel} we see that \textsc{FT-Model+SAE} underperforms our two base model systems by fairly wide margins on out-of-distribution data. As we will discuss in the next section, the surprising aspect of this is the strong performance of these base model systems, not the failure of the \textsc{FT-Model+SAE} system to surpass the performance of the underlying detector it was trained on.

\paragraph{Classifiers built from base model SAEs are accurate and robust AI detectors.}
In Table \ref{tab:results} we see that \textsc{Base-SAE} and \textsc{FT-SAE} perform very well, especially when applied to out-of-distribution data---beating well established baselines such as Binoculars \cite{binoculars} and FastDetectGPT \cite{bao2023fast} by around 8-12 F1 score. This is especially surprising for the \textsc{Base-SAE} condition, as the only trainable parameters in the system are the logistic regression weights.

We follow \citet{quaremba2026linearprobingprovidesrobust} in hypothesizing that the base model has learned certain linearly separable ``AI text'' features during pre-training. In the next sections, we include all three models as strong detectors for testing and leave further investigation into the performance of these classifiers to future work.

\paragraph{AI-generated Peer Reviews are too easy to detect.} In Table \ref{tab:prdb_results_indomain} we report results on PRDB and find that all trained classifiers achieve nearly perfect performance, even at very low false positive rates. This suggests the presence of confounds or other spurious features in the dataset that make the task artificially easy. This is a good use case for our pipeline. In Section \ref{sec:qualitative_analysis}, we will audit the specific features our PRDB detectors are using and determine whether they are abusing spurious cues.

\begin{table}[t]
\centering
\footnotesize
\setlength{\tabcolsep}{1.6pt}
\begin{tabular}{lcccccc}
\toprule
& \multicolumn{3}{c}{\textbf{GPT-4o}} & \multicolumn{3}{c}{\textbf{Llama 3.1 70B}} \\
\cmidrule(lr){2-4} \cmidrule(lr){5-7}
& AUC & F1 & TPR@1 & AUC & F1 & TPR@1 \\
\midrule
Binoculars$^\dagger$     & 95.7 & 89.8 & 34.5 & 99.9 & 99.0 & 99.1 \\
SpecDetect$^\dagger$     & 92.2 & 84.4 & 29.4 & 98.4 & 93.8 & 80.9 \\
ReMoDetect$^\dagger$     & 99.4 & 98.4 & 89.6 & 99.8 & 99.0 & 98.5 \\
Fast-DetectGPT$^\dagger$ & 81.0 & 74.1 & 10.6 & 99.3 & 98.6 & 85.4 \\
\midrule
\multicolumn{7}{l}{\textbf{Baselines}} \\
gram2vec  & 100 & 99.6 & 100 & 100 & 99.4 & 99.7 \\
hf-gemma2 & 100 & 100  & 100 & 100 & 100  & 100 \\
\midrule
\multicolumn{7}{l}{\textbf{w/o Mask}} \\
\textsc{Base-SAE}     & 100 & 99.8 & 100 & 100 & 99.8 & 100 \\
\textsc{FT-SAE}       & 100 & 99.9 & 100 & 100 & 99.9 & 100 \\
\textsc{FT-Model+SAE} & 100 & 99.9 & 100 & 100 & 99.9 & 100 \\
\midrule
\multicolumn{7}{l}{\textbf{w/ Mask ($\tau\!>\!0.7$)}} \\
\textsc{Base-SAE}     & 100 & 99.8 & 100 & 100 & 99.8 & 100 \\
\textsc{FT-SAE}       & 100 & 99.9 & 100 & 100 & 99.9 & 100 \\
\textsc{FT-Model+SAE} & 100 & 99.9 & 100 & 100 & 99.9 & 100 \\
\bottomrule
\end{tabular}
\caption{Performance on in-distribution test split of PRDB. Dagger $(\dagger)$ indicates off-the-shelf models. We see that all models achieve near perfect performance.}
\label{tab:prdb_results_indomain}
\end{table}

\begin{figure*}[t]
  \centering
  \includegraphics[width=\textwidth]{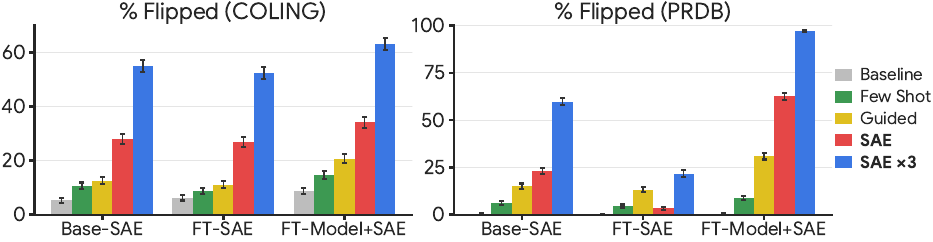}
  \caption{Percentage of AI-written documents whose predictions were flipped to ``Human'' after paraphrasing with Gemini using each of our five prompting strategies: Baseline, Few Shot, and Guided are generic paraphrase prompts, while SAE and SAEx3 use our interpretable features one-shot and iteratively ($n=3$). We see that having access to SAE features in the prompt allows the model to more easily humanize the text. See Table \ref{tab:adv_appendix_all} for extended metrics.}
  \label{fig:adv_flip}
\end{figure*}

\begin{table*}[t]
\centering
\small
\setlength{\tabcolsep}{4pt}
\begin{tabular}{lcccccccc}
\toprule
& \multicolumn{4}{c}{\textbf{w/o Mask}} & \multicolumn{4}{c}{\textbf{w/ Mask ($\tau\!>\!0.7$)}} \\
\cmidrule(r){2-5} \cmidrule(l){6-9}
& Num & Avg. Feat Shift & Same & \textbf{Importance} & Num & Avg. Feat Shift & Same & \textbf{Importance} \\[-1pt]
& Rewrites & $\{\Delta_{\text{tgt}},\ \Delta_{\text{non}}\}$ & Sign & \textbf{vs.\ Para.\ $\boldsymbol{\rho}$} & Rewrites & $\{\Delta_{\text{tgt}},\ \Delta_{\text{non}}\}$ & Sign & \textbf{vs.\ Para.\ $\boldsymbol{\rho}$} \\
\midrule
\textbf{COLING} & & & & & & & & \\
\textsc{Base-SAE}     & 16,052 & \{0.31,\ 0.17\} & 65.0\% & 0.36 & 19,318 & \{0.38,\ 0.12\} & 59.9\% & 0.41 \\
\textsc{FT-SAE}       & 15,968 & \{0.36,\ 0.23\} & 72.9\% & 0.52 & 17,826 & \{0.37,\ 0.14\} & 80.8\% & 0.54 \\
\textsc{FT-Model+SAE} & 28,965 & \{0.25,\ 0.22\} & 68.9\% & 0.54 & 27,866 & \{0.63,\ 0.27\} & 73.2\% & 0.62 \\
\midrule
\textbf{PRDB} & & & & & & & & \\
\textsc{Base-SAE}     & 35,339 & \{0.55,\ 0.12\} & 92.1\% & 0.76 & 30,803 & \{0.63,\ 0.07\} & 89.2\% & 0.81 \\
\textsc{FT-SAE}       & 37,037 & \{0.43,\ 0.19\} & 91.3\% & 0.74 & 36,720 & \{0.55,\ 0.12\} & 92.4\% & 0.80 \\
\textsc{FT-Model+SAE} & 40,032 & \{0.41,\ 0.16\} & 93.9\% & 0.69 & 37,494 & \{0.74,\ 0.14\} & 99.1\% & 0.80 \\
\bottomrule
\end{tabular}
\caption{Results of the feature-level paraphrase ablation. We report correlation between importance score and paraphrase ablated logit difference. $\Delta_{\text{tgt}}$ and $\Delta_{\text{non}}$ give the normalized activation shift of the targeted and non-targeted features under the paraphrase as well as the sign agreement. We see features are attributed faithfully across both datasets and see meaningful improvement with $\tau > 0.7$ indicating its utility as an interpretability heuristic.}
\label{tab:counterfactual}
\end{table*}

\section{Paraphrase Ablation}
\label{sec:causal_ablation}
In the previous section we established that our detectors are highly accurate. However, we still need to provide causal evidence that our feature descriptions faithfully represent the underlying dynamics of the model beyond simple auto-interpretation scores. We do this in two ways: a feature-level paraphrase ablation and a document-level ablation.\footnote{For an extended analysis of both experiments along with examples and more plots, see Appendix \ref{app:paraphrase-ablation}.}

\subsection{Feature-Level Paraphrase Ablation}

\paragraph{Setup} In this experiment we want to ablate individual features and observe whether the change in downstream prediction matches what our pipeline expects. To do this we start by subsampling 500 human and 500 AI documents from both COLING and PRDB. For a given model, we run a forward pass on all documents and get all active features that contribute to the downstream prediction. For each feature we do a targeted paraphrase; we feed the text description of that feature to \texttt{gemini-3.1-flash-lite} and ask it to paraphrase the document to remove only that feature. We report correlation between the mean paraphrase ablated logit delta and the mean importance score per feature. If our pipeline is faithful, then the downstream logit effect of paraphrasing away a feature will exactly equal the importance score of that feature.

\paragraph{Findings} Table \ref{tab:counterfactual} shows the results of the ablation. Across all twelve configurations, the importance score for a feature positively predicts the effect of paraphrasing it away. This effect is strongest on PRDB, where Spearman correlations reach $\rho = 0.69$--$0.81$ and 89--99\% of features agree in sign between the two interventions; on COLING agreement is positive but weaker ($\rho = 0.36$--$0.62$). We also see that restricting the detector to features with auto-interp score above $0.7$ improves the importance-vs-paraphrase correlation, most sharply for \textsc{FT-Model+SAE} ($0.54 \rightarrow 0.62$ on COLING, $0.69 \rightarrow 0.80$ on PRDB).

\begin{table*}[t]
\centering
\small
\setlength{\tabcolsep}{5pt}
\renewcommand{\arraystretch}{1.2}
\begin{tabular}{@{}llp{7.5cm}ccc@{}}
\toprule
\textbf{Model} & \textbf{Feature} & \textbf{Description} \textit{(auto-interp: gemini-3.1-flash-lite)} & $\tau$ & \textbf{Logit $\Delta$} & \textbf{\#Docs} (Human\,/\,AI)\\
\midrule
\textbf{GPT-4o} & 13\_15950 & academic paper review questions asking for more information details or clarification & 0.866 & $+1.002$ & $1{,}400$\,/\,$79{,}922$ \\
                & 17\_8459  & academic review feedback or peer review decisions & 0.870 & $+0.814$ & $229$\,/\,$96{,}612$ \\
                & 22\_4800  & double colons following bolded markdown section headers & 0.860 & $+0.672$ & $11$\,/\,$54{,}293$ \\
                & 17\_12216 & activates on the word ``potential'' & 0.884 & $+0.515$ & $15{,}012$\,/\,$92{,}929$ \\
\midrule
\textbf{Llama}  & 17\_4373  & the word ``thorough'' & 0.892 & $+0.993$ & $9{,}606$\,/\,$79{,}708$ \\
                & 13\_12061 & numbered list items in parentheses within question sections of research papers (\emph{e.g.,} \texttt{(1)}, \texttt{(2)}) & 0.858 & $+0.898$ & $21$\,/\,$27{,}425$ \\
                & 7\_5460   & the word one when it appears at the start of a sentence or bullet point & 0.851 & $+0.745$ & $3{,}339$\,/\,$53{,}564$ \\
                & 13\_164   & opening square bracket in list or string literals & 0.866 & $+0.610$ & $7$\,/\,$22{,}541$ \\
\midrule
\textbf{Human}  & 13\_5785  & first person singular pronoun I at start of sentences or after punctuation & 0.866 & $-0.919$ & $78{,}205$\,/\,$10{,}209$ \\
                & 7\_1616   & the word see used as a directive or reference to look elsewhere in academic reviews & 0.867 & $-0.836$ & $42{,}744$\,/\,$755$ \\
                & 7\_10672  & word seem or seems when used to express an opinion or observation about a paper or argument & 0.866 & $-0.808$ & $59{,}993$\,/\,$1{,}133$ \\
                & 22\_773   & closing parenthesis followed by a period or similar closing punctuation & 0.853 & $-0.745$ & $45{,}779$\,/\,$30$ \\
\bottomrule
\end{tabular}
\caption{Top features for \textsc{FT-Model+SAE} ($\tau > 0.85$) on the PRDB dataset \cite{yu2026paperreviewedllmbenchmarking}, ranked by the mean change in detector logit when the feature is zero-ablated (Logit $\Delta$). Each row reports a feature's auto-interp explanation, simulation score $\tau$, Logit $\Delta$, and number of occurrences in human and AI text.}
\label{tab:top_features}
\end{table*}

\subsection{Document-Level Paraphrase Ablation}

\paragraph{Setup} In this experiment we want to show that the set of features given by our pipeline in aggregate represent all of the features used by the model to make its prediction. To do this we conduct an adversarial humanization attack. We start by sampling 2,500 AI documents from both datasets. For each document we query \texttt{gemini-3.1-flash-lite} with five different prompts to get five different paraphrases. The first prompt ``Baseline'' simply asks the model to rewrite the text to make it more human. The second ``Few Shot'' includes few shot examples of human documents from the train split of the dataset. The ``Guided'' prompt includes instructions with features that are colloquially associated with AI and explicitly asks to remove them.

The final two prompts explicitly use our system. The ``SAE'' prompt includes the list of active features with their associated descriptions and importances and asks the model to remove all AI features. Finally, the ``SAE x3'' prompt iterates over the ``SAE'' prompt three times, re-querying the pipeline for the new most active features each time. We do each of these once per model condition. All prompts explicitly tell the model to edit less than 30\% of tokens in the document.

\paragraph{Results} In Figure \ref{fig:adv_flip} we report the results of the adversarial humanization. We see that our interpretable detectors are surprisingly robust to humanizing paraphrases, with the baseline  flipping at most 8.7\% of the documents' predictions and frequently 0\%. Next, we see that including our pipeline's interpretable features improves flip rate substantially more than either of the generic prompts across all models. Allowing the model to iterate further increases this gap, with a success rate of 60-100\% across all conditions all while maintaining a low edit distance (<40\% of tokens in the document edited). While it is not definitive proof, this result does suggest that our lists of features are relatively complete and high recall---paraphrasing away existing features flips predictions rather than revealing new AI features. 

\section{Qualitative Analysis}
\label{sec:qualitative_analysis}
In Table \ref{tab:top_features} we list the most predictive features for \textsc{FT-Model+SAE} $(\tau > 0.85)$ on AI Peer Reviews, ranked by mean feature importance score (Logit $\Delta$). We report the top features pushing the detector towards GPT-4o and Llama (positive Logit $\Delta$) as well as those pushing towards Human (negative Logit $\Delta$). Extended tables covering all three systems on both PRDB and COLING can be found in Tables \ref{tab:top_features_coling_ftsae} through \ref{tab:top_features_prdb_ftsae_llama} in the Appendix.

We first note that the most predictive features are highly specific, frequently activating on only one or two tokens or on formatting characters and punctuation marks. For example, feature 22\_4800 (``double colons following bolded markdown section headers'') appears in only 11 human documents but over 54,000 GPT-4o documents, and feature 13\_164 (``opening square bracket in list or string literals'') appears in just 7 human documents but over 22,000 Llama documents. 

In addition, we observe that the most predictive features for human text are largely stylistic rather than structural: first person pronouns at the start of sentences (13\_5785), hedges such as ``seem'' or ``seems'' used to express an opinion (7\_10672), and the word ``see'' used as a directive to look elsewhere in the review (7\_1616). The extended tables reveal further differences between systems: on COLING, the top \textsc{FT-SAE} feature (13\_3738) activates on ``introductory phrases or filler text commonly used by ai language models when declining personal questions'' (\emph{e.g.,} ``As an AI language model''), while the top \textsc{Base-SAE} features are dominated by topical and domain-level signals such as HTML attributes and software license boilerplate. This suggests that detectors built on base model representations may rely more heavily on domain shortcuts than on stylistic properties of the text itself.

Given the near perfect performance of these models at detecting AI-generated peer reviews (Section \ref{sec:findings}), one could reasonably claim that such features are spurious and should not be used in detection. In their analysis, \citet{kantamneni2025sparseautoencodersusefulcase} claimed as much, and recommended all formatting based features be removed. However, it is unclear what should and should not be considered a spurious feature for the purposes of AI detection. If a formatting quirk is consistently present in AI-written text across contexts, there's a good argument to be made that it should be included. Ultimately, the ability for a model provider to audit and decide on a case-by-case basis whether or not a particular feature should be used by a particular AI detector is a key affordance of our system.

\section{Conclusion}
In order to be able to use AI detectors in real-world high-stakes situations, they need to be auditable. It is unreasonable to expect major decisions to be made at the behest of an AI detector that cannot provide a rationale for its output. Current interpretable stylometric AI detectors underperform state-of-the-art neural AI detection methods by wide margins---making them infeasible for widespread use.

In this paper we propose a technique to take an existing black box neural AI detector and convert it into a detector that operates on human-interpretable concepts. We accomplish this by training Sparse Autoencoders on the detector's residual stream and training a lightweight classifier over only the features that pass a given global interpretability threshold. We find that classifiers constructed in this way retain the performance and generalization properties of the underlying base model while being substantially more understandable. Extensive ablations show that our feature descriptions have high recall and high precision, faithfully representing what the features fire on and how they contribute to the final prediction.

In addition to this, we find that classifiers based on SAE decompositions of base model activations are surprisingly robust detectors of AI generated text. We find that they outperform strong well-established baselines in the AI detection literature and encourage future work to follow up with further experimentation on using internal activations of language models as input to AI detectors.

Finally, we analyze the features used by AI detectors and find that they frequently use highly specific token sequences and formatting elements which may fail to generalize to other domains and models. Our findings raise questions about which features are valid to use in an AI detection system and we encourage future work to investigate adversarial hardening techniques using this style of unsupervised concept discovery.

\section*{Limitations}
\paragraph{Feature descriptions are biased towards recall} One core limitation of our work is a limitation shared by the current literature, namely that auto-interpretation pipelines systematically produce descriptions that are high recall but low precision \cite{gao2024scalingevaluatingsparseautoencoders}. For example, say an SAE feature fires on the token ``Tom'' but only after the token ``Hello''. If our dataset only ever contains the token ``Tom'' in this context, then our pipeline will likely generate a description for this feature that misses the dependency. That same description would also receive a perfect auto-interpretation score by the validation pipeline for the same reason. 

In their work, \citet{gao2024scalingevaluatingsparseautoencoders} attempt to measure the effect size of this issue using Neuron2Graph explanations \cite{foote2023neurongraphinterpretinglanguage} targeting specific ngrams in the training corpus. We attempted something similar in an informal capacity and asked language models to generate regular expressions that matched the generated description of a subset of our features. We measured the precision and recall using the regular expression as a ground truth and found that, while roughly half of our features did manage to get perfect descriptions (99.5+ F1 score), the other half exhibited this same recall bias to varying degrees. We believe that better sampling of hard negatives and synthetic data generation are good potential avenues to reduce this issue.

\paragraph{Lack of testing on different base models} Another limitation of our work is the fact that we only test on fine-tuned GemmaScope SAEs and do not train SAEs on other base models. While other trained SAEs do exist for models like Llama and Qwen, we chose GemmaScope as it is by far the most popular and was used by previous work for AI detection \cite{kuznetsov-etal-2025-feature}. Future work should seek to expand the breadth of base models tested, especially as it relates to the performance of the \textsc{Base-SAE} and \textsc{FT-SAE} conditions.

\section*{Acknowledgements}
This work was greatly helped by discussions and feedback from John Wieting, Neel Nanda, Adam Lelkes, Maty Bohacek, Elyas Masrour, Arthur Szlam, and Asic Chen. We also thank the Penn NLP group, in particular Josh Ludan, Grace Kim, Hita Kambhamettu, Allen Chang, and Alyssa Hwang for their help during the writing process.

This research was developed with funding from the Defense Advanced Research Projects Agency's (DARPA) SciFy program (Agreement No. HR00112520300) and is based upon work supported in part by the Office of the Director of National Intelligence (ODNI), Intelligence Advanced Research Projects Activity (IARPA), via 56000026C0019 (the BENGAL program). The views and conclusions contained herein are those of the authors and should not be interpreted as necessarily representing the official policies, either expressed or implied, of DARPA, ODNI, IARPA, the Department of Defense, or the U.S. Government. The U.S. Government is authorized to reproduce and distribute reprints for governmental purposes notwithstanding any copyright annotation therein.

\bibliography{custom}

\appendix

\clearpage
\section{Training Sparse Autoencoders}
\label{app:train_saes}
\subsection{Hyperparameters \& Training Settings}
For our trained SAEs we fine-tune GemmaScope \cite{lieberum-etal-2024-gemma} on in-domain data for the AI detection task. In Table \ref{tab:sae_hparams} we report the hyperparameters used to train our sparse autoencoders. We initialize GemmaScope using the \texttt{width\_16k/canonical} checkpoints from the \texttt{gemma-scope-2b-pt-res-canonical} release and train on the following five settings: 
\begin{enumerate}[itemsep=0.15em, topsep=0.6em]
    \item \textbf{COLING (base)} Reconstruct base Gemma-2-2B activations on the COLING train set
    \item \textbf{COLING (Open)} Reconstruct activations of the Gemma-2-2B model fine-tuned to detect generic AI generated text on the COLING train set
    \item \textbf{PRDB (base)} Reconstruct base Gemma-2-2B activations on the PRDB train set
    \item \textbf{PRDB (GPT-4o)} Reconstruct activations of the Gemma-2-2B model fine-tuned to detect GPT-4o on the PRDB train set
    \item \textbf{PRDB (LLama)} Reconstruct activations of the Gemma-2-2B model fine-tuned to detect Llama on the PRDB train set
\end{enumerate}
For each setting we train four models (one for each layer 7, 13, 17, 22). When training on PRDB we do not filter by target model and train all SAEs to reconstruct activations from both GPT-4o and Llama as well as human text.

We use the JumpReLU SAE \cite{rajamanoharan2024jumpingaheadimprovingreconstruction} architecture and rescale $\mathbf{W}_{\text{dec}}$ rows to unit norm after each update. We project out all parallel gradient components before the optimizer step following \cite{bricken2023monosemanticity}. Our downstream objective is the standard reconstruction loss plus a sparsity penalty without any explicit auxiliary loss terms or resampling of dead latents.

\subsection{Re-parameterization}
During training, GemmaScope SAEs train to reconstruct standardized model activations of the form
\begin{equation}
\mathbf{d}_{\text{std}} = \frac{\mathbf{d_{\text{raw}}}}{\text{C}} - \mathbf{b}_{\text{dec}}
\end{equation}
where $\text{C}$ is a scalar constant such that $\mathbf{E}[||\frac{\mathbf{d_{\text{raw}}}}{C}||_2^2] = 1$. Before release, these scaling parameters were folded back into the model weights so that the trained models could be applied directly to raw model activations without needing to standardize. 

\begin{table}[t]
\centering
\begin{tabular}{lr}
\toprule
\multicolumn{2}{c}{\textbf{SAE Model Hyperparameters}}\\
\midrule
\texttt{optimizer}* & \texttt{optim.Adam}\\
$\beta_1, \beta_2, \epsilon$* & $0.0, 0.999, 10^{-8}$\\
\texttt{sae\_width}* & $16,384$\\
\texttt{sparsity\_coeff}* & $6\times 10^{-4}$\\
\texttt{sparsity\_warmup\_steps}* & $10,000$\\
\texttt{sp\_scheduler\_type}* & \texttt{linear}\\
\texttt{learning\_rate}*&$7\times 10^{-5}$\\
\texttt{lr\_scheduler\_type}* &\texttt{cosine}\\
\texttt{warmup\_steps}*& $1000$\\
\texttt{batch\_size}*&$4096$\\
\texttt{bandwidth}*&$0.001$\\
\texttt{num\_epochs} & $4$\\
\texttt{layers} & $7,13,17,22$\\
\texttt{release}& \texttt{res-canonical}\\
\bottomrule
\end{tabular}
\caption{Hyperparameters used for fine-tuning the GemmaScope \cite{lieberum-etal-2024-gemma} SAEs. Asterisks indicate settings inherited from GemmaScope.}
\label{tab:sae_hparams}
\end{table}

\begin{table*}[t]
\centering
\small
\begin{tabular}{lcccccccc}
\toprule
& \multicolumn{2}{c}{\textbf{Layer 7}} & \multicolumn{2}{c}{\textbf{Layer 13}} & \multicolumn{2}{c}{\textbf{Layer 17}} & \multicolumn{2}{c}{\textbf{Layer 22}} \\
\cmidrule(lr){2-3} \cmidrule(lr){4-5} \cmidrule(lr){6-7} \cmidrule(lr){8-9}
 & \textsc{L0} & \textsc{FVU} & \textsc{L0} &  \textsc{FVU}  & \textsc{L0} &  \textsc{FVU}  & \textsc{L0} &  \textsc{FVU}  \\
\midrule
\textbf{COLING (base)} & 48.73 & 11.27\% & 67.77 & 15.49\% & 66.60 & 13.53\% & 73.97 & 13.8\% \\
\textbf{COLING (open)} & 40.19 & 3.71\% & 24.03 & 3.37\% & 16.04 & 3.78\% & 7.18 & 5.22\% \\
\textbf{PRDB (base)} & 31.57 & 7.18\% & 46.35 & 10.50\% & 45.01 & 8.53\% & 44.87 & 8.56\% \\
\textbf{PRDB (GPT-4o)} & 12.59 & 2.50\% & 7.20 & 2.95\% & 5.87 & 2.68\% & 4.64 & 3.47\% \\
\textbf{PRDB (Llama)} & 16.02 & 1.50\% & 9.28 & 1.38\% & 6.67 & 1.50\% & 5.24 & 2.15\% \\
\bottomrule
\end{tabular}
\caption{Evaluation results for our fine-tuned SAEs on the eval splits of the COLING and PRDB dataset. We report number of active features per example (L0) and reconstruction fidelity (FVU) and note that on average, base model SAEs have worse reconstruction performance than the fine-tuned model SAEs.}
\label{tab:sae_results}
\end{table*}

While this is convenient for inference, it complicates fine-tuning, since we must undo this procedure without knowledge of the constant $\text{C}$. In our pipeline, we empirically compute a local scaling constant $C_\ell$ on our train set for each layer $\ell$
\begin{equation}
\text{C}_{\ell} = \sqrt{\frac{1}{N} \sum_{i=1}^{N} \|\mathbf{d}^{(\ell)}_i\|_2^2}
\end{equation}
and re-scale the GemmaScope model weights by
\begin{align}
\mathbf{\hat{\theta}}^{(\ell)} &= \mathbf{\theta}^{(\ell)} \;/\; \text{C}_{\ell}\\
\mathbf{\hat{b}}_{\text{dec}}^{(\ell)} &= \mathbf{b}_{\text{dec}}^{(\ell)} \;/\; \text{C}_{\ell}\\
\mathbf{\hat{b}}_{\text{enc}}^{(\ell)} &= \mathbf{b}_{\text{enc}}^{(\ell)} \;/\; \text{C}_{\ell} + \mathbf{W}_{\text{enc}}(\mathbf{b}_{\text{dec}}^{(\ell)} \;/\; \text{C}_{\ell})
\end{align}
before the start of training.

After training is finished, we follow \citet{lieberum-etal-2024-gemma} and re-fold the normalization constant and pre-encoder bias back into the model parameters so that inference can once again be applied on raw model activations.

\subsection{Training Optimizations \& Hardware}
\paragraph{Buffered Token Shuffling}
When constructing batches for inference we want to uniformly shuffle tokens across different documents to avoid training instability. However, actually implementing this requires caching activations on the entire dataset, which requires a large amount of disk space.

To implement this, we first sample a pool of documents and bin them according to sequence length. We then run model inference on these batches and extract the per-token activations at each layer. We load these activations into an activations buffer and shuffle that buffer uniformly before sampling batches for inference. Using a buffer of 1,000,000 tokens results in a reasonably uniform shuffle across token sequences during inference without requiring us to cache activations.

To speed up training and save time caching activations, we trained all SAEs for all layers of a given model simultaneously. This limited the need for many costly forward passes of the base model.

\paragraph{Hardware and Training Time}
We trained using NVIDIA DGX B200 GPUs for a total of 185 GPU hours across all model conditions.

\subsection{Evaluation}
In Table \ref{tab:sae_results} we report the average number of active features per example (L0) and fraction of variance unexplained (FVU) achieved by our fine-tuned SAEs across each layer on their respective validation sets. We see that on average the fine-tuned model SAEs have much lower L0 and reconstruction error than the base model SAEs. This is despite the fact that in our main results, the base model SAEs performed far better on the downstream generated text detection task. This echoes findings from \citet{karvonen2025saebenchcomprehensivebenchmarksparse} that show simple reconstruction-based evaluations for SAEs can fail to correlate with actual desired downstream effects.

We also observe that the fine-tuned model SAEs gradually exhibit lower L0 as we go deeper in the model. This is in contrast to the base model SAEs which roughly have similar L0 regardless of depth. This hints at potential ways in which fine-tuned models may process information internally in fundamentally different ways than base models. While a full deep dive into the representational dynamics of these models is beyond the scope of this work, we nonetheless think this is a potentially promising direction for future work. 

\section{Training AI Detectors}
\label{app:training_details}
\subsection{Fine-tuning Neural Baseline Models}
For our Neural baseline we fine-tune gemma-2-2b\footnote{\url{https://huggingface.co/google/gemma-2-2b}} on the downstream AI detection task. In Table \ref{tab:ft_hparams} we list the hyperparameters and settings used for all fine-tuning runs. We fine-tune one model each on the following three conditions:
\begin{enumerate}[itemsep=0.15em, topsep=0.6em]
    \item \textbf{COLING}: Human vs. AI generated open-domain text from the COLING dataset \cite{wang-etal-2025-genai}
    \item \textbf{PRDB (GPT-4o)}: Human vs. GPT-4o generated Peer Reviews from the PRDB dataset \cite{yu2026paperreviewedllmbenchmarking}
    \item \textbf{PRDB (Llama)}: Human vs. Llama generated Peer Reviews from the PRDB dataset \cite{yu2026paperreviewedllmbenchmarking}
\end{enumerate}

For each setting we replace the language modeling head with a classification head using \texttt{AutoModelForSequenceClassification} with \texttt{num\_labels=2} and minimize downstream cross entropy loss using the Huggingface \texttt{Trainer} interface \cite{wolf-etal-2020-transformers}. We use the official development set as validation data for each setting and retain the model at the end of training that had the lowest validation loss. We report evaluation metrics in the main paper. In Table \ref{tab:neural_baselines_coling} we confirm that our trained model surpasses the performance of the baseline RoBERTa classifier provided by the shared task organizers.

\paragraph{Hardware \& Training Time} For training we used one NVIDIA RTX A6000 GPU for a total of 80 GPU hours.

\subsection{Training Logistic Regression Classifiers}
For both the SAE model classifiers and the linguistic feature baseline, we train sparse logistic regression classifiers in PyTorch \cite{paszke2019pytorch}, minimizing binary cross-entropy with minibatch SGD with momentum. The L1 penalty is implemented as a soft-thresholding update applied to the weight vector after each optimizer step. 
\begin{equation}
  w \leftarrow \mathrm{sign}(w) \cdot \max(|w| - \eta\lambda_1, 0)  
\end{equation}
which produces exact zeros in the learned weights rather than the near-zero weights produced by including the L1 term in the loss directly \cite{parikh2014}. We use SGD rather than Adam because, in preliminary experiments, sparsity under Adam+L1 plateaued at a high floor regardless of $\lambda_1$. We report our hyperparameters in Table \ref{tab:logreg_hparams}.

Because raw SAE max activations span several orders of magnitude, we normalize each feature by its root-mean-square (RMS) over the training corpus before training. Scaling without centering keeps zero activations at zero, preserving sparsity, and means the L1 penalty acts in standardized feature space. For our gram2vec baseline, whose features are dense rates rather than sparse activations, we instead fully standardize via z-scoring ($(x - \mu)/\sigma$). In both cases the normalization is folded back into the saved weights and bias after training, so the resulting detectors operate directly on raw features.

To cut down on inference time, we cached max pooled activations for each SAE for each document in the training data as memory-mapped arrays on disk, gathering only the rows needed for each minibatch. This allowed us to skip having to do any forward passes of language models during training, allowing us to efficiently train on CPU.

\subsection{Training XGBoost Classifiers}
For both the SAE model classifiers and the linguistic feature baseline, we also train lightweight XGBoost classifiers \cite{chen2016xgboost} using the \texttt{xgboost} Pypi package.\footnote{\url{https://pypi.org/project/xgboost/}} We report our hyperparameters in Table \ref{tab:xgboost_hparams}.

We train these models as a comparison point for the logistic regression, to see whether or not our features have large interaction terms that the logistic regression is not taking advantage of. In Table \ref{tab:head_comparison_coling} we show that XGBoost recovers most of the remaining gap between our FT-Model+SAE condition and the base model, meaning that in the high false positive rate regime, the feature interaction terms are decisive. However, in general we note that our logistic regression classifiers do about as well as XGBoost, indicating that our features are primarily additive with minimal interaction terms. On the other hand, the gram2vec features have very strong interaction terms and benefit substantially from having XGBoost applied, thus suggesting the weakness of the initial feature basis.

\begin{table}[t]
\centering
\setlength{\tabcolsep}{4pt}
\begin{tabular}{lcc}
\toprule
& \textbf{SAE} & \textbf{gram2vec} \\
\midrule
Optimizer & \multicolumn{2}{c}{SGD + Momentum} \\
Momentum & $0.9$ & $0.9$ \\
Learning rate & $3{\times}10^{-2}$ & $3{\times}10^{-2}$ \\
L1 penalty ($\lambda_1$) & $10^{-1}$ & $10^{-2}$ \\
Epochs & $3$ & $3$ \\
Batch size & $1024$ & $1024$ \\
Normalization & RMS & z-score \\
\bottomrule
\end{tabular}
\caption{Hyperparameters used for training the logistic regression classifiers for the SAE detectors and the gram2vec linguistic feature baselines.}
\label{tab:logreg_hparams}
\end{table}

\begin{figure*}[t]
\centering
\includegraphics[width=\textwidth]{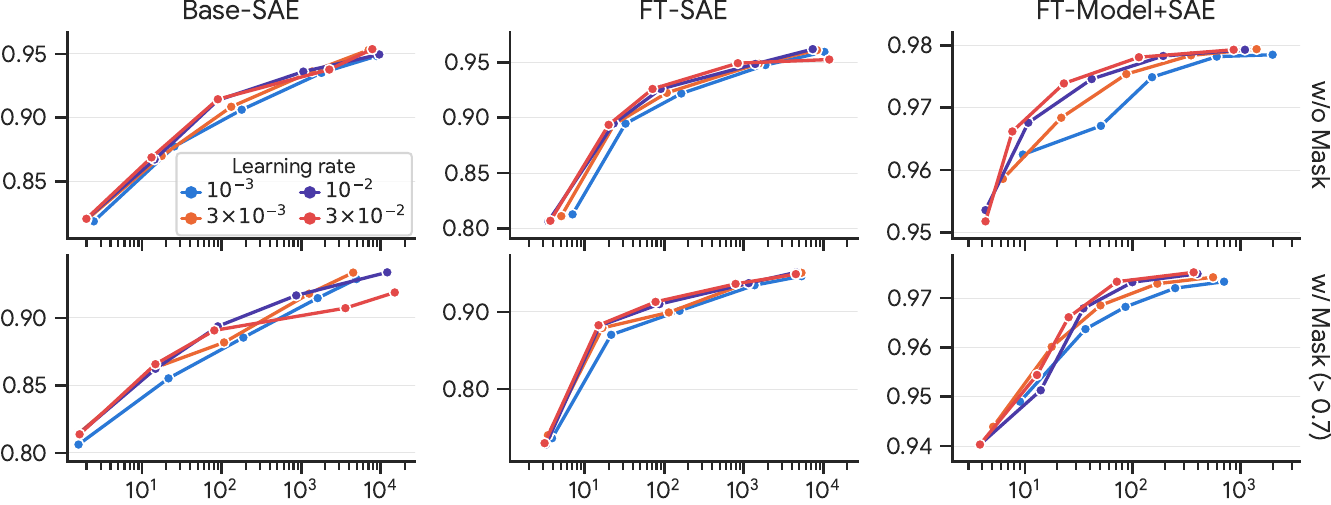}
\caption{COLING-dev F1 against the mean number of active features per document
(log scale) for the SGD~+~$L_1$ grid, three epochs. Columns are the feature
settings, rows the mask thresholds; each line follows one learning rate through the
five $\lambda$ values (leftmost point $\lambda=1$, rightmost $\lambda=10^{-2}$).}
\label{fig:l1_grid_search}
\end{figure*}

\paragraph{Choosing the $L_1$ penalty.}
The strength of the $L_1$ penalty sets how many SAE features a head uses, and the
interpretability analyses in the main paper depend on that number being small.
Following \citet{movva_sparse_2025}, who tune the penalty of an $L_1$-regularised
linear model to select a target number of SAE neurons, we did not pick $\lambda$ by
held-out accuracy alone but by the number of features that carry non-zero weight in
a typical document: we aimed for 30--100 active features per document, few enough
that the features behind a single prediction can be listed and read, and then took
the best COLING-dev F1 inside that band. We swept the learning rate
$\eta \in \{10^{-3}, 3\times10^{-3}, 10^{-2}, 3\times10^{-2}\}$ and $L_1$ penalty
$\lambda \in \{10^{-2}, 3\times10^{-2}, 10^{-1}, 3\times10^{-1}, 1\}$ for every
feature setting and both mask thresholds;
Figure~\ref{fig:l1_grid_search} traces each learning rate's $\lambda$ path on COLING
dev. We choose the values $\eta=3\times10^{-2}$, $\lambda=10^{-1}$ and three epochs
for every head in the paper.

\begin{table}[t]
\centering
\begin{tabular}{lr}
\toprule
\multicolumn{2}{c}{\textbf{XGBoost Hyperparameters}}\\
\midrule
\texttt{objective} & \texttt{reg:squarederror} \\
\texttt{n\_estimators} & 150 \\
\texttt{max\_depth}* & 6 \\
\texttt{subsample}* & 1 \\
\texttt{reg\_lambda}* & 1 \\
\texttt{learning\_rate} & 0.1 \\
\texttt{tree\_method} & \texttt{hist} \\
\bottomrule
\end{tabular}
\caption{Hyperparameters used for fine-tuning the XGBoost classifiers for both the linguistic feature baselines and the SAE classifiers.}
\label{tab:xgboost_hparams}
\end{table}

\begin{table}[t]
\centering
\setlength{\tabcolsep}{3.5pt}
\begin{tabular}{lccc}
\toprule
& \texttt{dev} & \texttt{devtest} & \texttt{test} \\
\midrule
RoBERTa-baseline & 96.2 & 83.1 & 74.9 \\
gemma-2-2b & \textbf{98.0} & \textbf{86.8} & \textbf{75.2} \\
\bottomrule
\end{tabular}
\caption{F1 score of the RoBERTa baseline provided by the COLING shared task organizers against our fine-tuned gemma-2-2b neural detector (\S\ref{sec:baselines}).}
\label{tab:neural_baselines_coling}
\end{table}

\section{Data}
\subsection{Processing}
\label{app:processing}
The PRDB dataset \cite{yu2026paperreviewedllmbenchmarking} consists of human and AI generated peer review text from NeurIPS and ICLR for the previous 8 years (2016-2024). During that time period, the structure of peer reviews has changed. In the early years, reviewers were often given just a single text box to write their review. However, recently conferences have been breaking up the review into different sections (e.g. ``Summary of Contributions'', ``Strengths and Weaknesses'', ``Questions''). PRDB respects these sections and separates the text of reviews into different columns based on the section they come from. This requires us to decide on a standardized format to use when converting a peer review with multiple text fields into a pure text document.

For every document that has multiple fields, we take the name of the field (e.g. \texttt{summary\_of\_the\_paper}), remove the underscores and convert it to title case. We then insert double asterisks to the left and right to simulate a bold Markdown header and add the raw text of each section directly afterwards. This is accomplished in a very quick and simple one liner \texttt{f"**\{title\}**\textbackslash n\{value.strip()\}\textbackslash n\textbackslash n"}.

Given that both human written and AI generated peer reviews are required to conform to this structure, adding these Markdown elements does not add any formatting artifacts that would be easily exploitable by our detectors. On the contrary, this artificially reduces the predictive power of any markdown-based features. We do this intentionally to bias our models towards using more substantive features for detection.

\subsection{Generator Models and Domains}
\label{app:stats}
In Table \ref{tab:datasets} we report the extended lists of the generator models and domains present in each dataset. We note that there are substantial overlaps in data coverage, ArXiv appears many times between training and testing, along with Wikipedia and QA. While this is the standard dataset for training AI detectors in the literature, it is worth noting that transfer to these collections may not be a perfect representation of true OOD performance of detectors.

\begin{table}[t]
\centering
\small
\setlength{\tabcolsep}{4pt}
\begin{tabular}{ccllc}
\toprule
\textbf{Bin} & \textbf{Auto-interp $\tau$} & \textbf{\# Feats} & \textbf{\# Pairs} & \textbf{$\rho$} \\
\midrule
0 & $[\,0.700,\ 0.707\,]$ & 44 & 14{,}733 & 0.778 \\
1 & $[\,0.707,\ 0.719\,]$ & 20 &  6{,}162 & 0.791 \\
2 & $[\,0.719,\, 0.750\,]$ & 31 & 10{,}276 & 0.825 \\
3 & $[\,0.750,\, 0.866\,]$ & 29 & 10{,}169 & 0.871 \\
\midrule
All & $[\,0.700,\, 0.866\,]$ & 124 & 41{,}340 & 0.809 \\
\bottomrule
\end{tabular}
\caption{Agreement between zero-ablation and paraphrase logit deltas as a function of auto-interpretation score (\textsc{FT-Model+SAE}, $\tau\!>\!0.7$, PRDB, GPT-4o target). Features are binned by quartiles of their auto-interp score; within each bin, Spearman $\rho$ is computed between the per-feature mean zero-ablation $\Delta$logit and the per-feature mean paraphrase $\Delta$logit. We see that correlation between paraphrase ablation and zero-ablation ablation score goes up with increasing autointerp score.}
\label{tab:cf_autointerp_bins}
\end{table}

\begin{table}[t]
\centering
\begin{tabular}{lr}
\toprule
\multicolumn{2}{c}{\textbf{Neural AI Detector Hyperparamters}}\\
\midrule
\texttt{optimizer}* & \texttt{adamw\_torch}\\
$\beta_1, \beta_2, \epsilon$* & $0.9, 0.999, 10^{-8}$\\
\texttt{precision}* & \texttt{float32}\\
\texttt{lr\_scheduler\_type}* & \texttt{linear}\\
\texttt{gradient\_accum\_steps}*&$1$\\
\texttt{max\_grad\_norm}* & $1.0$\\
\texttt{num\_epochs} & $1$\\
\texttt{batch\_size} & $4$\\
\texttt{warmup\_steps}& $500$\\
\texttt{weight\_decay}&$0.01$\\
\texttt{learning\_rate}&$5\times 10^{-5}$\\
\texttt{load\_best\_model\_at\_end}&\texttt{True}\\
\texttt{metric\_for\_best\_model}&\texttt{eval\_loss}\\
\bottomrule
\end{tabular}
\caption{Hyperparameters used for fine-tuning Gemma-2-2B model on AI detection. Asterisks indicate defaults inherited from Huggingface Trainer.}
\label{tab:ft_hparams}
\end{table}

\section{Autointerpretation Pipeline Prompts}
\label{app:autointerp}
In Figure \ref{fig:generation-prompt} and Figure \ref{fig:validation-prompt} we report the full text of the prompts used in the description generation and validation portions of our pipeline respectively. During generation we ask the model to provide three different flavors of description: one terse, one keyword-based, and one long. For all experiments and validation in our work we always use the terse description and ignore the keyword and long variants. The latter are only used when visualizing features in our interface (\S\ref{app:interface}).

In Figure \ref{fig:feature-examples} we show an example of how documents are formatted in our description generation prompt. We provide the language model with an extended list of documents from across the feature's activation distribution. For each document we include special angled bracket tokens to indicate on which tokens a feature is most active and crop each document to a small window ($\pm$20 tokens) around the max activating token. During description validation we follow the same process but group all documents together in the same bulleted list. For negative documents, during validation, we randomly sample a token as the maximum activating token and highlight and crop as necessary.

\begin{figure*}
\centering
\includegraphics[width=\textwidth]{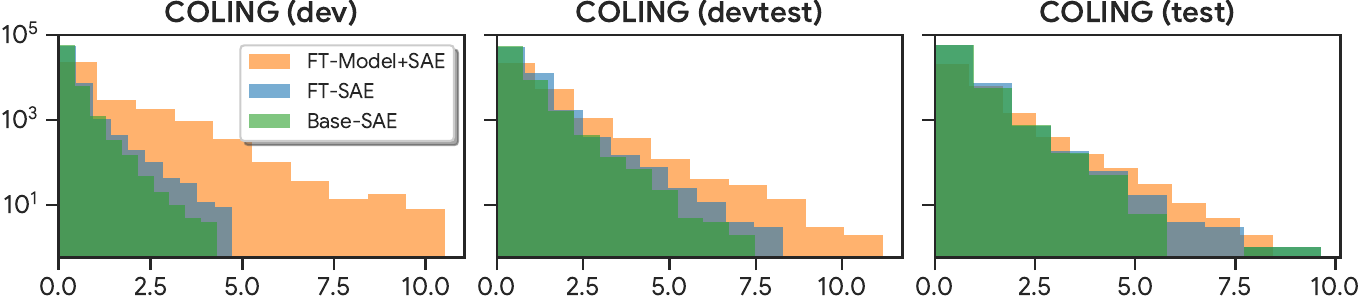}
\caption{Histogram of Log Odds Ratio (x-axis) of all features from the Open Domain AI generated text detectors. We see that \textsc{FT-Model+SAE} produces far more highly predictive features on the in-distribution data but once out-of-distribution the features are no more predictive than those from \textsc{FT+SAE} or \textsc{Base-SAE}.}
\label{fig:lor_hist}
\end{figure*}

\begin{table*}[t]
\centering
\small
\setlength{\tabcolsep}{4pt}
\begin{tabular}{ll rrr rrr rrr}
\toprule
& & \multicolumn{3}{c}{\textbf{Interpretable}} & \multicolumn{3}{c}{\textbf{Interp.\ \& $w>0$}} & \multicolumn{3}{c}{\textbf{Avg.\ \# active / doc}} \\
\cmidrule(lr){3-5} \cmidrule(lr){6-8} \cmidrule(lr){9-11}
\textbf{Data} & \textbf{SAE} & $\tau\!>\!0.0$ & $\tau\!>\!0.7$ & $\tau\!>\!0.85$ & $\tau\!>\!0.0$ & $\tau\!>\!0.7$ & $\tau\!>\!0.85$ & $\tau\!>\!0.0$ & $\tau\!>\!0.7$ & $\tau\!>\!0.85$ \\
\midrule
COLING & \textsc{Base-SAE} & 65{,}512 & 33{,}624 & 9{,}355 & 201 & 189 & 149 & 101 & 107 & 108 \\
COLING & \textsc{FT-SAE} & 66{,}245 & 43{,}480 & 13{,}051 & 118 & 136 & 155 & 100 & 101 & 101 \\
COLING & \textsc{FT-Model+SAE} & 29{,}489 & 16{,}893 & 2{,}983 & 63 & 65 & 65 & 53.7 & 41.9 & 25.6 \\
\midrule
PRDB & \textsc{Base-SAE} & 65{,}153 & 22{,}332 & 5{,}544 & 82 & 156 & 71 & 79.6 & 116 & 68.7 \\
PRDB & \textsc{FT-SAE} & 64{,}997 & 39{,}231 & 7{,}967 & 71 & 85 & 115 & 99.3 & 83.7 & 85.1 \\
PRDB & \textsc{FT-Model+SAE} & 17{,}518 & 11{,}451 & 1{,}903 & 111 & 71 & 52 & 78.6 & 49.7 & 33.6 \\
\bottomrule
\end{tabular}
\caption{SAE feature counts summed over the four probed layers. \textbf{Interpretable}: features whose auto-interp score exceeds $\tau$. \textbf{Avg.\ \# active / doc}: mean number of nonzero-weight interpretable features that fire on a held-out test document, i.e.\ that contribute a nonzero logit term to the prediction. PRDB rows use the GPT-4o target.}
\label{tab:active_feature_counts}
\end{table*}

\section{Paraphrase Ablation Details}
\label{app:paraphrase-ablation}

Both ablations of \S\ref{sec:causal_ablation} use \texttt{gemini-3.1-flash-lite} (temperature 1.0, top\_p$=0.95$, top\_k$=64$, one sample per prompt). The model returns the rewrite inside \texttt{<rewrite>} tags, from which we extract the paraphrase; empty responses (e.g., safety-filtered) are discarded and counted.

\subsection{Feature-Level Paraphrase Ablation}

\paragraph{Prompt.}
Figure~\ref{fig:paraphrase-prompt} gives the full prompt. It includes all three components of the feature description (terse summary, keywords, detailed description) and instructs the model to change only the target feature.

\paragraph{Document Subsampling.}
For each dataset we draw a class-balanced sample of 1{,}000 documents (500 human, 500 AI) from the held-out test split with a fixed seed. On PRDB the AI class is the detector's target generator (GPT-4o); on COLING, where detectors are trained on all generators pooled, AI documents are sampled from every generator in proportion to its share of the corpus.

\paragraph{Feature Sampling.}
For each document we compute the zero-ablation logit delta of every active feature receiving weight in the head (in the w/ Mask condition only features with simulation score above $0.7$ can be active) and uniformly sample up to 50 of them, without replacement, as paraphrase targets. Sampling is restricted to \emph{describable} features, i.e., those with an auto-interp description and a defined simulation score; this excludes features too rare for auto-interp to run and a small set of document-start artifacts whose descriptions cannot guide an edit. Sampling is seeded per document. This yields roughly 40{,}000--50{,}000 (document, feature) rewrite records per configuration.

\paragraph{Measurements.}
Logit deltas are measured in log-odds space with a consistent sign convention. The activation shifts $\Delta_{\text{tgt}}$ and $\Delta_{\text{non}}$ are the absolute change in a feature's activation between original and rewrite, normalized by that feature's mean activation over the training corpus (so $1.0$ is a shift equal to its typical activation); $\Delta_{\text{tgt}}$ is this quantity for the targeted feature and $\Delta_{\text{non}}$ its mean over all other active features. Per-feature analyses exclude features with $|\text{Ablate }\Delta| < 0.01$ or sampled in fewer than 10 documents.

\begin{figure*}[t]
\centering
\includegraphics[width=\textwidth]{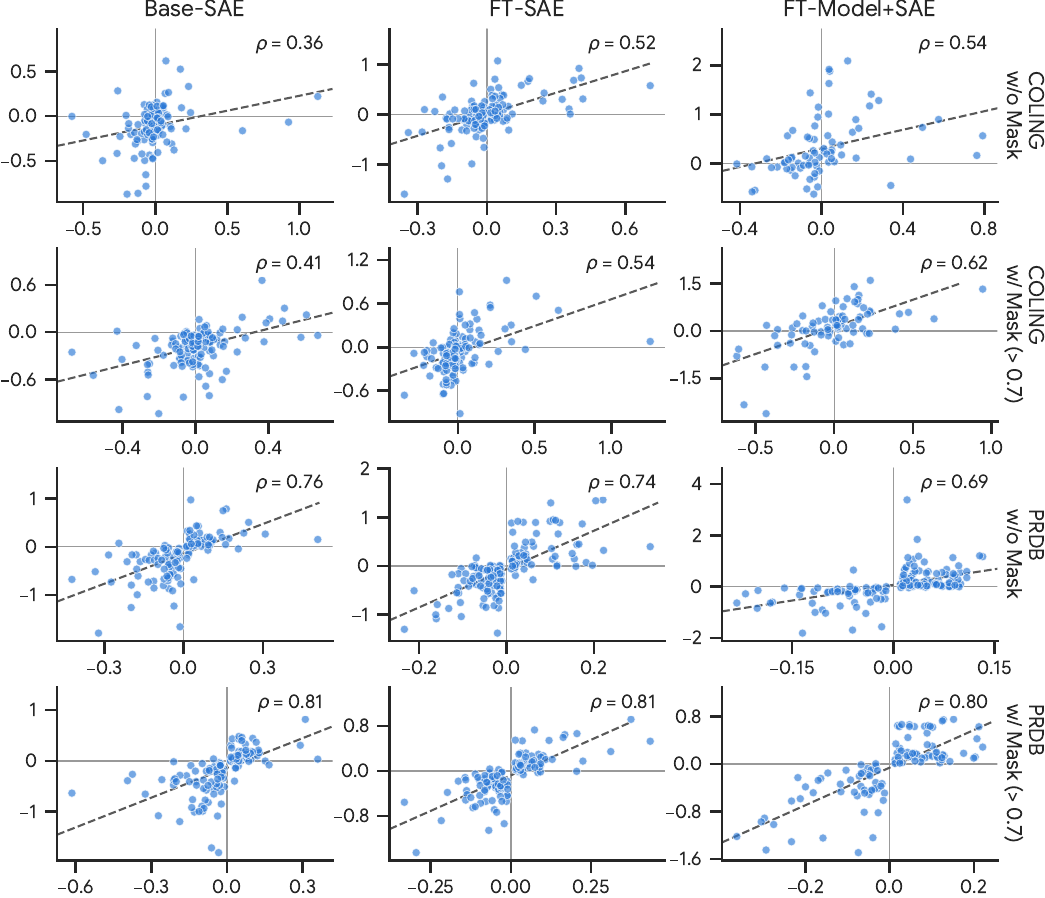}
\caption{Per-feature agreement between the two interventions
(\S\ref{sec:causal_ablation}), for the logistic-regression heads. Each point is one
SAE feature. The $x$-axis is the feature's mean change in the detector's AI logit
when the feature is zero-ablated, and the $y$-axis is the mean change when the text
is instead paraphrased to remove the feature, both averaged over the documents in
which the feature was sampled. Positive values on both axes mean the feature pushed
the prediction toward AI and removing it from the text lowered the AI logit.
Columns are the feature settings; the top two rows are COLING without and with the
interpretability mask ($\tau > 0.7$), the bottom two rows the same for PRDB (GPT-4o).}
\label{fig:cf_feature_agreement}
\end{figure*}

\paragraph{Extended Results.}
Figure~\ref{fig:cf_feature_agreement} plots each feature's zero-ablation delta against its paraphrase delta for every head: agreement is moderate on COLING ($\rho = 0.36$--$0.62$), strong on PRDB ($0.69$--$0.81$), and higher with the mask in every setting. Table~\ref{tab:cf_autointerp_bins} bins features by auto-interp score; $\rho$ rises from $0.78$ in the lowest quartile to $0.87$ in the highest, so better-described features are edited away more precisely. Tables~\ref{tab:cf_frequent_prdb} and~\ref{tab:cf_frequent_coling} list the most frequently targeted features. On PRDB rewrites are precise ($\Delta_{\text{tgt}}$ typically $0.5$--$1.0$, $\Delta_{\text{non}}$ $0.02$--$0.3$); on COLING the base-model SAEs' most common features are punctuation and sentence-boundary detectors that a meaning-preserving rewrite can rarely remove, which accounts for the weaker agreement.

\subsection{Document-Level Paraphrase Ablation}

\paragraph{Setup.}
We attack the $\tau > 0.85$ logistic-regression heads of all three feature settings on COLING and PRDB (Llama-trained heads in Figure~\ref{fig:adv_flip}; GPT-4o-trained heads as a robustness check). Each detector uses the Youden threshold of its development-set evaluation, fixed before the attack; we verified that this reproduces its reported F1. Per dataset we draw 2{,}500 AI and 2{,}500 human documents with a fixed seed (30--900 Gemma tokens, so paraphrases fit the 1{,}024-token extractor window), shared by the dataset's three detectors; PRDB AI documents come from the target generator only. Only AI documents are paraphrased, so false positives are held fixed and F1 moves only through recall.

\paragraph{Conditions.}
The three unguided prompts (Figures~\ref{fig:adv-prompt-baseline}--\ref{fig:adv-prompt-guided}) are generated once per dataset and scored by all of its detectors; they share an edit budget of ``well under 20\%'' of the words. \emph{Few Shot} draws three human training-split documents per target from the same sub-source (COLING) or conference (PRDB), length-matched and, for PRDB, from papers absent from the test split. \emph{Guided} adds six generic rules against machine-writing habits written from general knowledge, independent of any detector. The \emph{SAE} prompt (Figures~\ref{fig:adv-prompt-sae}, \ref{fig:adv-prompt-sae-evidence}) is detector-specific: it lists the ten largest AI-direction and ten largest human-direction feature contributions with their descriptions, keywords, signed contributions and the three passages where each fires most strongly, states the max-pooling counting rule, sets a target equal to the score's distance above the threshold plus 3, and asks for a plan before the rewrite. \emph{SAE $\times 3$} repeats this up to three times, recomputing the evidence on each rewrite and re-prompting only documents still classified as AI; a document is frozen at the first pass that flips. Failed generations were rare (0--6 per condition) and are excluded.

\paragraph{Metrics.}
F1 at the frozen threshold with human scores fixed; attack success rate (ASR), the share of base-detected AI documents classified human after the attack; and Lev, the word-level Levenshtein distance to the original as a share of the longer text. For SAE $\times 3$ all quantities are measured on the text at which a document was frozen, so Lev is cumulative. Uncertainty is a 1{,}000-sample bootstrap over AI documents. Table~\ref{tab:adv_appendix_all} reports all metrics for every detector and condition.

\paragraph{Extended Results.}
The feature-guided attack is also the most economical: one SAE pass edits 6--29\% of words and three passes 11--38\%, against 24--39\% for the unguided prompts, while flipping two to ten times as many documents. The one exception is \textsc{FT-SAE} on PRDB (ASR 3.4\% after one pass, 21.8\% after three), whose evidence is spread over about 150 densely firing base-model features that no ten-feature list captures. The GPT-4o-trained PRDB heads behave like the Llama-trained ones (ASR after three passes 69\%, 20\%, 83\% vs.\ 60\%, 22\%, 97\%), so the conclusions do not depend on the generator. Because the attack only needs to cross the frozen threshold, AUROC degrades far less than F1 (e.g., 0.96 to 0.82 for \textsc{FT-Model+SAE} on COLING).

\section{Interface}
\label{app:interface}
In Figure \ref{fig:interface} we show a screenshot of the interface we built for our auditable detectors. The main window accepts raw text input and is where users would copy-paste in their text to be run through the detector. Once the user hits enter, we run the base model, extract SAE activations from the four hook points, and run the classification head to get the final prediction.

On the right we visualize the model's prediction as well as a break down of the top most influential features for the model's decision towards both AI and Human text. Each feature card includes the description generated by our auto-interpretation pipeline and it's associated score (\S\ref{sec:autointerp}), the net logit contribution of the feature, and the number of instances of that feature appearing on both the human and AI side of our training data.

For each feature, we localize it in the main text by highlight the tokens on which the feature fires. We shade tokens in proportion to how strongly the feature is firing. Hovering over a feature card will temporarily remove highlights from all other features to allow for individual feature inspection. Hovering over highlighted tokens will reveal a popout window with a more detailed description of the feature and a list of common keywords. We allow users to dynamically modify the target $\tau$ they would like to run inference on to help control the desired accuracy-interpretability tradeoff.

\begin{figure}[t]
\centering
\includegraphics[width=\columnwidth]{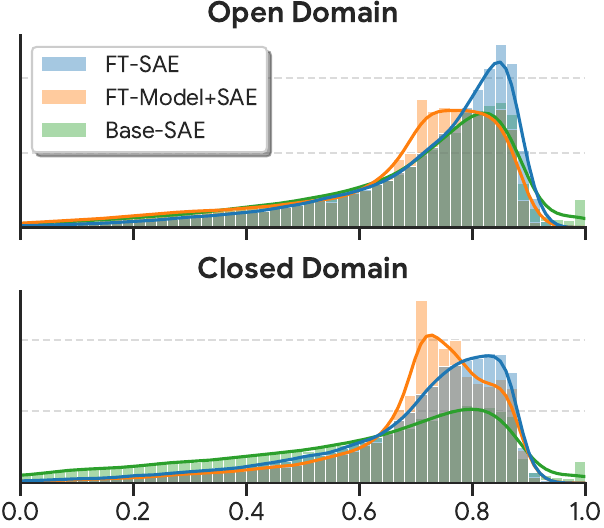}
\caption{Density normalized histograms of feature auto-interp correlation scores (x-axis) across our conditions. We can see that in both cases, the \textsc{FT-SAE} model retains a higher proportion of highly interpretable features.}
\label{fig:corr}
\end{figure}

\section{Extra Tables and Figures}
\label{app:extra}
\begin{figure*}
\centering
\includegraphics[width=\textwidth]{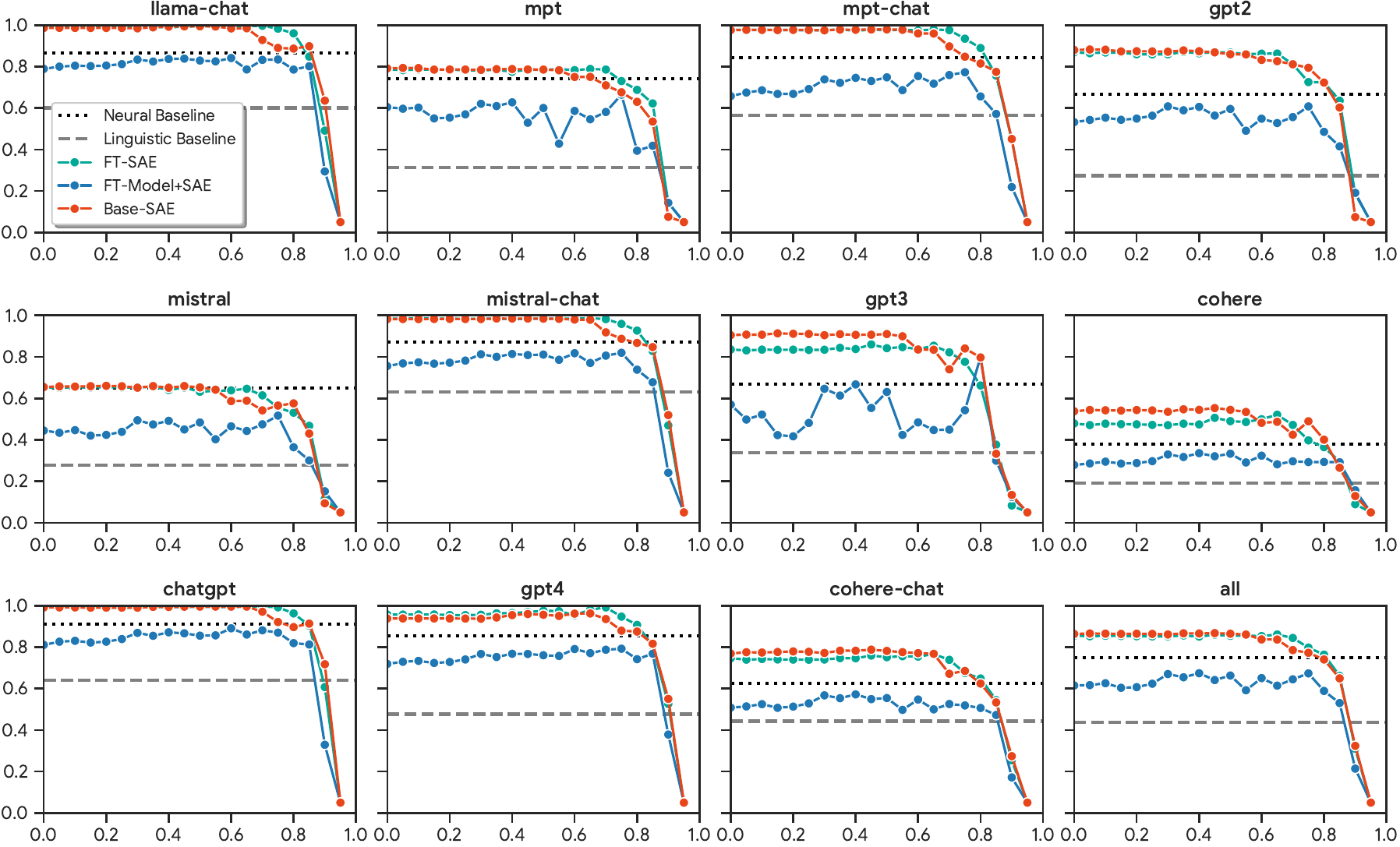}
\caption{Recall at FPR=5\% (y-axis) vs. Interpretability Threshold (x-axis) for $L_1$-regularized logistic regression classifiers trained on the COLING dataset evaluated on various splits of the RAID dataset \cite{dugan-etal-2024-raid}.}
\label{fig:raid_full_results}
\end{figure*}

In Figure \ref{fig:raid_full_results} we report the AUROC scores for our different conditions for various interpretability thresholds on the RAID dataset. We observe that the \textsc{FT-Model+SAE} condition seems to roughly match the performance of the full neural baseline for most conditions and that both the \textsc{Base-SAE} and \textsc{FT-SAE} perform better on models present in the COLING dataset.

In Table \ref{tab:prdb_results_full} we report evaluation results on the PRDB dataset. We see that training on Llama generally leads to worse performance on the three held out models (Qwen, Claude, and Gemini). We suspect that this may be due to Llama sharing fewer specific stylistic tendencies with other comparable models. Previous work by \citet{jiang2025artificialhivemindopenendedhomogeneity} has noted the surprising degree of stylistic similarity across  different language models. We hope that these types of SAE-based feature analyses may give us a more precise vocabulary for describing the subtle differences between models.

In Tables \ref{tab:top_features_coling_ftsae}, \ref{tab:top_features_coling_ftmodelsae}, and \ref{tab:top_features_coling_basesae} we report the most predictive features found by each of our model conditions on the COLING dataset, and in Tables \ref{tab:top_features_prdb_basesae_gpt4o} through \ref{tab:top_features_prdb_ftsae_llama} we report the most predictive features found on the PRDB dataset. Both tables are ranked by mean counterfactual logit change (Logit $\Delta$) in other words, the average importance score of a feature.

Tables \ref{tab:worked_example_gpt-4o-batch-ICLR2019-1259-r1g5pn1527} through \ref{tab:worked_example_llama-3.1-70b-ICLR2021-1934-6POeD7QDvXS} show full example outputs of our pipeline on both COLING and PRDB.

\begin{table*}[t]
\centering
\small
\setlength{\tabcolsep}{4pt}
\renewcommand{\arraystretch}{1.2}
\begin{tabular}{@{}lp{8.5cm}rrrrr@{}}
\toprule
\multicolumn{7}{@{}c}{\textbf{Feature-Level Paraphrase Results $\tau > 0.7$ --- Paper Reviews}}\\
\midrule
\textbf{Feature} & \textbf{Concept Description} & \textbf{\# Docs} & \textbf{Ablate $\Delta$} & \textbf{Para.\ $\Delta$} & $\boldsymbol{\Delta_{\text{tgt}}}$ & $\boldsymbol{\Delta_{\text{non}}}$\\
\midrule
\multicolumn{7}{@{}l}{\textbf{\textsc{Base-SAE} (GPT-4o)}}\\
7\_16119 & words related to claims assertions and disclaimers & 495 & $+0.0139$ & $+0.0382$ & 0.70 & 0.02 \\
13\_6062 & period punctuation mark at the end of sentences & 434 & $-0.0690$ & $-0.151$ & 0.67 & 0.09 \\
13\_377 & markdown formatting for section breaks headers or newlines in academic reviews & 424 & $-0.0495$ & $-0.109$ & 0.22 & 0.06 \\
22\_632 & subject pronoun it followed by a verb phrase & 386 & $-0.0189$ & $-0.230$ & 0.48 & 0.06 \\
22\_14264 & words potential and potentially & 384 & $+0.0158$ & $+0.0733$ & 0.54 & 0.02 \\
22\_15232 & starts of sentences or paragraphs, specifically the word this & 382 & $-0.0703$ & $-0.188$ & 0.90 & 0.05 \\
\midrule
\multicolumn{7}{@{}l}{\textbf{\textsc{FT-SAE} (GPT-4o)}}\\
7\_9550 & period character used as end of sentence punctuation & 464 & $+0.0467$ & $+0.319$ & 0.37 & 0.14 \\
22\_8261 & forms of the verb to be such as is and are & 450 & $-0.0283$ & $-0.538$ & 0.39 & 0.19 \\
22\_10248 & punctuation marks like periods commas colons and potentially punctuation attached to preceding tokens & 442 & $-0.0447$ & $-0.733$ & 0.61 & 0.32 \\
22\_6121 & word paper often followed by a verb in academic reviews & 427 & $+0.0172$ & $+0.0451$ & 0.29 & 0.07 \\
13\_4198 & positive feedback or evaluation terms in academic peer reviews & 422 & $-0.111$ & $-0.0545$ & 0.44 & 0.17 \\
7\_11925 & activates on commas used within sentences or parenthetical clauses & 419 & $-0.164$ & $-0.181$ & 0.95 & 0.17 \\
\midrule
\multicolumn{7}{@{}l}{\textbf{\textsc{FT-Model+SAE} (GPT-4o)}}\\
7\_9670 & initial sentence capitalization at the beginning of a paragraph or after a period & 554 & $-0.0184$ & $-0.330$ & 0.95 & 0.22 \\
7\_9027 & the word is in questions or declarations often following a subject or noun phrase & 522 & $-0.298$ & $-0.911$ & 0.94 & 0.18 \\
7\_11114 & first person singular pronouns referring to the author or reviewer & 491 & $-0.0459$ & $-0.457$ & 0.85 & 0.27 \\
7\_11298 & period punctuation mark at the end of sentences or list items & 481 & $-0.170$ & $-0.275$ & 0.92 & 0.13 \\
22\_13620 & sentence initial transition words & 471 & $-0.0932$ & $-0.269$ & 0.53 & 0.13 \\
13\_9322 & activates on the word paper when it appears in academic reviews or research contexts & 460 & $-0.219$ & $-0.229$ & 1.04 & 0.04 \\
\bottomrule
\end{tabular}
\caption{The 6 features most often sampled for paraphrase by each of the \textsc{Base-SAE}, \textsc{FT-SAE} and \textsc{FT-Model+SAE} ($\tau > 0.7$) counterfactual experiment on PRDB, ranked by the number of documents in which they were targeted (descriptions from gemini-3.1-flash-lite auto-interp). Each row reports the feature's mean change in detector logit under zero-ablation (Ablate $\Delta$) and under Gemini's rewrite (Para.\ $\Delta$), and the mean normalized change in activation that the rewrite induced in the targeted feature ($\Delta_{\text{tgt}}$; 1.0 is complete removal) and in all other active features ($\Delta_{\text{non}}$).}
\label{tab:cf_frequent_prdb}
\end{table*}

\begin{table*}[t]
\centering
\small
\setlength{\tabcolsep}{6pt}
\renewcommand{\arraystretch}{1.2}
\begin{tabular}{@{}lp{7.5cm}rrrrr@{}}
\toprule
\multicolumn{7}{@{}c}{\textbf{Feature-Level Paraphrase Results $\tau > 0.7$ --- General Domain Text}}\\
\midrule
\textbf{Feature} & \textbf{Concept Description} & \textbf{\# Docs} & \textbf{Ablate $\Delta$} & \textbf{Para.\ $\Delta$} & $\boldsymbol{\Delta_{\text{tgt}}}$ & $\boldsymbol{\Delta_{\text{non}}}$\\
\midrule
\multicolumn{7}{@{}l}{\textbf{\textsc{Base-SAE}}}\\
22\_1866 & introductory phrases or transitional clauses introducing topics or evidence & 523 & $+0.148$ & $+0.0391$ & 0.09 & 0.26 \\
17\_9942 & punctuation marks like periods and apostrophes & 519 & $+0.374$ & $-0.122$ & 0.01 & 0.22 \\
17\_15019 & opening quotation marks followed by quoted dialogue or official statements & 422 & $+0.0545$ & $-0.121$ & 0.16 & 0.05 \\
13\_13241 & period punctuation mark at the end of sentences & 407 & $-0.326$ & $-0.169$ & 0.35 & 0.11 \\
7\_14055 & words indicating significance or clarity such as important clear challenging and known & 381 & $+0.0230$ & $-0.0803$ & 0.24 & 0.08 \\
17\_9782 & forms of the verb to be such as is was were are & 380 & $-0.0474$ & $-0.152$ & 0.66 & 0.17 \\
\midrule
\multicolumn{7}{@{}l}{\textbf{\textsc{FT-SAE}}}\\
13\_3849 & period punctuation at the end of sentences & 482 & $-0.0223$ & $+0.0659$ & 0.31 & 0.12 \\
13\_3106 & punctuation marks at the end of sentences such as periods exclamation marks and parentheses & 461 & $-0.0470$ & $+0.104$ & 0.70 & 0.15 \\
22\_8962 & labels or headers appearing at the start of paragraphs or sections often formatted as document metadata & 395 & $-0.0219$ & $-0.0818$ & 0.14 & 0.08 \\
13\_12507 & neuron fires on segments of complex sentences often containing punctuation or mid-sentence conjunctions & 384 & $+0.105$ & $-0.215$ & 0.20 & 0.30 \\
22\_3589 & subjective value judgment descriptors and evaluative adjectives often appearing at sentence starts & 371 & $-0.0669$ & $-0.0767$ & 0.20 & 0.16 \\
13\_5570 & words and short phrases following the verb are or similar auxiliary verbs in instructional or descriptive contexts & 369 & $-0.0852$ & $-0.171$ & 0.20 & 0.13 \\
\midrule
\multicolumn{7}{@{}l}{\textbf{\textsc{FT-Model+SAE}}}\\
13\_2232 & noun phrases or verb phrases involving pronoun subjects or multi-word descriptive segments & 952 & $+0.0266$ & $+0.344$ & 0.20 & 0.32 \\
17\_12278 & words and phrases that contribute to a descriptive or observational narrative style & 812 & $+0.212$ & $+0.535$ & 0.28 & 0.38 \\
22\_13711 & words indicating actions of resolving problems or achieving tasks & 770 & $+0.119$ & $+0.181$ & 0.17 & 0.11 \\
17\_15712 & words and phrases related to common discourse or importance markers & 766 & $+0.634$ & $+0.383$ & 0.18 & 0.25 \\
7\_15524 & words following functional prepositions or within complex syntactic clauses & 764 & $+0.233$ & $+0.588$ & 0.56 & 0.47 \\
7\_8482 & capitalized words starting sentences or proper nouns & 750 & $+0.229$ & $+1.60$ & 0.77 & 0.36 \\
\bottomrule
\end{tabular}
\caption{The 6 features most often sampled for paraphrase by each of the \textsc{Base-SAE}, \textsc{FT-SAE} and \textsc{FT-Model+SAE} ($\tau > 0.7$) counterfactual experiment on COLING, ranked by the number of documents in which they were targeted (descriptions from gemini-3.1-flash-lite auto-interp). Each row reports the feature's mean change in detector logit under zero-ablation (Ablate $\Delta$) and under Gemini's rewrite (Para.\ $\Delta$), and the mean normalized change in activation that the rewrite induced in the targeted feature ($\Delta_{\text{tgt}}$; 1.0 is complete removal) and in all other active features ($\Delta_{\text{non}}$).}
\label{tab:cf_frequent_coling}
\end{table*}

\section{AI Assistance}
\label{app:ai_assistance}
AI assistants were used to implement the interface in section \ref{app:interface}, write the initial drafts of the introduction and abstract, and gather initial collections of related work for literature review. We also used AI frequently in smaller less autonomous capacities such as when brainstorming the organization of information in this paper, writing specific self-contained functions, creating tables and visualizations, and adding documentation to our codebase.

We used Gemini 2.5 Pro for everyday questions and Claude Opus 4.7 for larger more agentic tasks. Wherever possible, in our code, we indicate when AI assistants were involved with comments.

\begin{table*}[t]
\centering
\footnotesize
\setlength{\tabcolsep}{4pt}
\renewcommand{\arraystretch}{1.05}
\begin{tabular}{@{}p{7.3cm}p{5.5cm}l@{}}
\toprule
\textbf{Input text} & \textbf{Active features} (Logit $\Delta$) & \textbf{Score}\\
\midrule
\textit{(GPT-4o)}~The paper titled ``Sub\hlai[2]{gradient} \hlai[2]{Descent}\fk{5} \hlai[2]{Lear}\fk{5,e}\hlai[3]{ns} \hlai[3]{Orthogonal} \hlai[3]{Dictionaries}\fk{5}'' \hlai[3]{presents}\fk{3} a \hlai[1]{significant} \hlai[1]{contribution} \hlai[1]{to}\fk{3} the field of dictionary learning, \hlai[3]{particularly}\fk{4} in addressing the challenge of learning orthogonal dictionaries through a novel approach involving subgradient descent. The authors successfully demonstrate that a subgradient descent algorithm with random initialization can learn orthogonal dictionaries using a nonsmooth, nonconvex \newline $\ell$1 minimization formulation. This approach \hlhu[1]{contrasts}\fk{d,e} with prior methods that necessitate complex initialization strategies or computationally expensive operations. The research marks \hlai[1]{a}\fk{7} noteworthy \hlai[2]{advancement}\fk{7} by providing the \hlhu[1]{first}\fk{c} theoretical guarantee for dictionary recovery via subgradient descent on this specific problem formulation.\newline\mbox{}\newline The paper's originality stems from the novel application of subgradient descent in a nonconvex, nonsmooth optimization landscape, tackling issues directly rather than circumventing them through smoothing techniques. This contributes \hlai[2]{valu}\fk{2,a}\hlai[2]{ably} \hlai[2]{to} \hlai[2]{the}\fk{2} dictionary learning domain \hlai[2]{and} \hlai[3]{offers} \hlai[3]{insights} \hlai[3]{that} \hlai[3]{could} \hlai[2]{extend}\fk{2} to other nonconvex, nonsmooth optimization \hlai[3]{problems}\fk{6}. The use of Riemannian subgradient descent on a nonsmooth formulation is \hlai[3]{particularly}\fk{2,4} \hlai[3]{innovative}\fk{1}, and their theoretical analysis of the optimization landscape dynamics is rigorous.\newline\mbox{}\newline In \hlhu[1]{terms}\fk{b} of scientific rigor, the paper is comprehensive in its theoretical underpinnings.
&
\textit{Toward AI}\newline
\fk{1}\,$+0.27$\; innovative variations like innovative innovations or invasive\newline
\fk{2}\,$+0.22$\; academic phrases predicting future influence or impact on researchers\newline
\fk{3}\,$+0.22$\; academic paper verbs introducing research contributions\newline
\fk{4}\,$+0.19$\; word particular and variations like particularly in academic writing context\newline
\fk{5}\,$+0.16$\; multi-word academic terminology specifically within titles or descriptive naming conventions\newline
\fk{6}\,$+0.16$\; word environments appearing in technical literature regarding system deployment or model applicability\newline
\fk{7}\,$+0.13$\; words related to progress or improvements such as advances advancement advancements or shifts\newline
\textit{+ 15 more}\; $+0.88$\newline
\textit{Toward human}\newline
\fk{a}\,$-0.10$\; typos or misspellings within technical or academic documents\newline
\fk{b}\,$-0.03$\; idiomatic transition phrases like to put it bluntly to be clear and to illustrate this with a\newline
\fk{c}\,$-0.02$\; phrase to the best of our knowledge or similar personal epistemic qualifiers\newline
\fk{d}\,$-0.02$\; comparing similarity or difference between methods or concepts\newline
\fk{e}\,$-0.01$\; first person subjective expressions or comparisons of ideas to previous work
&
\begin{tabular}[t]{@{}lr@{}}
$\Sigma$ AI    & $+2.22$ \\
$\Sigma$ human & $-0.18$ \\
bias           & $-0.51$ \\
\midrule
logit          & $+1.55$ \\
$P(\mathrm{AI})$ & $82\%$ \\
\multicolumn{2}{@{}l}{\textbf{$\rightarrow$ AI}} \\
\end{tabular}
\\
\bottomrule
\end{tabular}
\caption{The \textsc{FT-SAE} ($\tau = 0.85$) logistic-regression detector applied to a GPT-4o-written review from the PRDB test set (\texttt{ICLR2019-1259-r1g5pn1527}). Tokens on which a listed feature fires are shaded red (feature pushes toward AI) or blue (toward human), with three intensities set by the token-level contribution $|w_i a_{i,t}|$ of the strongest listed feature on that token.}
\label{tab:worked_example_gpt-4o-batch-ICLR2019-1259-r1g5pn1527}
\end{table*}

\begin{table*}[t]
\centering
\footnotesize
\setlength{\tabcolsep}{4pt}
\renewcommand{\arraystretch}{1.05}
\begin{tabular}{@{}p{7.3cm}p{5.5cm}l@{}}
\toprule
\textbf{Input text} & \textbf{Active features} (Logit $\Delta$) & \textbf{Score}\\
\midrule
\textit{(Llama3)}~Based on the lab \hlhu[1]{values}\fk{e}\hlhu[1]{,}\fk{6,a} it appears that the neonate is experiencing a \hlhu[1]{mild} \hlhu[1]{to}\fk{f} moderate neonatal \hlhu[1]{anemia,}\fk{f} likely related \hlhu[1]{to} \hlhu[1]{the}\fk{g} mother\hlhu[1]{'s}\fk{g} chorioamnionitis\hlhu[1]{.}\fk{a} The \hlhu[1]{low}\fk{f} MCH, MCHC, and platelet \hlhu[1]{count}\fk{e,f}, along with the high RDW and neutrophils, suggest that the neonate may \hlhu[1]{be}\fk{f} experiencing a \hlhu[1]{hem}\fk{g}olytic process, possibly due to \hlhu[1]{maternal}\fk{g} infection\hlhu[1]{.}\fk{a} The \hlhu[1]{slight}\fk{e,f} enlarged \hlhu[1]{platelets}\fk{f} and poly\hlhu[1]{chrom}\fk{b}asia also \hlhu[1]{support}\fk{f} this diagnosis\hlhu[1]{.}\fk{a} The anisocytosis may \hlai[1]{indicate} \hlai[1]{that}\fk{6} the neonate \hlhu[1]{is}\fk{f} \hlai[1]{experiencing}\fk{6} a \hlhu[1]{degree} \hlhu[1]{of}\fk{d} hemolysis, which is consistent with the mother's ch\hlai[1]{orio}\fk{3}amnionitis.\hlhu[1]{It}\fk{4,a}'\hlai[2]{s}\fk{1,4} \hlai[2]{essential}\fk{1,2} \hlai[2]{to}\fk{1,a} \hlhu[1]{monitor} \hlhu[1]{the}\fk{a} neon\hlhu[1]{ate}\fk{a}'\hlhu[1]{s} \hlhu[1]{condition} \hlhu[1]{closely}\fk{a}\hlhu[1]{,}\fk{6,a} \hlhu[1]{as}\fk{a} chorioamnionitis can lead \hlhu[1]{to}\fk{c} neonatal sepsis and other \hlhu[1]{complications.} \hlhu[1]{The}\fk{a} neon\hlhu[1]{ate} \hlhu[1]{may} \hlhu[1]{require}\fk{a} close \hlhu[1]{monitoring}\fk{a}\hlai[2]{,}\fk{1,6,a} \hlhu[1]{possible}\fk{6,a} \hlhu[1]{antibiotics}\fk{a}\hlhu[1]{,}\fk{a,g} \hlhu[1]{and}\fk{6,a,g} \hlhu[1]{possibly}\fk{a,g} \hlhu[1]{blood} \hlhu[1]{trans}\fk{g}fusions to \hlai[1]{manage}\fk{6} \hlhu[1]{the}\fk{g} anemia\hlhu[1]{.}\fk{a}\hlai[2]{Please}\fk{1,a} \hlai[3]{note}\fk{1,2,5,7} \hlai[3]{that}\fk{1,5,a} \hlai[3]{this} \hlai[3]{is}\fk{1} \hlai[3]{a}\fk{1,a,e} \hlai[3]{general} \hlai[3]{interpretation}\fk{1,b} \hlai[3]{and}\fk{1,a,b} \hlai[3]{a} \hlai[3]{more}\fk{1,a} \hlai[3]{detailed}\fk{1,a,b} \hlai[3]{evaluation} \hlai[2]{should} \hlai[2]{be} \hlai[3]{performed}\fk{1,a} \hlai[3]{by}\fk{1,a,c} \hlai[2]{a}\fk{1,a,g} \hlai[2]{healthcare}\fk{1} \hlai[3]{professional.}\fk{1,a}
&
\textit{Toward AI}\newline
\fk{1}\,$+2.10$\; introductory phrases or filler text commonly used by ai language models when declining personal questions\newline
\fk{2}\,$+0.26$\; word important and its synonyms like crucial when used to emphasize the significance of advice or caution\newline
\fk{3}\,$+0.19$\; repeated phrases or sequences in text\newline
\fk{4}\,$+0.11$\; activates on the phrase it is specifically when used as a formal or impersonal subject\newline
\fk{5}\,$+0.07$\; transitions used to summarize or conclude a text\newline
\fk{6}\,$+0.06$\; words and phrases that introduce supplementary information or instructions within a sequence\newline
\fk{7}\,$+0.06$\; dates and timestamps in forum posts or metadata\newline
\textit{+ 7 more}\; $+0.25$\newline
\textit{Toward human}\newline
\fk{a}\,$-0.23$\; text following or concluding medical advice or health consultations\newline
\fk{b}\,$-0.18$\; words related to estimations or imprecise approximations\newline
\fk{c}\,$-0.13$\; parenthetical clauses or aside remarks in text\newline
\fk{d}\,$-0.08$\; words related to quantifying or subsetting groups like part some or partially\newline
\fk{e}\,$-0.06$\; accounting and bookkeeping ledger entries involving debit credit and bank account transactions\newline
\fk{f}\,$-0.06$\; subjective judgment phrases or opinions about quality\newline
\fk{g}\,$-0.04$\; words related to physical sensations or gore such as bleeding footprints or bloodcurdling screams\newline
\textit{+ 6 more}\; $-0.76$
&
\begin{tabular}[t]{@{}lr@{}}
$\Sigma$ AI    & $+3.10$ \\
$\Sigma$ human & $-1.54$ \\
bias           & $+0.61$ \\
\midrule
logit          & $+2.14$ \\
$P(\mathrm{AI})$ & $89\%$ \\
\multicolumn{2}{@{}l}{\textbf{$\rightarrow$ AI}} \\
\end{tabular}
\\
\bottomrule
\end{tabular}
\caption{The \textsc{FT-SAE} ($\tau = 0.85$) logistic-regression detector applied to a Llama3-written document from the COLING test set (\texttt{coling29322\_CUDRT}). Tokens on which a listed feature fires are shaded red (feature pushes toward AI) or blue (toward human), with three intensities set by the token-level contribution $|w_i a_{i,t}|$ of the strongest listed feature on that token.}
\label{tab:worked_example_coling29322-CUDRT}
\end{table*}

\begin{table*}[t]
\centering
\footnotesize
\setlength{\tabcolsep}{4pt}
\renewcommand{\arraystretch}{1.05}
\begin{tabular}{@{}p{5cm}p{8cm}l@{}}
\toprule
\textbf{Input text} & \textbf{Active features} (Logit $\Delta$) & \textbf{Score}\\
\midrule
\textit{(Human)}~This paper deals with neuromorphic computing architectures for solving deep learning problems. Given that the energy-efficiency is a critical need for modern machine learning tools, this is an important study in \hlhu[3]{my}\fk{b,c} opinion. Until \hlai[1]{now,}\fk{8} most of the spiking neuron based works don't \hlhu[2]{really} \hlhu[2]{apply}\fk{e} for realistic deep model learning. Therefore, \hlai[2]{advancements}\fk{2,3,4,7,8} \hlai[1]{such} \hlai[1]{as} \hlai[1]{these}\fk{8} \hlai[1]{would}\fk{5,8} \hlai[1]{be}\fk{8} \hlhu[2]{very}\fk{d,f} \hlai[1]{useful.}\fk{8} \newline\mbox{}\newline However, my biggest concern is that the paper may be a bit \hlhu[2]{difficult}\fk{1,6,a} \hlhu[3]{to}\fk{1,a} \hlai[2]{grasp} \hlai[2]{for} \hlai[2]{the} \hlai[1]{general}\fk{1} ML \hlai[1]{audience}\fk{1}\hlai[1]{.}\fk{6} I \hlhu[2]{myself}\fk{c} \hlhu[2]{am}\fk{c,g,h} \hlhu[1]{not} \hlhu[1]{an} \hlhu[2]{expert} \hlhu[1]{in}\fk{g,h} \hlhu[1]{this}\fk{g} \hlhu[1]{topic}\fk{g,h} \hlhu[1]{and} \hlhu[1]{had} \hlhu[1]{quite} \hlhu[1]{a} \hlhu[1]{bit} \hlhu[1]{of}\fk{h} \hlhu[1]{difficulty}\fk{a,h} \hlhu[1]{as}\fk{h} an ML researcher \hlhu[1]{in} \hlhu[1]{understanding}\fk{g,h} \hlhu[1]{parts} \hlhu[1]{of}\fk{h} the \hlai[1]{paper}\fk{6}\hlhu[1]{.}\fk{h}
&
\textit{Toward AI}\newline
\fk{1}\,$+0.19$\; text concerning accessibility or difficulty of academic papers for specific readerships\newline
\fk{2}\,$+0.11$\; words related to progress or improvements such as advances advancement advancements or shifts\newline
\fk{3}\,$+0.09$\; words related to advancement or progress including advances and advancing\newline
\fk{4}\,$+0.09$\; words related to research and development progress specifically experimentation and advancement\newline
\fk{5}\,$+0.07$\; academic phrases predicting future influence or impact on researchers\newline
\fk{6}\,$+0.06$\; academic writing style describing paper accessibility or technical complexity\newline
\fk{7}\,$+0.04$\; academic terminology related to progress or improvements in research\newline
\fk{8}\,$+0.04$\; advancements or improvements in technical research or methodology\newline
\textit{+ 6 more}\; $+0.09$\newline
\textit{Toward human}\newline
\fk{a}\,$-0.39$\; readability of academic papers specifically phrases describing ease or difficulty of comprehension\newline
\fk{b}\,$-0.25$\; first person singular pronouns referring to the author or reviewer\newline
\fk{c}\,$-0.18$\; first-person opinion markers like i believe or i would say\newline
\fk{d}\,$-0.14$\; intensifier words like extremely and very\newline
\fk{e}\,$-0.11$\; contractions and auxiliary verb phrases like ve, s, and t\newline
\fk{f}\,$-0.11$\; intensifier adverb very or so often used to modify adjectives or quantifiers in critical evaluation contexts\newline
\fk{g}\,$-0.08$\; expressing lack of domain knowledge or expertise\newline
\fk{h}\,$-0.08$\; first person singular phrases expressing confusion or inability to understand something\newline
\textit{+ 9 more}\; $-0.21$
&
\begin{tabular}[t]{@{}lr@{}}
$\Sigma$ AI    & $+0.78$ \\
$\Sigma$ human & $-1.53$ \\
bias           & $-0.51$ \\
\midrule
logit          & $-1.26$ \\
$P(\mathrm{AI})$ & $22\%$ \\
\multicolumn{2}{@{}l}{\textbf{$\rightarrow$ Human}} \\
\end{tabular}
\\
\bottomrule
\end{tabular}
\caption{The \textsc{FT-SAE} ($\tau = 0.85$) logistic-regression detector applied to a human-written review from the PRDB test set (\texttt{ICLR2020-205-rkxbmCrTKS}). Tokens on which a listed feature fires are shaded red (feature pushes toward AI) or blue (toward human), with three intensities set by the token-level contribution $|w_i a_{i,t}|$ of the strongest listed feature on that token.}
\label{tab:worked_example_human-ICLR2020-205-rkxbmCrTKS}
\end{table*}

\begin{table*}[t]
\centering
\footnotesize
\setlength{\tabcolsep}{4pt}
\renewcommand{\arraystretch}{1.05}
\begin{tabular}{@{}p{6.5cm}p{6.5cm}l@{}}
\toprule
\textbf{Input text} & \textbf{Active features} (Logit $\Delta$) & \textbf{Score}\\
\midrule
\textit{(Human)}~Post \hlhu[3]{rebuttal:} \hlhu[2]{the} \hlhu[3]{author} \hlhu[3]{feedback}\fk{d} \hlhu[3]{addresses}\fk{6,d} \hlhu[3]{my} \hlhu[3]{concerns}\fk{a,d}\hlhu[3]{,} \hlhu[2]{though} \hlhu[2]{I}\fk{d} still think \hlhu[2]{the}\fk{d} readability of the \hlai[1]{paper}\fk{7} needs \hlai[1]{improvement}\fk{5}\hlhu[2]{.}\fk{d} I increased the overall score. -{}--{}--{}--{}-- 

My main concern is that the main algorithm (\hlhu[3]{Sec}\fk{b} \hlhu[3]{4}\fk{f}) and the privacy analysis are not clearly stated, making it hard to check the quality of the paper. \hlai[1]{More}\fk{3} specifically, In terms of \hlai[1]{clarity}\fk{2}, 

- Algorithm \hlhu[1]{1}\fk{f} and 2 \hlai[1]{needs}\fk{3,5} \hlai[1]{to}\fk{5} \hlai[1]{be}\fk{3,5} \hlai[1]{better}\fk{3} \hlhu[2]{explained}\fk{1,c} \hlhu[2]{in}\fk{c} the text. \hlhu[2]{The} \hlhu[2]{pseudo} \hlhu[2]{code}\fk{c} can \hlai[1]{also} \hlai[1]{be} \hlai[1]{made}\fk{5} \hlai[1]{more}\fk{3,5} \hlai[2]{readable}\fk{1,5}\hlai[1]{,}\fk{1} for \hlhu[3]{example,} \hlhu[3]{Line}\fk{c} \hlhu[3]{4}\fk{c} and \hlhu[2]{7} \hlhu[2]{are} \hlhu[2]{actually} \hlhu[2]{if-}\fk{c}then-else statements but the \hlhu[2]{else} \hlhu[2]{parts}\fk{c} are implicit (\hlhu[2]{the}\fk{c} full version is better). 

- \hlhu[3]{PTERR}\fk{e}IFICBALL \hlhu[2]{should}\fk{c} be \hlai[1]{explained}\fk{4} more\hlhu[2]{.}\fk{c} It is unclear \hlhu[2]{how} \hlhu[2]{much}\fk{c} \hlhu[3]{noise}\fk{e} is added. 

\hlhu[2]{-}\fk{c} It \hlhu[2]{seems}\fk{g} to \hlhu[1]{me}\fk{g} Algorithm \hlhu[2]{1} \hlhu[3]{would} \hlhu[2]{find}\fk{c} \hlhu[3]{k}\fk{e} clusters. Yet \hlhu[2]{the}\fk{c} recursive \hlhu[3]{part} \hlhu[2]{calls}\fk{c} it twice with k-1\hlhu[2]{.}\fk{c} Doesn't that mean we would have 2(k-1) clusters? 

- Line \hlhu[2]{3}\fk{c,e} in \hlhu[3]{Algorithm}\hlhu[3]{2} \hlhu[3]{mentioned}\fk{c} \hlhu[3]{a}\fk{c,e} \hlhu[2]{subroutine} \hlhu[2]{P}\fk{e}GE which is not defined. PGE \hlhu[3]{seems}\fk{g} to be important as it is the main component in Algorithm 2 \hlhu[2]{that}\fk{c} guarantees Differential Privacy.
&
\textit{Toward AI}\newline
\fk{1}\,$+0.16$\; words related to requests for improved clarity or explanation in academic reviews\newline
\fk{2}\,$+0.06$\; clarity clarification and related headings in academic review forms\newline
\fk{3}\,$+0.05$\; academic peer review feedback requesting more clarity or examples\newline
\fk{4}\,$+0.04$\; words related to research analysis and critique like elaborated explored or clarified\newline
\fk{5}\,$+0.03$\; criticism of document clarity or need for simplification\newline
\fk{6}\,$+0.02$\; academic paper verbs introducing research contributions\newline
\fk{7}\,$+0.02$\; academic evaluation of research papers focusing on methodology impact and practical applicability\newline
\textit{+ 3 more}\; $+0.05$\newline
\textit{Toward human}\newline
\fk{a}\,$-0.36$\; first person singular pronouns referring to the author or reviewer\newline
\fk{b}\,$-0.36$\; figures and tables in technical documents\newline
\fk{c}\,$-0.34$\; academic peer review feedback focusing on mathematical notation equation parts and specific paper reference structure\newline
\fk{d}\,$-0.27$\; post-rebuttal review comments and feedback\newline
\fk{e}\,$-0.26$\; mathematical notation and variables in technical papers\newline
\fk{f}\,$-0.24$\; referencing sections or specific numbers within technical or academic papers\newline
\fk{g}\,$-0.20$\; forms of the word seem including seem seemingly seems\newline
\textit{+ 31 more}\; $-1.99$
&
\begin{tabular}[t]{@{}lr@{}}
$\Sigma$ AI    & $+0.42$ \\
$\Sigma$ human & $-4.02$ \\
bias           & $-0.51$ \\
\midrule
logit          & $-4.11$ \\
$P(\mathrm{AI})$ & $2\%$ \\
\multicolumn{2}{@{}l}{\textbf{$\rightarrow$ Human}} \\
\end{tabular}
\\
\bottomrule
\end{tabular}
\caption{The \textsc{FT-SAE} ($\tau = 0.85$) logistic-regression detector applied to a human-written review from the PRDB test set (\texttt{NeurIPS2019-16-af0539fdca}). Tokens on which a listed feature fires are shaded red (feature pushes toward AI) or blue (toward human), with three intensities set by the token-level contribution $|w_i a_{i,t}|$ of the strongest listed feature on that token.}
\label{tab:worked_example_human-NeurIPS2019-16-af0539fdca}
\end{table*}

\begin{table*}[t]
\centering
\footnotesize
\setlength{\tabcolsep}{4pt}
\renewcommand{\arraystretch}{1.05}
\begin{tabular}{@{}p{7.3cm}p{5.5cm}l@{}}
\toprule
\textbf{Input text} & \textbf{Active features} (Logit $\Delta$) & \textbf{Score}\\
\midrule
\textit{(Human)}~The reason is, stores want customers to use cash. By giving us cash, we are more likely to use cash next time. I feel a little guilty when using my bank card at the store because I \hlai[1]{know}\fk{5} I'm \hlai[1]{giving}\fk{1} about 2-3\% of the sale to the bank. Unless I don't really like where I'm \hlhu[2]{shopping} \hlhu[2]{(ie} \hlhu[3]{Walmart),} \hlhu[2]{I}\fk{c} try to use cash if I have it. I \hlai[1]{doubt}\fk{5} these large \hlhu[2]{stores} \hlhu[2]{pay}\fk{e} \hlhu[2]{extra}\fk{3,e} \hlai[1]{for}\fk{3} \hlai[2]{supplying} \hlai[1]{the}\fk{1} cash portion. \hlhu[2]{They} \hlhu[2]{just}\fk{e} need to keep the cash onhand. In \hlhu[3]{other}\fk{6,a} countries, do \hlhu[2]{they}\fk{e} not mind paying banks a percentage of each transaction? That's a \hlhu[3]{huge}\fk{f} loss for retailers. (\hlhu[3]{I}\fk{2,d} \hlai[3]{also}\fk{2,4} heard \hlhu[3]{tipping}\fk{e} isn'\hlhu[2]{t} \hlhu[2]{popular} \hlhu[3]{in}\fk{e} some countries, maybe the lack \hlhu[2]{of}\fk{e} regard for \hlhu[3]{vendors}\fk{e} \hlhu[3]{is}\fk{g} related somehow\hlai[1]{??)}\fk{7} Oh\hlhu[3]{,}\fk{a} plus, it\hlhu[2]{'}\fk{e}s \hlhu[2]{a}\fk{e} value \hlhu[3]{added}\fk{3,a} \hlhu[2]{service}\fk{e}. A customer is more likely to \hlhu[3]{return}\fk{b} to a store if \hlai[1]{they}\fk{5} \hlai[3]{provide}\fk{1,e} \hlai[2]{this}\fk{1,5,e} \hlhu[2]{service}\fk{e}.
&
\textit{Toward AI}\newline
\fk{1}\,$+0.12$\; forms of the verb provide or related synonyms\newline
\fk{2}\,$+0.11$\; introductory phrases used to preface an additional observation or note\newline
\fk{3}\,$+0.08$\; words indicating an increase or supplement such as addition additional and extra\newline
\fk{4}\,$+0.04$\; word also as a transition or additive marker\newline
\fk{5}\,$+0.03$\; words or phrases used in contemplative or introspective personal reflections\newline
\fk{6}\,$+0.03$\; word such used to introduce examples\newline
\fk{7}\,$+0.03$\; exclamation points at the end of sentences or phrases\newline
\textit{+ 12 more}\; $+0.22$\newline
\textit{Toward human}\newline
\fk{a}\,$-0.20$\; conversational meta-commentary phrases like off topic or similar topic\newline
\fk{b}\,$-0.20$\; refund and return policies or processes\newline
\fk{c}\,$-0.15$\; words or phrases commonly found in customer reviews or informal first-person narratives\newline
\fk{d}\,$-0.12$\; words or phrases indicating imprecision or approximate estimations\newline
\fk{e}\,$-0.11$\; words indicating service quality or restaurant experiences including staff interactions and feelings about service\newline
\fk{f}\,$-0.11$\; repeated capitalized or emphatic adverbs used for emphasis\newline
\fk{g}\,$-0.11$\; insulting or dismissive colloquial language directed at a person or their ideas\newline
\textit{+ 31 more}\; $-0.89$
&
\begin{tabular}[t]{@{}lr@{}}
$\Sigma$ AI    & $+0.67$ \\
$\Sigma$ human & $-1.89$ \\
bias           & $+0.66$ \\
\midrule
logit          & $-0.58$ \\
$P(\mathrm{AI})$ & $36\%$ \\
\multicolumn{2}{@{}l}{\textbf{$\rightarrow$ Human}} \\
\end{tabular}
\\
\bottomrule
\end{tabular}
\caption{The \textsc{Base-SAE} ($\tau = 0.85$) logistic-regression detector applied to a human-written document from the COLING test set (\texttt{coling53434\_CUDRT}). Tokens on which a listed feature fires are shaded red (feature pushes toward AI) or blue (toward human), with three intensities set by the token-level contribution $|w_i a_{i,t}|$ of the strongest listed feature on that token.}
\label{tab:worked_example_coling53434-CUDRT}
\end{table*}

\begin{table*}[t]
\centering
\footnotesize
\setlength{\tabcolsep}{4pt}
\renewcommand{\arraystretch}{1.05}
\begin{tabular}{@{}p{7.3cm}p{5.5cm}l@{}}
\toprule
\textbf{Input text} & \textbf{Active features} (Logit $\Delta$) & \textbf{Score}\\
\midrule
\textit{(Human)}~This paper proposes an \hlhu[1]{out}\fk{g}-of-\hlai[1]{distribution}\fk{3} (OOD) \hlai[1]{detection}\fk{3} method, where the features of in-\hlai[1]{distribution}\fk{3} (ID) \hlai[1]{samples}\fk{3,6} and those of the OOD samples are decor\hlai[1]{related}\fk{3} in a class-\hlai[1]{wise}\fk{3} manner by using HSIC. A theoretical guarantee for decorrelation is \hlhu[2]{provided}\fk{b,c,e}. The \hlhu[2]{prop}\fk{b}posed method, Conditional-i, is \hlhu[2]{evaluated} \hlhu[2]{on}\fk{a} \hlai[1]{O}\fk{5}\hlhu[1]{OD}\fk{5,e} \hlhu[2]{detection}\fk{a,d} benchmarks in \hlhu[2]{image}\fk{a} classification and NLP, and shows S\hlhu[1]{OTA}\fk{d} results.\newline\mbox{}\newline \hlhu[2]{Reviewers}\fk{1,a} \hlhu[2]{had} \hlhu[3]{many} \hlhu[3]{major} \hlhu[3]{concerns,} \hlhu[3]{e.g.,} \hlhu[3]{novelty} \hlhu[3]{from} \hlhu[3]{HOOD,} \hlhu[3]{missing}\fk{a} baselines \hlhu[2]{with}\fk{a} \hlhu[3]{OOD}\fk{5,a} \hlhu[2]{exposure}\fk{a,b}\hlhu[2]{,} \hlhu[2]{hyperparameter} \hlhu[2]{tuning,} \hlhu[3]{and} \hlhu[2]{insufficient} \hlhu[2]{theory}\fk{a} to \hlhu[2]{support} \hlhu[2]{the} \hlhu[2]{proposed} \hlhu[2]{method}\fk{a}, \hlhu[2]{which} \hlhu[2]{the}\fk{a} \hlai[1]{authors}\fk{4} \hlhu[2]{addressed} \hlhu[3]{well,} \hlhu[2]{and} \hlhu[3]{most} \hlhu[2]{of} \hlhu[2]{the}\fk{a} \hlhu[2]{reviewers}\fk{a,b} \hlhu[3]{have}\fk{a} \hlhu[2]{been}\fk{2,7,a} \hlhu[3]{convinced}\fk{a,c}\hlhu[3]{.\textvisiblespace} \hlhu[2]{$\hookleftarrow$}\newline\mbox{}\newline \hlhu[3]{The} \hlhu[3]{paper} \hlhu[3]{is} \hlhu[3]{well-}\fk{a}\hlhu[3]{written}\fk{a,b,g} \hlhu[3]{and} \hlhu[3]{the} \hlhu[3]{mathematical} \hlhu[3]{notation} \hlhu[3]{seems} \hlhu[3]{fine} \hlhu[3]{for} \hlhu[3]{machine} \hlhu[3]{learners.\textvisiblespace} \hlhu[3]{However,} \hlhu[3]{the}\fk{a} \hlhu[3]{relation}\fk{a,b} \hlhu[3]{between} \hlhu[3]{Theorem\textvisiblespace} \hlhu[3]{1} \hlhu[2]{and} \hlhu[3]{Assumption}\fk{a} \hlhu[3]{1} \hlhu[3]{is} \hlhu[3]{weird.\textvisiblespace} \hlhu[3]{Any} \hlhu[3]{theorem} \hlhu[3]{must} \hlhu[3]{always} \hlhu[3]{hold} \hlhu[3]{(otherwise} \hlhu[3]{it's} \hlhu[2]{not} \hlhu[3]{a}\fk{a} theorem\hlhu[2]{),} \hlhu[3]{and} \hlhu[3]{the}\fk{a} \hlhu[3]{sentence}\fk{a,b} \hlhu[3]{``}\fk{a}\hlhu[2]{Theorem}\fk{a,f}\hlhu[2]{\textvisiblespace} \hlhu[3]{1} \hlhu[3]{holds} \hlhu[2]{if} \hlhu[3]{Assumption\textvisiblespace} \hlhu[3]{1} \hlhu[3]{holds''} \hlhu[3]{doesn't}\fk{a} make sense.
&
\textit{Toward AI}\newline
\fk{1}\,$+0.49$\; academic and technical publication identifiers including urls and scholarly citation elements\newline
\fk{2}\,$+0.17$\; past participle of to be\newline
\fk{3}\,$+0.10$\; mathematical and computer science technical terminology and formal definitions\newline
\fk{4}\,$+0.09$\; words related to authorship and citation such as author and authors\newline
\fk{5}\,$+0.02$\; proper names and capitalized personal identifiers in narrative or academic text\newline
\fk{6}\,$+0.02$\; words related to news announcements corporate updates and official company statements\newline
\fk{7}\,$+0.01$\; past participle been\newline
\textit{+ 1 more}\; $+0.03$\newline
\textit{Toward human}\newline
\fk{a}\,$-1.26$\; academic peer review feedback context\newline
\fk{b}\,$-0.88$\; word article when referring to specific texts or media content\newline
\fk{c}\,$-0.30$\; past participle verbs indicating future availability or state change such as released or available\newline
\fk{d}\,$-0.22$\; words ending in logist or surgeon related to medical professions\newline
\fk{e}\,$-0.22$\; narrative fiction character actions and pronouns\newline
\fk{f}\,$-0.07$\; abbreviations for change my view or reddit forum markers\newline
\fk{g}\,$-0.06$\; words and phrases describing restaurant and service quality experiences\newline
\textit{+ 2 more}\; $-0.89$
&
\begin{tabular}[t]{@{}lr@{}}
$\Sigma$ AI    & $+0.94$ \\
$\Sigma$ human & $-3.90$ \\
bias           & $+0.53$ \\
\midrule
logit          & $-2.43$ \\
$P(\mathrm{AI})$ & $8\%$ \\
\multicolumn{2}{@{}l}{\textbf{$\rightarrow$ Human}} \\
\end{tabular}
\\
\bottomrule
\end{tabular}
\caption{The \textsc{FT-Model+SAE} ($\tau = 0.85$) logistic-regression detector applied to a human-written document from the COLING test set (\texttt{coling12300\_peersum}). Tokens on which a listed feature fires are shaded red (feature pushes toward AI) or blue (toward human), with three intensities set by the token-level contribution $|w_i a_{i,t}|$ of the strongest listed feature on that token.}
\label{tab:worked_example_coling12300-peersum}
\end{table*}

\begin{table*}[t]
\centering
\footnotesize
\setlength{\tabcolsep}{4pt}
\renewcommand{\arraystretch}{1.05}
\begin{tabular}{@{}p{7.3cm}p{5.5cm}l@{}}
\toprule
\textbf{Input text} & \textbf{Active features} (Logit $\Delta$) & \textbf{Score}\\
\midrule
\textit{(Llama)}~The paper presents a novel approach for verifying the local \hlai[1]{robustness}\fk{7} of \hlai[1]{networks} \hlai[1]{with} \hlai[1]{piecewise} \hlai[1]{linear} \hlai[1]{activation} \hlai[1]{functions,} \hlai[1]{which} \hlai[1]{relies} \hlai[1]{on} \hlai[1]{geometric} \hlai[1]{projections}\fk{7} \hlai[1]{rather}\fk{7,c} \hlai[1]{than} \hlai[1]{constraint} \hlai[1]{solving} \hlai[1]{or} \hlai[1]{conservative} \hlai[2]{over-approximations}\fk{7}. The approach is particularly suited for \hlai[1]{certifying} \hlai[1]{$\ell$2-robustness} \hlai[1]{and}\fk{7} has \hlhu[1]{been}\fk{a,b} \hlai[1]{shown} \hlai[1]{to} \hlai[1]{outperform} \hlai[1]{existing}\fk{7} exact tools \hlai[1]{by} \hlai[1]{multiple} \hlai[1]{orders} \hlai[1]{of} \hlai[1]{magnitude}\fk{7} while \hlai[1]{maintaining}\fk{7} the same \hlai[1]{or}\fk{7} \hlai[1]{better}\fk{7,d} \hlai[1]{precision} \hlai[1]{under}\fk{7} a \hlai[1]{time} \hlai[1]{constraint}\fk{7}. The authors \hlai[2]{provide}\fk{6} a \hlai[3]{thorough}\fk{1} evaluation of their approach, including comparisons with other tools and a discussion of its \hlai[2]{limitations}\fk{5}. The paper is well-\hlai[3]{written}\fk{2} and the presentation is clear and concise. The authors have also made their implementation available on GitHub, which is a significant contribution to the research community. The only \hlai[2]{potential}\fk{4} drawback of the approach is its incompleteness, which may lead to unknown results in some cases. However, the authors \hlai[2]{provide}\fk{6} a \hlai[3]{thorough}\fk{1} discussion of this \hlai[2]{limitation}\fk{5} and its implications. Overall, the paper is a significant contribution to the \hlai[3]{field}\fk{3} of neural network verification and is likely to be of interest to researchers and practitioners in this area.
&
\textit{Toward AI}\newline
\fk{1}\,$+0.99$\; the word thorough\newline
\fk{2}\,$+0.63$\; word written used to describe paper quality\newline
\fk{3}\,$+0.50$\; academic field noun when used in singular or plural form\newline
\fk{4}\,$+0.45$\; activates on the word potential\newline
\fk{5}\,$+0.32$\; word limitations within academic paper reviews\newline
\fk{6}\,$+0.23$\; forms of the verb provide in academic paper reviews\newline
\fk{7}\,$+0.22$\; academic paper method descriptions and technical approach summaries\newline
\textit{+ 5 more}\; $+0.37$\newline
\textit{Toward human}\newline
\fk{a}\,$-0.16$\; past participle been often following have or had\newline
\fk{b}\,$-0.11$\; the word been in passive voice constructions\newline
\fk{c}\,$-0.05$\; adversative transition words like rather and instead\newline
\fk{d}\,$-0.03$\; word better or best used to compare performance or quality\newline
\textit{+ 2 more}\; $-0.08$
&
\begin{tabular}[t]{@{}lr@{}}
$\Sigma$ AI    & $+3.71$ \\
$\Sigma$ human & $-0.43$ \\
bias           & $-0.37$ \\
\midrule
logit          & $+2.91$ \\
$P(\mathrm{AI})$ & $95\%$ \\
\multicolumn{2}{@{}l}{\textbf{$\rightarrow$ AI}} \\
\end{tabular}
\\
\bottomrule
\end{tabular}
\caption{The \textsc{FT-Model+SAE} (Llama) ($\tau = 0.85$) logistic-regression detector applied to a Llama-written review from the PRDB test set (\texttt{ICLR2021-1934-6POeD7QDvXS}). Tokens on which a listed feature fires are shaded red (feature pushes toward AI) or blue (toward human), with three intensities set by the token-level contribution $|w_i a_{i,t}|$ of the strongest listed feature on that token.}
\label{tab:worked_example_llama-3.1-70b-ICLR2021-1934-6POeD7QDvXS}
\end{table*}

\begin{table*}[t]
\centering
\small
\setlength{\tabcolsep}{3.5pt}
\resizebox{\textwidth}{!}{%
\begin{tabular}{lcccccccccccc}
\toprule
& \multicolumn{3}{c}{\textbf{COLING (dev)}} & \multicolumn{3}{c}{\textbf{COLING (devtest)}} & \multicolumn{3}{c}{\textbf{COLING (test)}} & \multicolumn{3}{c}{\textbf{RAID (test)}} \\
\cmidrule(lr){2-4} \cmidrule(lr){5-7} \cmidrule(lr){8-10} \cmidrule(lr){11-13}
 & AUC & F1 & TPR@$5$ & AUC & F1 & TPR@$5$ & AUC & F1 & TPR@$5$ & AUC & F1 & TPR@$5$ \\
\midrule
gram2vec & 81.9 & 77.9 & 36.2 & 80.1 & 73.1 & 45.3 & 77.7 & 72.0 & 33.6 & 77.5 & 73.5 & 43.6 \\
\quad \textcolor{gray}{+XGBoost} & \textcolor{gray}{93.1} & \textcolor{gray}{87.1} & \textcolor{gray}{69.2} & \textcolor{gray}{82.0} & \textcolor{gray}{75.1} & \textcolor{gray}{55.0} & \textcolor{gray}{81.5} & \textcolor{gray}{70.1} & \textcolor{gray}{53.8} & \textcolor{gray}{83.5} & \textcolor{gray}{82.1} & \textcolor{gray}{47.7} \\
\midrule
\textbf{w/o Mask} & & & & & & & & & & & & \\
\textsc{Base-SAE} & 97.1 & 91.4 & 84.0 & 96.2 & 93.2 & 90.4 & 94.2 & 88.9 & 81.8 & 95.2 & 92.3 & 86.4 \\
\quad \textcolor{gray}{+XGBoost} & \textcolor{gray}{98.9} & \textcolor{gray}{94.8} & \textcolor{gray}{92.8} & \textcolor{gray}{95.7} & \textcolor{gray}{91.6} & \textcolor{gray}{87.8} & \textcolor{gray}{92.9} & \textcolor{gray}{86.3} & \textcolor{gray}{79.1} & \textcolor{gray}{96.3} & \textcolor{gray}{93.3} & \textcolor{gray}{87.8} \\
\addlinespace
\textsc{FT-SAE} & 97.8 & 92.6 & 86.8 & 96.7 & 93.5 & 91.2 & 95.9 & 92.2 & 89.1 & 95.3 & 91.8 & 85.5 \\
\quad \textcolor{gray}{+XGBoost} & \textcolor{gray}{99.2} & \textcolor{gray}{96.1} & \textcolor{gray}{95.6} & \textcolor{gray}{96.4} & \textcolor{gray}{93.9} & \textcolor{gray}{91.4} & \textcolor{gray}{96.0} & \textcolor{gray}{91.9} & \textcolor{gray}{88.6} & \textcolor{gray}{96.8} & \textcolor{gray}{93.7} & \textcolor{gray}{89.6} \\
\addlinespace
\textsc{FT-Model+SAE} & 99.5 & 97.5 & 98.1 & 91.0 & 86.6 & 63.5 & 83.1 & 76.1 & 47.8 & 91.7 & 92.5 & 61.5 \\
\quad \textcolor{gray}{+XGBoost} & \textcolor{gray}{99.6} & \textcolor{gray}{97.8} & \textcolor{gray}{98.5} & \textcolor{gray}{91.3} & \textcolor{gray}{86.3} & \textcolor{gray}{71.8} & \textcolor{gray}{83.0} & \textcolor{gray}{75.6} & \textcolor{gray}{46.3} & \textcolor{gray}{92.5} & \textcolor{gray}{91.5} & \textcolor{gray}{71.3} \\
\midrule
\textbf{w/ Mask ($\tau\!>\!0.7$)} & & & & & & & & & & & & \\
\textsc{Base-SAE} & 95.2 & 89.4 & 75.5 & 93.8 & 88.6 & 80.5 & 88.2 & 80.9 & 61.3 & 93.0 & 89.8 & 78.6 \\
\quad \textcolor{gray}{+XGBoost} & \textcolor{gray}{97.2} & \textcolor{gray}{91.8} & \textcolor{gray}{82.7} & \textcolor{gray}{92.4} & \textcolor{gray}{87.0} & \textcolor{gray}{79.8} & \textcolor{gray}{90.2} & \textcolor{gray}{80.4} & \textcolor{gray}{69.7} & \textcolor{gray}{92.3} & \textcolor{gray}{87.8} & \textcolor{gray}{76.5} \\
\addlinespace
\textsc{FT-SAE} & 96.9 & 91.0 & 81.4 & 96.8 & 94.2 & 92.3 & 95.3 & 92.2 & 89.2 & 94.5 & 91.7 & 84.4 \\
\quad \textcolor{gray}{+XGBoost} & \textcolor{gray}{98.3} & \textcolor{gray}{93.6} & \textcolor{gray}{89.1} & \textcolor{gray}{96.6} & \textcolor{gray}{94.4} & \textcolor{gray}{92.6} & \textcolor{gray}{95.2} & \textcolor{gray}{90.5} & \textcolor{gray}{86.2} & \textcolor{gray}{95.0} & \textcolor{gray}{91.6} & \textcolor{gray}{86.0} \\
\addlinespace
\textsc{FT-Model+SAE} & 99.2 & 96.8 & 96.7 & 91.1 & 84.8 & 67.9 & 82.4 & 75.4 & 49.1 & 92.0 & 93.2 & 64.5 \\
\quad \textcolor{gray}{+XGBoost} & \textcolor{gray}{98.5} & \textcolor{gray}{94.8} & \textcolor{gray}{90.9} & \textcolor{gray}{90.8} & \textcolor{gray}{86.4} & \textcolor{gray}{74.1} & \textcolor{gray}{82.9} & \textcolor{gray}{74.8} & \textcolor{gray}{50.9} & \textcolor{gray}{89.9} & \textcolor{gray}{91.7} & \textcolor{gray}{54.8} \\
\bottomrule
\end{tabular}}
\caption{Performance comparison between Logistic regression and XGBoost detector heads trained on the COLING train set, evaluated on the held-out COLING and RAID splits. We see that XGBoost improves performance of gram2vec by a large margin. It also improves performance of our models but does so by a comparatively smaller amount. This suggests that our features are likely predominantly linear with relatively negligible interaction terms.}
\label{tab:head_comparison_coling}
\end{table*}

\begin{table*}[t]
\centering
\small
\setlength{\tabcolsep}{4pt}
\begin{tabular}{@{}lcccccccccccccccc@{}}
\toprule
\multicolumn{17}{@{}c}{\textbf{Document-Level Paraphrase Ablation Results}}\\
\midrule
 & & \multicolumn{3}{c}{Baseline} & \multicolumn{3}{c}{Few Shot} & \multicolumn{3}{c}{Guided} & \multicolumn{3}{c}{\textbf{SAE}} & \multicolumn{3}{c}{\textbf{SAE $\times 3$}}\\
\cmidrule(lr){3-5} \cmidrule(lr){6-8} \cmidrule(lr){9-11} \cmidrule(lr){12-14} \cmidrule(lr){15-17}
 & F1 & F1 & ASR & Lev & F1 & ASR & Lev & F1 & ASR & Lev & F1 & ASR & Lev & F1 & ASR & Lev\\
\midrule
\multicolumn{17}{@{}l}{\textbf{COLING}}\\
\textsc{Base-SAE} & 80.3 & 81.1 & 5.3 & 24.0 & 78.1 & 10.7 & 31.5 & 77.2 & 12.6 & 39.0 & 66.8 & 28.1 & 26.1 & 46.9 & 55.1 & 34.5\\
\textsc{FT-SAE} & 84.4 & 83.3 & 6.3 & 24.0 & 82.0 & 8.8 & 31.5 & 80.9 & 11.1 & 39.0 & 70.5 & 26.9 & 28.7 & 51.7 & 52.6 & 37.5\\
\textsc{FT-Model+SAE} & 88.3 & 86.0 & 8.7 & 24.0 & 82.9 & 14.7 & 31.5 & 79.5 & 20.7 & 39.0 & 69.9 & 34.2 & 26.5 & 45.3 & 63.2 & 33.7\\
\midrule
\multicolumn{17}{@{}l}{\textbf{PRDB (GPT-4o)}}\\
\textsc{Base-SAE} & 99.7 & 99.7 & 0.1 & 13.8 & 99.3 & 0.8 & 25.6 & 95.7 & 7.8 & 36.5 & 90.9 & 16.2 & 6.1 & 47.0 & 69.1 & 11.2\\
\textsc{FT-SAE} & 99.8 & 99.9 & 0.0 & 13.8 & 99.8 & 0.2 & 25.6 & 97.1 & 5.2 & 36.5 & 98.0 & 3.6 & 8.7 & 88.9 & 19.7 & 16.5\\
\textsc{FT-Model+SAE} & 99.8 & 99.8 & 0.0 & 13.8 & 99.0 & 1.6 & 25.6 & 82.6 & 29.3 & 36.5 & 76.3 & 38.1 & 6.7 & 29.0 & 83.0 & 11.3\\
\midrule
\multicolumn{17}{@{}l}{\textbf{PRDB (Llama)}}\\
\textsc{Base-SAE} & 99.1 & 99.0 & 0.6 & 21.9 & 96.1 & 6.2 & 39.4 & 90.9 & 15.2 & 36.8 & 86.2 & 23.0 & 10.5 & 56.7 & 59.7 & 17.8\\
\textsc{FT-SAE} & 99.5 & 99.4 & 0.1 & 21.9 & 97.1 & 4.6 & 39.4 & 92.4 & 13.2 & 36.8 & 97.7 & 3.4 & 14.1 & 87.2 & 21.8 & 23.3\\
\textsc{FT-Model+SAE} & 99.1 & 98.9 & 0.6 & 21.9 & 94.6 & 8.7 & 39.4 & 80.8 & 30.9 & 36.8 & 53.8 & 62.5 & 8.4 & 5.5 & 97.1 & 11.6\\
\bottomrule
\end{tabular}
\caption{Adversarial paraphrasing of 2500 AI-written documents per dataset (2500 human documents fixed). The first column is each detector's F1 before the attack. For every intervention we report F1 after paraphrasing, the attack success rate (ASR: share of documents detected before paraphrasing that are classified human afterwards) and the mean word-level Levenshtein distance to the original as a share of the longer text (Lev), all in percent.}
\label{tab:adv_appendix_all}
\end{table*}

\begin{table*}[t]
\centering
\small
\setlength{\tabcolsep}{3pt}
\renewcommand{\arraystretch}{1.15}
\begin{tabular}{@{}lrrp{6cm}p{6cm}@{}}
\toprule
\textbf{Split} & \textbf{\#Docs} & \textbf{\#Gen.} & \textbf{Generators (families)} & \textbf{Domains}\\
\midrule
\multicolumn{5}{@{}l}{\textbf{COLING}}\\
\texttt{train} & 610,767 & 39 & GPT-3.5, GPT-3 (davinci) (3), OPT (9), Llama (6), T0/BLOOM (4), FLAN-T5 (5), Cohere, Gemma (2), GPT-4/4o (2), Mistral, Dolly (2), GPT-J/NeoX (2), GLM & Q\&A, Reddit, stories, WikiHow, arXiv abstracts, Wikipedia, student essays, Yelp reviews, news, commonsense, opinion (CMV), scientific tables, peer reviews\\
\texttt{dev} & 261,758 & 39 & GPT-3.5, GPT-3 (davinci) (3), OPT (9), Llama (6), T0/BLOOM (4), FLAN-T5 (5), Cohere, Gemma (2), GPT-4/4o (2), Mistral, Dolly (2), GPT-J/NeoX (2), GLM & Q\&A, Reddit, stories, WikiHow, arXiv abstracts, Wikipedia, student essays, Yelp reviews, news, commonsense, opinion (CMV), scientific tables, peer reviews\\
\texttt{devtest} & 32,557 & 5 & GPT-4/4o, Gemini, Gemini (humanized), Mistral, Mistral (humanized) & arXiv abstracts, Wikipedia, WikiHow, Reddit, student essays, peer reviews, books, news, poetry, recipes, reddit, reviews, wiki\\
\texttt{test} & 73,941 & 14 & GPT-4/4o (4), Llama (3), GPT-4/Llama-2 (mixed), GLM, GPT-3.5, Qwen, Baichuan, Dolly, StableLM & IELTS essays, editing (expand, polish, refine, rewrite, summarise), peer-review summaries, peer reviews, mixed human--AI text (news, blog, email, speech, reviews, abstracts, QA)\\
\midrule
\multicolumn{5}{@{}l}{\textbf{PRDB}}\\
\texttt{train} & 319,239 & 2 & GPT-4o, Llama 3.1 70B & ICLR/NeurIPS peer reviews (2016--2024)\\
\texttt{test} & 171,837 & 5 & GPT-4o, Llama 3.1 70B, Claude 3.5 Sonnet, Gemini 1.5 Pro, Qwen 2.5 72B & ICLR/NeurIPS peer reviews (2016--2024)\\
\midrule
\multicolumn{5}{@{}l}{\textbf{RAID}}\\
\texttt{test} & 56,000 & 11 & MPT (2), Mistral (2), Cohere (2), GPT-2, Llama 2 70B, GPT-3 (davinci), ChatGPT, GPT-4 & arXiv abstracts, book reviews, news articles, poetry, recipes, reddit, IMDb movie reviews, Wikipedia introductory paragraphs\\
\bottomrule
\end{tabular}
\caption{Composition of the datasets used for training and evaluation. COLING \texttt{train} and \texttt{dev} are a stratified split of one pool (MAGE, M4GT-Bench and HC3). RAID is the non-adversarial portion of its test split. We see that all of our splits cover a wide spread of models and domains.}
\label{tab:datasets}
\end{table*}

\setlength{\fboxsep}{0pt}\setlength{\fboxrule}{0.4pt} 

\begin{figure*}[t]
\centering
\fbox{\includegraphics[width=\dimexpr\textwidth-2\fboxrule]{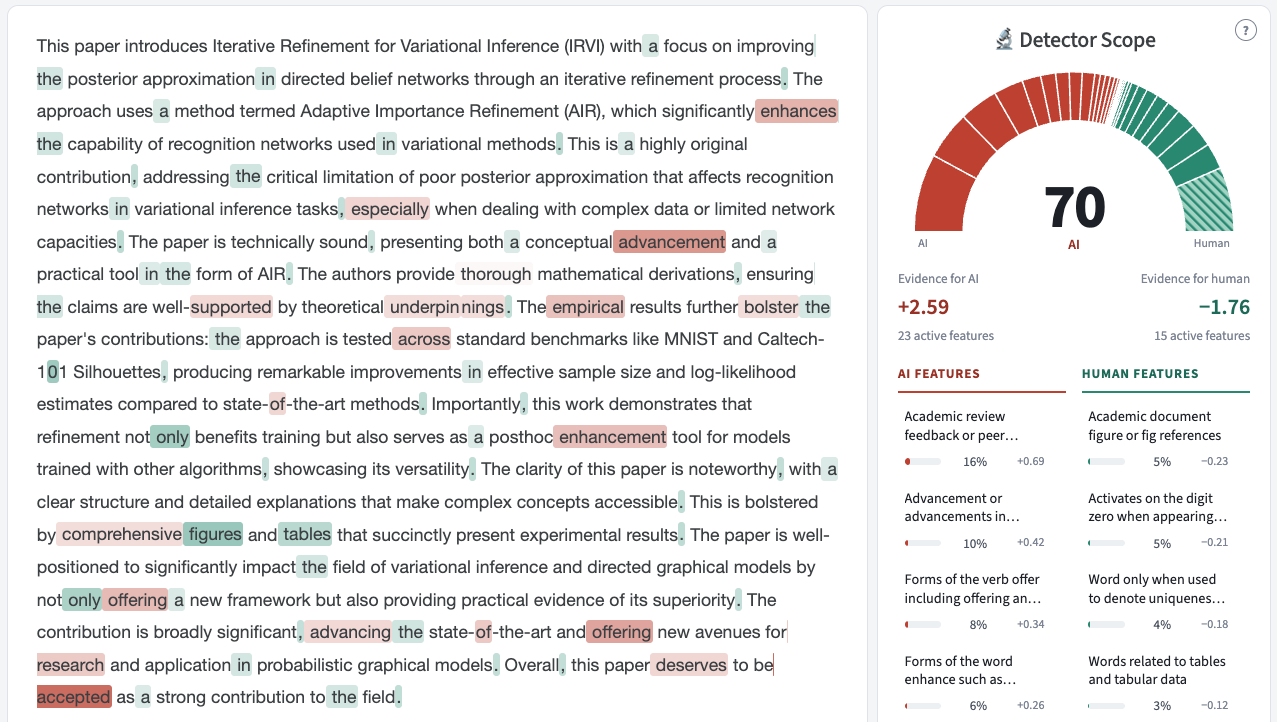}}

\vspace{1em}

\fbox{\includegraphics[width=\dimexpr\textwidth-2\fboxrule]{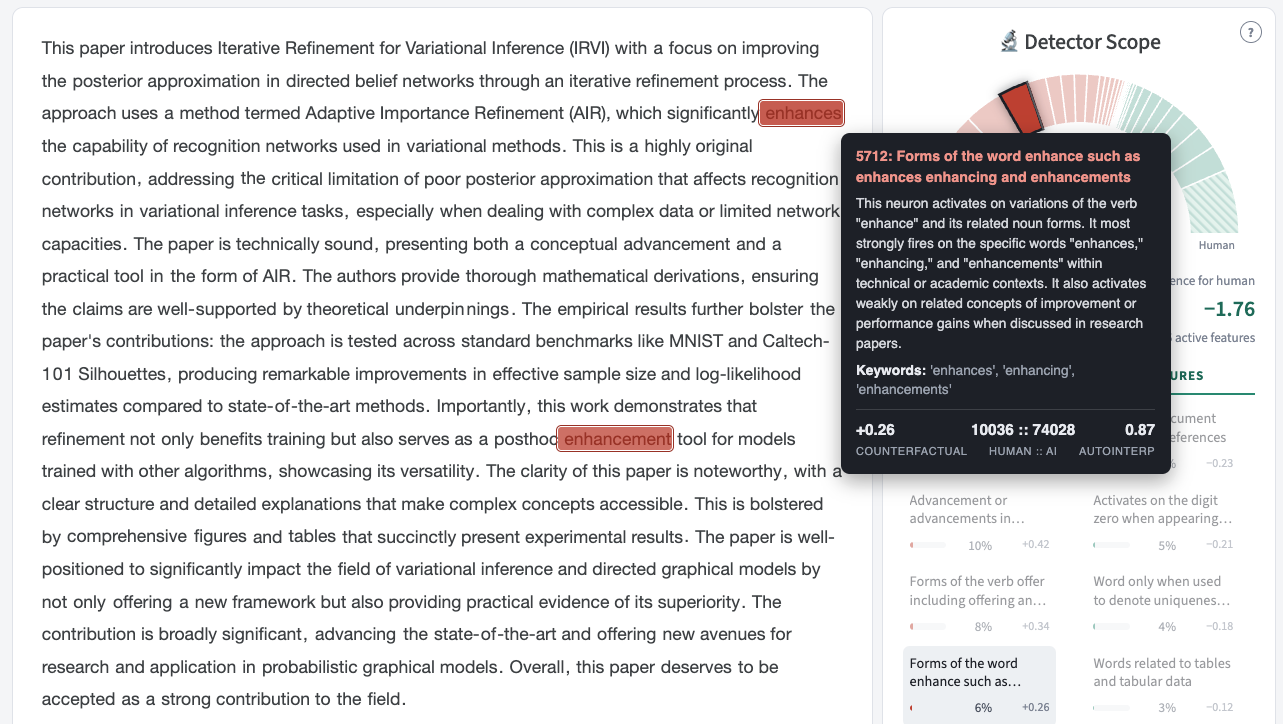}}
\caption{Interface for our interpretable AI detector. Top features for the given text are displayed on the right bar along with their description, autointerp score, local importance, and odds ratios. Text in the input is highlighted in proportion to feature strength. Hovering over highlighted tokens provides a more detailed description of the feature along with auto-interpretation score and number of occurrences in human vs. AI text.}
\label{fig:interface}
\end{figure*}

\begin{figure*}
  \small
  \centering
  \begin{lstlisting}
We're studying neurons in a neural network. Each neuron activates on some particular word or concept in a short document.
Each neuron is highly sparse, and will only activate on a small number of documents (and possibly a small number of words within each document).
There are a range of possible hypotheses, for example the neuron could activate on:

- specific syntactic features, e.g. punctuation like ")", ".", ":" or markdown like "**", or code syntax
- a specific set of words or phrases, e.g. "protest" or "protesters" (or other-language variants)
- specific words in particular contexts, e.g. "security" when it's used in the context of the bigram "social security"
- specific situations contexts, e.g. dialogue in online chat forums
- broader concepts, e.g. discussion of political statistics

You will be given a list of documents, sorted into groups according to the maximum strength of the activation across tokens in that document. Based on this, you will be asked to explain what this neuron is activating on.
For each sequence, the strongest activating words are indicated with <<...>> - note that these tags "<<" and ">>" are not part of the actual sentence, they're just used to highlight the places where the neuron is most active.
You'll be asked for several different kinds of explanations, which you should enclose in tags like <explanation_type>...</explanation_type> (remember to close all open tags!). In all cases, your explanation should be maximally specific and but still capture all the relevant information.

The different types of explanations you should provide are:

- <terse>: A short explanation, limited to 20 words or less. This explanation should be shorthand, omitting punctuation and formatting and capitalization. You can omit prefixes like 'This neuron fires on', instead describe it directly.
- <keywords>: A comma-separated list of up to 3 key words or short phrases that activate this neuron, in quotation marks '...'. Note that this should be actual examples where the neuron might fire, not descriptions of those examples (so it doesn't have to literally be single words, although often it will be).
- <long>: A comprehensive but concise explanation of what this neuron activates on. This should be 3 sentences: the first describes the concept briefly, the second describes what words or phrases most commonly activate it in the top activating documents, and the third describes what words or phrases cause it to also activate (but weaker) in some lower quantiles. If only one word or phrase activates the neuron within the document, make sure you specify which one (rather than just saying what kinds of documents activate it). Don't use overly formal language, be as simple and concise as possible. Don't include lists of more than 5 words or phrases in a row. Note that you don't have to list words if it's not relevant for concisely explaining the neuron.

So your final answer should look like: <terse>...</terse><keywords>...</keywords><long>...</long>

Some final instructions:
- Try to be specific in your explanations, although not so specific that you exclude most/all of the examples from matching your explanation.
- Pay attention to things like the capitalization and punctuation of the activating words or concepts, only if that seems relevant (e.g. if the activating words were all proper nouns).
- We're dealing with sentences that have been tokenized, so the neuron might only be active on tokens from a tokenized word rather than full words or phrases.
- Some neurons are token-level (i.e. activating on just a few specific words or very short phrases), some are concept-level (activating on most or all words in a sentence, if that sentence fits a particular class). In your long explanations, make sure you take this into account.

The activating documents are given below.

(*@\textcolor{green!50!black}{\{examples\}}@*)
  \end{lstlisting}
  \caption{Full system prompt for generating feature descriptions in the autointerpretation pipeline.}
  \label{fig:generation-prompt}
\end{figure*}

\begin{figure*}
  \small
  \centering
  \begin{lstlisting}
These are the highest-activating documents in the dataset:
1. 'compared to the results of the VQ-VAE method (https://arxiv.org/pdf(*@\textcolor{magenta}{{<}{<}}@*)/(*@\textcolor{magenta}{{>}{>}}@*)1711.00937.pdf or VQ-VAE '
2. ']). For nullness guards, you should take a look at [https://arxiv.org/pdf(https://arxiv.org/pdf(*@\textcolor{magenta}{{<}{<}}@*)/(*@\textcolor{magenta}{{>}{>}}@*)1902.06111.pdf]. I think it would improve the'

These are some documents from the group with score 3/4:
1. '[Transcriptomics-guided Slide Representation Learning in Computational Pathology](https://arxiv.org/html(*@\textcolor{magenta}{{<}{<}}@*)/(*@\textcolor{magenta}{{>}{>}}@*)2405.11618v1) that in cases where representations come'
2. 'results reported from the SAVi++ paper [here](https://browse.arxiv.org/pdf(*@\textcolor{magenta}{{<}{<}}@*)/(*@\textcolor{magenta}{{>}{>}}@*)2206.07764.pdf)). This hinders the evaluation of this'

These are some documents from the group with score 2/4:
1. '(Gonnet & Deselaers 2019) https://arxiv.org/abs(*@\textcolor{magenta}{{<}{<}}@*)/(*@\textcolor{magenta}{{>}{>}}@*)1903.08023\n\n**Questions**\n- Why does IM'
2. '\n\n[1] Torsional Diffusion for Molecular Conformer Generation: https://arxiv.org/abs(*@\textcolor{magenta}{{<}{<}}@*)/(*@\textcolor{magenta}{{>}{>}}@*)2206.01729\n\n[2] Swallowing the Bitter Pill'

These are some of the lowest-activating documents in the dataset:
1. '2019EA000812\nhttps://www.mdpi.com/2073-4433/10/5/244\n\n'
2. 'transfer learning for neurophysiological recordings, \nhttps://iopscience.iop.org/article/10.1088/1741-2552/abb'

These are documents that do not activate the feature at all:
1. 'larger training data set alterations.\n\nGood solid work, but lacking non-obvious results, and I do not see manuscript adressing the the harder challenges. However, the quantified results may have a notable'
2. 'novel in approach, the paper relies heavily on existing theories and methods, potentially limiting its ground-breaking appeal. Not all components, such as the lifting technique, are entirely new to the field.\n'
  \end{lstlisting}
  \caption{Example formatting for documents during feature description generation for feature 7\_7269 ``\textit{arxiv url path indicators}'' FT-Model+SAE, Closed Domain (GPT-4o). Added highlight tokens are highlighted in {\color{magenta} magenta}.}
  \label{fig:feature-examples}
\end{figure*}

\begin{figure*}
  \small
  \centering
  \begin{lstlisting}
We're studying neurons in a neural network. Each neuron activates on some particular word or concept in a short document.
You will be given a short explanation of what this neuron activates for, and then be shown (*@\textcolor{green!50!black}{\{group\_size\}}@*) example sequences.
For each sequence, the strongest activating words are indicated with <<...>> - note that these tags "<<" and ">>" are not part of the actual sentence, they're just used to highlight the places where the neuron is most active. 
Each of those documents has a different score: 0 means no neurons activate in the document, and (*@\textcolor{green!50!black}{\{max\_quantile\}}@*) means the neuron activates on the highlighted word or words in the document with the maximum strength.
You will have to return an estimated score for each sequence, in the form of a comma-separated list of (*@\textcolor{green!50!black}{\{group\_size\}}@*) integers from 0 to (*@\textcolor{green!50!black}{\{max\_quantile\}}@*), one for each sequence.
You should include nothing else in your response other than a list of comma-separated numbers, e.g. '1, 3, 7, ...'. Make sure not to include any reasoning or other text, just the plain response.

Here is the explanation for what the neuron should activate on: **(*@\textcolor{green!50!black}{\{explanation\}}@*)**.

Here are the examples:

1. (*@\textcolor{green!50!black}{\{example\_1\}}@*)
2. (*@\textcolor{green!50!black}{\{example\_2\}}@*)
(*@\textcolor{gray}{...}@*)
  \end{lstlisting}
  \caption{The full validation prompt for our autointerpretation pipeline. We sample groups of documents according to activation strength and then ask a large language model to predict the activation quantile of the SAE feature.}
  \label{fig:validation-prompt}
\end{figure*}

\begin{figure*}
  \small
  \centering
  \begin{lstlisting}
You are an expert text editor. You will be given a document and a description of a single stylistic or linguistic feature that appears in it.

Your task is to rewrite the document so that the described feature is removed entirely, while changing as little else as possible. You must preserve the original meaning, length, tone, topic, and every other stylistic feature of the text. Only the described feature should disappear; do not introduce new distinctive stylistic quirks in its place.

Feature to remove:
Summary: (*@\textcolor{green!50!black}{\{terse\}}@*)
Keywords: (*@\textcolor{green!50!black}{\{keywords\}}@*)
Detailed description: (*@\textcolor{green!50!black}{\{long\}}@*)

Document:
(*@\textcolor{green!50!black}{\{text\}}@*)

Return only the rewritten document, wrapped in <rewrite></rewrite> tags, with no additional commentary.
  \end{lstlisting}
  \caption{The prompt used in our paraphrase based causal ablation study (\S\ref{sec:causal_ablation}). Note that we use all three components of the feature's auto-interpretation description so as to give the model as much information as possible as to how to edit the document. We extract the paraphrase from the model response by parsing for the rewrite tags and retry if tags are missing or incomplete. We generate at temperature=1.0, top\_p=0.95, top\_k=64.}
  \label{fig:paraphrase-prompt}
\end{figure*}

\begin{figure*}
  \small
  \centering
  \begin{lstlisting}
You are an expert writer and editor. You will be given a document that was written by an AI language model.

Your task is to edit the document so that it reads as if it had been written by a human, while changing as few words as possible. Keep the original wording wherever you can and change only what is needed to make the text read as human-written; a good edit typically touches well under 20% of the words. Preserve the original meaning, topic, level of detail and length (about the same number of words).

You will be graded on two things: (1) how human-written the result reads, and (2) the percentage of tokens you swap relative to the original, which must stay low. Keep most of the text exactly the same: leave sentences that already read as human untouched, and inside the sentences you do edit, change only the words that need to change. Do not rewrite for style, do not shorten, do not restructure, and do not swap words that are not needed. Do not add commentary, disclaimers or any text about this task.

Document:
(*@\textcolor{green!50!black}{\{text\}}@*)

Reminder: keep most of the text exactly the same and change at most 15% of the words. Return only the rewritten document, wrapped in <rewrite></rewrite> tags, with no additional commentary.
  \end{lstlisting}
  \caption{The \textit{Baseline} (minimal-edit) paraphrase prompt used in our adversarial paraphrasing experiments (\S\ref{sec:causal_ablation}). The paraphraser is asked to make the document read as human-written while changing as few words as possible. We extract the paraphrase by parsing for the rewrite tags; failed responses are counted and excluded. All paraphrases are generated with gemini-3.1-flash-lite-preview at temperature=1.0, top\_p=0.95, top\_k=64.}
  \label{fig:adv-prompt-baseline}
\end{figure*}

\begin{figure*}
  \small
  \centering
  \begin{lstlisting}
You are an expert writer and editor. You will be given a document that was written by an AI language model, together with 3 passages written by people, taken from the same kind of source as the document.

### Example 1
(*@\textcolor{green!50!black}{\{example\_1\}}@*)

### Example 2
(*@\textcolor{green!50!black}{\{example\_2\}}@*)

### Example 3
(*@\textcolor{green!50!black}{\{example\_3\}}@*)

Your task is to edit the document so that it reads as if one of these authors had written it, while changing as few words as possible. Match the register, sentence rhythm, vocabulary and level of polish of these authors; do not copy their content, opinions or wording. Keep the original wording wherever you can and change only what is needed; a good edit typically touches well under 20% of the words. Preserve the original meaning, topic, level of detail and length (about the same number of words).

You will be graded on two things: (1) how human-written the result reads, and (2) the percentage of tokens you swap relative to the original, which must stay low. Keep most of the text exactly the same: leave sentences that already read as human untouched, and inside the sentences you do edit, change only the words that need to change. Do not rewrite for style, do not shorten, do not restructure, and do not swap words that are not needed. Do not add commentary, disclaimers or any text about this task.

Document:
(*@\textcolor{green!50!black}{\{text\}}@*)

Reminder: keep most of the text exactly the same and change at most 15% of the words. Return only the rewritten document, wrapped in <rewrite></rewrite> tags, with no additional commentary.
  \end{lstlisting}
  \caption{The \textit{Few Shot} paraphrase prompt. Three human-written documents from the training split are drawn per target document (same sub-source for COLING, same conference for PRDB, between 0.5$\times$ and 1.5$\times$ the target's length, at most 400/450 words; PRDB exemplars exclude papers whose reviews appear in the test split) and shown as register exemplars; the task and edit budget are otherwise identical to the Baseline prompt.}
  \label{fig:adv-prompt-fewshot}
\end{figure*}

\begin{figure*}
  \small
  \centering
  \begin{lstlisting}
You are an expert writer and editor. You will be given a document that was written by an AI language model.

Your task is to edit the document so that it reads as if it had been written by a human, while changing as few words as possible. Keep the original wording wherever you can and change only what is needed to make the text read as human-written; a good edit typically touches well under 20% of the words. Preserve the original meaning, topic, level of detail and length (about the same number of words).

Follow these rules, which remove the habits that give machine-written text away:
1. Do not open a sentence with a transition word or phrase (Furthermore, Additionally, Moreover, Ultimately, Overall, Finally, In conclusion, Notably, Importantly).
2. Do not end with a sentence that summarises, generalises or looks ahead.
3. Vary sentence length deliberately: mix very short sentences with long ones, and never write several sentences of similar length in a row.
4. Avoid stock vocabulary such as delve, crucial, vital, essential, pivotal, landscape, comprehensive, robust, leverage, foster, navigate, testament, tapestry, underscore, showcase, seamless, multifaceted.
5. Do not introduce lists of examples with "such as" or "for instance", and do not write three-item parallel lists.
6. Do not add headers, labels, bullet points or paragraph breaks that the original does not have.
Contractions, first person, hedges, asides and plain everyday words are welcome.

You will be graded on two things: (1) how human-written the result reads, and (2) the percentage of tokens you swap relative to the original, which must stay low. Keep most of the text exactly the same: leave sentences that already read as human untouched, and inside the sentences you do edit, change only the words that need to change. Do not rewrite for style, do not shorten, do not restructure, and do not swap words that are not needed. Do not add commentary, disclaimers or any text about this task.

Document:
(*@\textcolor{green!50!black}{\{text\}}@*)

Reminder: keep most of the text exactly the same and change at most 15% of the words. Return only the rewritten document, wrapped in <rewrite></rewrite> tags, with no additional commentary.
  \end{lstlisting}
  \caption{The \textit{Guided} paraphrase prompt. The Baseline prompt is extended with six generic rules against machine-writing habits; the rules were written from general knowledge of LLM prose and do not depend on any detector or on our model outputs.}
  \label{fig:adv-prompt-guided}
\end{figure*}

\begin{figure*}
  \small
  \centering
  \begin{lstlisting}
You are an expert writer and editor helping to test how robust an AI-generated-text detector is. You will be given a document written by an AI language model, together with a precise account of how one specific detector reads it.

The detector reads the document with a language model and scores it with a linear classifier over interpretable features. Each feature has a short automatically generated description of the kind of text it responds to, and below you are shown the exact passages of this document where it fires. Every feature the detector finds either raises or lowers its "AI" score by the amount shown, and the verdict is the sum of those contributions.

Counting rule: the detector keeps only the STRONGEST occurrence of each feature. A feature stops counting only when every passage where it fires has been rewritten so the feature no longer applies; rewriting some of its passages and leaving others earns nothing.

## Your target
The detector's score for a document is the sum of the feature contributions below plus a fixed constant, and it labels the document "AI" when the score is above its decision boundary. In this document it sees (*@\textcolor{green!50!black}{\{ai\_mass\}}@*) of AI evidence and (*@\textcolor{green!50!black}{\{human\_mass\}}@*) of human evidence, and the score currently stands (*@\textcolor{green!50!black}{\{gap\}}@*) above the boundary. To flip the verdict with a comfortable margin you must lower the score by at least (*@\textcolor{green!50!black}{\{target\}}@*): remove AI evidence, add human evidence, or both. Removing the listed AI features completely is the most reliable way to get there.

(*@\textcolor{green!50!black}{\{evidence\}}@*)

## Rules
- For every AI feature you remove, rewrite ALL of its marked passages (see the counting rule). Start with the largest contributions.
- Change as few words as possible: edit the marked passages and what a substitution needs, and leave every other sentence exactly as it was. A good edit typically touches well under 20% of the words.
- Preserve the original meaning, topic, level of detail and length (about the same number of words); the result must read as natural, fluent text.
- You are graded on two things: how far the detector's score moves toward "human", and the percentage of tokens you swap, which must stay low.
- Do not add commentary, disclaimers or any text about this task.

## How to answer
First, inside <plan></plan> tags, write a short plan: for each feature you target, list its marked passages and the exact replacement wording you will use, then add up the contributions you expect to remove and check that the total reaches the target. Then write the rewritten document inside <rewrite></rewrite> tags.

Document:
(*@\textcolor{green!50!black}{\{text\}}@*)

Reminder: lower the score by at least (*@\textcolor{green!50!black}{\{target\}}@*) by rewriting every marked passage of the AI features you target, while changing at most 15-20% of the words. Return your plan inside <plan></plan> tags followed by the rewritten document inside <rewrite></rewrite> tags, with no other text.
  \end{lstlisting}
  \caption{The \textit{SAE} (feature-guided) paraphrase prompt. The {\color{green!50!black}\{evidence\}} block (Figure~\ref{fig:adv-prompt-sae-evidence}) lists the detector's own features for the document; the numeric target is the score's distance to the decision threshold plus a margin of 3, and the counting rule describes the detector's max-pooling. The model first writes a plan inside plan tags, then the rewrite. In the $\times 3$ condition the same prompt is issued up to three times, each time with the evidence recomputed on the previous rewrite, stopping once the document is classified human.}
  \label{fig:adv-prompt-sae}
\end{figure*}

\begin{figure*}
  \small
  \centering
  \begin{lstlisting}
## AI evidence the detector found in THIS document (largest first)
Each entry gives the feature, its contribution, and the passages where it fires, with the firing words marked as (*@\textcolor{magenta}{[[}@*)...(*@\textcolor{magenta}{]]}@*).
- (*@\textcolor{green!50!black}{\{description\}}@*) (e.g., (*@\textcolor{green!50!black}{\{keywords\}}@*)) [effect on the AI score: (*@\textcolor{green!50!black}{\{effect\}}@*)]
    fires at: ...(*@\textcolor{green!50!black}{\{before\}}@*)(*@\textcolor{magenta}{[[}@*)(*@\textcolor{green!50!black}{\{core\}}@*)(*@\textcolor{magenta}{]]}@*)(*@\textcolor{green!50!black}{\{after\}}@*)...
    fires at: ...(*@\textcolor{green!50!black}{\{before\}}@*)(*@\textcolor{magenta}{[[}@*)(*@\textcolor{green!50!black}{\{core\}}@*)(*@\textcolor{magenta}{]]}@*)(*@\textcolor{green!50!black}{\{after\}}@*)...
    fires at: ...(*@\textcolor{green!50!black}{\{before\}}@*)(*@\textcolor{magenta}{[[}@*)(*@\textcolor{green!50!black}{\{core\}}@*)(*@\textcolor{magenta}{]]}@*)(*@\textcolor{green!50!black}{\{after\}}@*)...
    (+(*@\textcolor{green!50!black}{\{n\_more\}}@*) more passages like these)
(*@\textcolor{gray}{...}@*)

## Human evidence already in THIS document (keep it; strengthening it is fine)
- (*@\textcolor{green!50!black}{\{description\}}@*) (e.g., (*@\textcolor{green!50!black}{\{keywords\}}@*)) [effect on the AI score: (*@\textcolor{green!50!black}{\{effect\}}@*)]
    fires at: ...(*@\textcolor{green!50!black}{\{before\}}@*)(*@\textcolor{magenta}{[[}@*)(*@\textcolor{green!50!black}{\{core\}}@*)(*@\textcolor{magenta}{]]}@*)(*@\textcolor{green!50!black}{\{after\}}@*)...
    fires at: ...(*@\textcolor{green!50!black}{\{before\}}@*)(*@\textcolor{magenta}{[[}@*)(*@\textcolor{green!50!black}{\{core\}}@*)(*@\textcolor{magenta}{]]}@*)(*@\textcolor{green!50!black}{\{after\}}@*)...
    fires at: ...(*@\textcolor{green!50!black}{\{before\}}@*)(*@\textcolor{magenta}{[[}@*)(*@\textcolor{green!50!black}{\{core\}}@*)(*@\textcolor{magenta}{]]}@*)(*@\textcolor{green!50!black}{\{after\}}@*)...
    (+(*@\textcolor{green!50!black}{\{n\_more\}}@*) more passages like these)
(*@\textcolor{gray}{...}@*)
  \end{lstlisting}
  \caption{The evidence block of the SAE prompt (Figure~\ref{fig:adv-prompt-sae}). For each of the detector's top-10 AI-direction and top-10 human-direction features in the document we list its terse auto-interpretation description and keywords, its contribution to the score, and up to three passages where it fires most strongly, with the firing tokens marked by {\color{magenta}[[ ]]} highlight tokens (six tokens of context on either side).}
  \label{fig:adv-prompt-sae-evidence}
\end{figure*}

\begin{table*}[t]
\centering
\small
\setlength{\tabcolsep}{2.1pt}
\begin{tabular}{lccccccccccccccc}
\toprule
& \multicolumn{3}{c}{\textbf{GPT-4o}} & \multicolumn{3}{c}{\textbf{Llama 3.1 70B}} & \multicolumn{3}{c}{\textbf{Qwen 2.5 72B}} & \multicolumn{3}{c}{\textbf{Claude 3.5 Sonnet}} & \multicolumn{3}{c}{\textbf{Gemini 1.5 Pro}} \\
\cmidrule(lr){2-4} \cmidrule(lr){5-7} \cmidrule(lr){8-10} \cmidrule(lr){11-13} \cmidrule(lr){14-16}
 & AUC & F1 & TPR@$5$ & AUC & F1 & TPR@$5$ & AUC & F1 & TPR@$5$ & AUC & F1 & TPR@$5$ & AUC & F1 & TPR@$5$ \\
\midrule
\multicolumn{16}{l}{\textit{Trained on GPT-4o}} \\[2pt]
\textbf{Baselines} & & & & & & & & & & & & & & & \\
gram2vec  & 100 & 99.6 & 100 & 97.9 & 93.5 & 90.0 & 99.7 & 98.6 & 99.9 & 99.7 & 98.2 & 99.7 & 99.1 & 96.6 & 98.1 \\
hf-gemma2 & 100 & 100  & 100 & 97.4 & 92.3 & 89.9 & 100  & 99.6 & 100  & 99.2 & 96.3 & 96.9 & 99.1 & 96.1 & 96.5 \\
\midrule
\textbf{w/o Mask} & & & & & & & & & & & & & & & \\
\textsc{Base-SAE}    & 100 & 99.8 & 100 & 99.5 & 98.6 & 99.9 & 99.9 & 99.7 & 100 & 99.6 & 98.6 & 100  & 99.6 & 98.4 & 99.8 \\
\textsc{FT-SAE}      & 100 & 99.9 & 100 & 99.6 & 99.1 & 100  & 99.9 & 99.8 & 100 & 99.7 & 99.0 & 100  & 99.8 & 99.0 & 100  \\
\textsc{FT-Model+SAE}& 100 & 99.9 & 100 & 97.2 & 91.3 & 82.9 & 99.9 & 99.0 & 99.7 & 92.1 & 85.6 & 52.6 & 93.4 & 85.3 & 69.9 \\
\midrule
\textbf{w/ Mask ($\tau\!>\!0.7$)} & & & & & & & & & & & & & & & \\
\textsc{Base-SAE}    & 100 & 99.8 & 100 & 99.2 & 97.3 & 99.2 & 99.9 & 99.5 & 100 & 99.4 & 97.7 & 99.6 & 99.4 & 97.2 & 99.0 \\
\textsc{FT-SAE}      & 100 & 99.9 & 100 & 99.6 & 98.6 & 99.9 & 99.9 & 99.7 & 100 & 99.7 & 98.7 & 99.9 & 99.7 & 98.7 & 99.9 \\
\textsc{FT-Model+SAE}& 100 & 99.9 & 100 & 97.9 & 92.9 & 87.9 & 100  & 99.2 & 99.9 & 95.7 & 89.2 & 73.9 & 96.2 & 88.8 & 79.8 \\
\specialrule{\heavyrulewidth}{3pt}{3pt}
\multicolumn{16}{l}{\textit{Trained on LLaMA}} \\[2pt]
\textbf{Baselines} & & & & & & & & & & & & & & & \\
gram2vec  & 97.4 & 93.5 & 85.5 & 100 & 99.4 & 100 & 99.5 & 97.5 & 99.3 & 95.2 & 89.4 & 70.9 & 96.7 & 92.2 & 79.1 \\
hf-gemma2 & 97.0 & 89.7 & 85.4 & 100 & 100  & 100 & 99.7 & 98.1 & 99.0 & 98.7 & 95.7 & 96.1 & 83.3 & 80.0 & 33.9 \\
\midrule
\textbf{w/o Mask} & & & & & & & & & & & & & & & \\
\textsc{Base-SAE}    & 99.6 & 98.2 & 99.9 & 100 & 99.8 & 100 & 100  & 99.7 & 100  & 99.0 & 96.5 & 97.8 & 98.5 & 95.2 & 93.9 \\
\textsc{FT-SAE}      & 99.6 & 98.6 & 100  & 100 & 99.9 & 100 & 100  & 99.9 & 100  & 99.6 & 98.1 & 99.8 & 99.2 & 96.9 & 98.6 \\
\textsc{FT-Model+SAE}& 94.9 & 87.6 & 69.6 & 100 & 99.9 & 100 & 99.4 & 96.1 & 97.1 & 91.0 & 84.2 & 48.0 & 84.8 & 78.7 & 35.6 \\
\midrule
\textbf{w/ Mask ($\tau\!>\!0.7$)} & & & & & & & & & & & & & & & \\
\textsc{Base-SAE}    & 99.1 & 96.4 & 97.6 & 100 & 99.8 & 100 & 99.9 & 98.9 & 100  & 99.2 & 96.3 & 97.5 & 98.0 & 93.8 & 89.5 \\
\textsc{FT-SAE}      & 99.5 & 97.7 & 99.6 & 100 & 99.9 & 100 & 100  & 99.9 & 100  & 99.3 & 96.7 & 98.2 & 98.7 & 95.1 & 94.0 \\
\textsc{FT-Model+SAE}& 93.7 & 85.9 & 62.7 & 100 & 99.9 & 100 & 99.3 & 95.6 & 96.2 & 90.8 & 84.9 & 43.1 & 82.8 & 77.6 & 30.0 \\
\midrule
\multicolumn{16}{l}{\textit{Off-the-shelf}} \\[2pt]
Binoculars$^\dagger$     & 95.7 & 89.8 & 78.9 & 99.9 & 99.0 & 99.7 & 99.9 & 99.5 & 100  & 98.3 & 94.7 & 93.8 & 98.9 & 95.9 & 96.6 \\
SpecDetect$^\dagger$     & 92.2 & 84.4 & 62.1 & 98.4 & 93.8 & 92.8 & 99.8 & 98.3 & 99.5 & 94.7 & 87.8 & 73.7 & 96.7 & 90.8 & 83.5 \\
ReMoDetect$^\dagger$     & 99.4 & 98.4 & 100  & 99.8 & 99.0 & 99.9 & 99.7 & 98.8 & 100  & 99.3 & 97.3 & 99.0 & 99.4 & 98.1 & 99.9 \\
Fast-DetectGPT$^\dagger$ & 81.0 & 74.1 & 30.5 & 99.3 & 98.6 & 99.1 & 99.4 & 99.3 & 99.9 & 94.0 & 87.7 & 62.9 & 97.8 & 93.7 & 87.5 \\
\bottomrule
\end{tabular}
\caption{Model performance across LLM-generated text targets in the PRDB dataset, for detectors trained on GPT-4o- and LLaMA-generated text respectively. Each column group corresponds to text generated by a different source model. $\dagger$ indicates off-the-shelf models that were not trained on any in-domain data.}
\label{tab:prdb_results_full}
\end{table*}

\begin{table*}[t]
\centering
\small
\setlength{\tabcolsep}{4pt}
\renewcommand{\arraystretch}{1.2}
\begin{tabular}{@{}lp{12.2cm}rr@{}}
\toprule
\multicolumn{4}{@{}c}{\textbf{Most Predictive Features: \textsc{FT-SAE} $\tau > 0.85$ --- General Domain Text (COLING)}}\\
\midrule
\textbf{Feature} & \textbf{Concept Description} \textit{(auto-interp: gemini-3.1-flash-lite)} & \textbf{Corr.} & \textbf{Logit $\Delta$}\\
\midrule
\multicolumn{4}{@{}l}{\textbf{AI Features}}\\
13\_3738  & introductory phrases or filler text commonly used by ai language models when declining personal questions (\emph{e.g.,} ``As an'', ``AI language model'', ``I do not have'') & 0.859 & $+0.987$ \\
13\_1545  & academic and scientific research methodology and technical analysis terminology (\emph{e.g.,} ``dynamics'', ``analytical'', ``methodology'') & 0.876 & $+0.909$ \\
17\_12178 & critical commentary or argumentative analysis of student policies and social responsibility (\emph{e.g.,} ``have'', ``effects'', ``motivated by'', ``can'', ``unfair'') & 0.892 & $+0.558$ \\
17\_8638  & neuron activates on random text fragments and noisy character sequences often found in garbled or fragmented data (\emph{e.g.,} ``astion'', ``Mol'', ``a priori'') & 0.867 & $+0.372$ \\
17\_1175  & repeated phrases or sequences in text (\emph{e.g.,} ``zero numbers are called'', ``long story'', ``of a Hamiltonian'') & 0.852 & $+0.363$ \\
13\_2153  & exclamation points used at the end of emotional or emphatic sentences (\emph{e.g.,} \texttt{!}, \texttt{!}, \texttt{!}) & 0.879 & $+0.318$ \\
22\_9579  & markdown formatting for headers or list items (\emph{e.g.,} \texttt{\#\#}, \texttt{\#\#\#}, \texttt{\textbackslash{}n\textbackslash{}n}) & 0.850 & $+0.297$ \\
7\_1076   & conversational filler phrases used in informal speech (\emph{e.g.,} ``like'', ``a really good'', ``aren't'') & 0.870 & $+0.261$ \\
7\_7991   & reads more or similar web page navigation links (\emph{e.g.,} ``Read more »'', ``more »'', ``more'') & 0.868 & $+0.239$ \\
17\_4337  & introductory transition words signaling a summary or final thought (\emph{e.g.,} ``In short,'', ``conclusion,'', ``Ultimately,'') & 0.860 & $+0.210$ \\
17\_14082 & word such when used to introduce examples or comparisons (\emph{e.g.,} ``such'', ``such as'') & 0.866 & $+0.192$ \\
22\_8299  & repetitive sequences of short functional words or repeated phrases (\emph{e.g.,} ``were grey'', ``the first'', ``computing power'') & 0.878 & $+0.179$ \\
17\_13162 & repetitive patterns or repeated words within a document (\emph{e.g.,} ``combo'', ``robust'', ``Materialism'') & 0.859 & $+0.178$ \\
22\_3891  & transitions used to summarize or conclude a text (\emph{e.g.,} ``conclusion'', ``summary'', ``overall'') & 0.854 & $+0.173$ \\
22\_232   & instructions on how to perform or accomplish specific tasks or processes (\emph{e.g.,} ``how to'', ``make a'', ``do'') & 0.885 & $+0.158$ \\
17\_546   & the word about used as a preposition (\emph{e.g.,} ``about'', ``surrounding'', ``regarding'') & 0.851 & $+0.149$ \\
13\_15327 & words or concepts associated with personal growth, achievement, societal contribution, and nuanced life experiences (\emph{e.g.,} ``addictive'', ``activities'', ``achievement'') & 0.885 & $+0.146$ \\
7\_6260   & wikipedia category label syntax (\emph{e.g.,} ``Category:'', ``Category'', ``category'') & 0.884 & $+0.146$ \\
\midrule
\multicolumn{4}{@{}l}{\textbf{Human Features}}\\
7\_14967  & comma punctuation following a newline or preceding sentence start (\emph{e.g.,} \texttt{,}, \texttt{, }) & 0.853 & $-1.563$ \\
7\_10954  & neuron fires on the digit 0 inside url placeholders like url 0 (\emph{e.g.,} ``URL\_0'', \texttt{0}) & 0.854 & $-0.951$ \\
7\_9535   & newline characters in scientific or technical text (\emph{e.g.,} \texttt{\textbackslash{}n}, \texttt{\textbackslash{}n\textbackslash{}n}) & 0.863 & $-0.842$ \\
7\_1262   & contractions containing the letter n and apostrophe specifically focusing on the n't ending (\emph{e.g.,} ``n'{}'', ``n't'', ``n't .'') & 0.851 & $-0.785$ \\
7\_5896   & newline characters in scientific or technical text (\emph{e.g.,} \texttt{\textbackslash{}n}, \texttt{\textbackslash{}n}, \texttt{\textbackslash{}n}) & 0.858 & $-0.658$ \\
13\_11151 & newline characters in mathematical or technical contexts (\emph{e.g.,} \texttt{\textbackslash{}n}, \texttt{\textbackslash{}n}, \texttt{\textbackslash{}n}) & 0.859 & $-0.473$ \\
7\_14666  & wikipedia category and navigation section headers (\emph{e.g.,} ``Film'', ``Television'', ``External'') & 0.898 & $-0.454$ \\
22\_3894  & semicolon followed by a newline character (\emph{e.g.,} \texttt{;\textbackslash{}n}) & 0.871 & $-0.394$ \\
17\_15548 & opening double quotation marks following a closing single quote and comma (\emph{e.g.,} \texttt{', }) & 0.855 & $-0.374$ \\
13\_15646 & markdown editing and formatting indicators (\emph{e.g.,} ``EDIT :'', \texttt{\textbackslash{}\textbackslash{}*}, ``( URL'') & 0.850 & $-0.366$ \\
13\_8665  & newline characters and line breaks within technical or academic documents (\emph{e.g.,} \texttt{\textbackslash{}n}, \texttt{\textbackslash{}n\textbackslash{}n}) & 0.855 & $-0.350$ \\
7\_15058  & wikipedia category and list formatting specifically numerical prefixes and lists (\emph{e.g.,} \texttt{21}, \texttt{1986}, \texttt{1929}) & 0.882 & $-0.344$ \\
22\_1014  & the symbol sequence -> commonly used for arrows or transition markers & 0.876 & $-0.289$ \\
22\_1117  & biographical section headers and metadata labels (\emph{e.g.,} ``Biography'', ``History'', ``Life'') & 0.867 & $-0.281$ \\
7\_9241   & contraction n suffix of dont doesnt isnt etc (\emph{e.g.,} ``n't'', ``n'', ``'t'') & 0.866 & $-0.279$ \\
13\_1665  & reddit subculture acronyms and abbreviations like cmv or eli5 (\emph{e.g.,} ``CMV'', ``CMV!'', ``ELI5'') & 0.866 & $-0.269$ \\
7\_4790   & contractions containing the letter n or apostrophes followed by t (\emph{e.g.,} ``wo'', ``n't'', ``ca'') & 0.871 & $-0.247$ \\
17\_8967  & words or phrases involving abstract concepts or situational descriptions in formal or narrative text (\emph{e.g.,} ``closer'', ``Death itself'', ``idered'') & 0.860 & $-0.189$ \\
\bottomrule
\end{tabular}
\caption{Top features of the \textsc{FT-SAE} $\tau > 0.85$ logistic-regression detector on the COLING dataset \cite{wang-etal-2025-genai}, ranked by mean logit contribution (Logit $\Delta$) along with their auto-interpretation simulation score (\S\ref{sec:autointerpretation}).}
\label{tab:top_features_coling_ftsae}
\end{table*}

\begin{table*}[t]
\centering
\small
\setlength{\tabcolsep}{4pt}
\renewcommand{\arraystretch}{1.2}
\begin{tabular}{@{}lp{12.2cm}rr@{}}
\toprule
\multicolumn{4}{@{}c}{\textbf{Most Predictive Features: \textsc{FT-Model+SAE} $\tau > 0.85$ --- General Domain Text (COLING)}}\\
\midrule
\textbf{Feature} & \textbf{Concept Description} \textit{(auto-interp: gemini-3.1-flash-lite)} & \textbf{Corr.} & \textbf{Logit $\Delta$}\\
\midrule
\multicolumn{4}{@{}l}{\textbf{AI Features}}\\
7\_8935   & the word the when appearing frequently & 0.865 & $+0.865$ \\
7\_2327   & the preposition on & 0.855 & $+0.860$ \\
7\_7227   & actions and physical reactions of characters in narrative prose (\emph{e.g.,} ``laughed nervously'', ``shook his head'', ``stood'') & 0.887 & $+0.796$ \\
13\_4399  & imperative instructional verbs in written guides or manuals (\emph{e.g.,} ``Remember'', ``Perform'', ``Write'') & 0.861 & $+0.642$ \\
13\_501   & academic and technical publication identifiers including urls and scholarly citation elements (\emph{e.g.,} ``preprint'', ``arxiv'', ``https'') & 0.908 & $+0.633$ \\
7\_870    & words indicating personal perspective experience or generalized statements of fact (\emph{e.g.,} ``assist'', ``suggests'', ``experience'') & 0.860 & $+0.522$ \\
7\_13576  & activates on the word article or articles often in titles or references & 0.876 & $+0.522$ \\
7\_6058   & the phrase such as & 0.878 & $+0.491$ \\
17\_10611 & conversational filler words and discourse markers like like and so that (\emph{e.g.,} ``like'', ``so that'', ``you know'') & 0.870 & $+0.463$ \\
7\_365    & introductory transition words or labels used to set up a logical argument or point (\emph{e.g.,} ``Counter Argument:'', ``Counter-argument:'', ``defecting'') & 0.865 & $+0.416$ \\
7\_4073   & math and physics formula notation specifically subscripts and bracketed terms in technical equations (\emph{e.g.,} ``eq:'', ``v0'', \texttt{12}) & 0.853 & $+0.382$ \\
13\_2854  & academic phrasing regarding new methods and technical approaches (\emph{e.g.,} ``a novel'', ``approach'', ``that'') & 0.862 & $+0.329$ \\
22\_200   & academic and scientific analysis terminology found in research papers (\emph{e.g.,} ``simulations'', ``analyze'', ``observations'') & 0.871 & $+0.317$ \\
22\_6780  & the word including or similar participle forms used to introduce examples or lists (\emph{e.g.,} ``including'', ``ranging'', ``inclusive'') & 0.851 & $+0.304$ \\
13\_9703  & the word such in various contexts & 0.866 & $+0.301$ \\
7\_3300   & academic and scientific research terminology especially in physics and cosmology (\emph{e.g.,} ``the'', ``models'', ``physics'') & 0.889 & $+0.273$ \\
\midrule
\multicolumn{4}{@{}l}{\textbf{Human Features}}\\
17\_13923 & text sequences often involving personal narratives bracketed punctuation or specific conversational phrases (\emph{e.g.,} ``grave'', ``closing'', ``message us'') & 0.876 & $-1.828$ \\
13\_543   & the phrase such as used to introduce examples & 0.863 & $-1.231$ \\
7\_6678   & scientific research vocabulary and terms related to analysis methodology or physical phenomena (\emph{e.g.,} ``explore'', ``subset'', ``plasmonics'') & 0.886 & $-0.884$ \\
17\_3367  & word article when referring to specific texts or media content (\emph{e.g.,} ``article'', ``article.'') & 0.851 & $-0.823$ \\
7\_14434  & commas occurring at the beginning of a line or after a newline character (\emph{e.g.,} \texttt{,}) & 0.855 & $-0.724$ \\
13\_2197  & football and tennis match commentary names and actions (\emph{e.g.,} ``Moussa Marega'', ``Plymouth Argyle'', ``Rodionova'') & 0.875 & $-0.596$ \\
7\_11800  & asterisk characters used for emphasis or formatting in markdown (\emph{e.g.,} \texttt{*}, \texttt{**}, \texttt{* *}) & 0.860 & $-0.562$ \\
7\_7417   & academic peer review feedback context (\emph{e.g.,} ``reviewers'', ``commenters'', ``feedback'') & 0.858 & $-0.548$ \\
13\_13833 & metadata or list item identifiers in wikipedia or reference sections like dates years or categories (\emph{e.g.,} \texttt{2003}, \texttt{1973}, ``References'') & 0.899 & $-0.515$ \\
7\_12299  & wikipedia category tags and structural metadata markers (\emph{e.g.,} \texttt{<{}<}, \texttt{>{}>}, ``1954 births'') & 0.856 & $-0.489$ \\
17\_12622 & markdown url placeholders in brackets (\emph{e.g.,} ``URL'', \texttt{\_0}, \texttt{\_5}) & 0.861 & $-0.482$ \\
7\_14956  & markdown formatting syntax and URL related punctuation like brackets underscores and parentheses (\emph{e.g.,} ``(\_URL\_'', \texttt{<{}<}, \texttt{>{}>}) & 0.853 & $-0.452$ \\
13\_6212  & pinyin or foreign language word sub-components within parentheses or following labels (\emph{e.g.,} ``inyin'', ``Tibetan'', ``pinyin:'', ``freguesia'') & 0.876 & $-0.414$ \\
13\_13054 & words ending in logist or surgeon related to medical professions (\emph{e.g.,} ``logist'', ``surgeon'', ``ped'') & 0.851 & $-0.274$ \\
17\_253   & narrative fiction character actions and pronouns (\emph{e.g.,} ``her'', ``and'', ``the'') & 0.884 & $-0.213$ \\
7\_5164   & acronyms and technical abbreviations in parentheses (\emph{e.g.,} ``CD'', ``DIS'', ``OBIE'') & 0.855 & $-0.204$ \\
22\_3509  & wikipedia list and reference section headers or index entries (\emph{e.g.,} \texttt{1895}, ``2010s'', ``See also'') & 0.878 & $-0.201$ \\
7\_16036  & newline characters that occur after a semicolon or punctuation mark (\emph{e.g.,} \texttt{\textbackslash{}n}, \texttt{;}, \texttt{.}) & 0.866 & $-0.198$ \\
\bottomrule
\end{tabular}
\caption{Top features of the \textsc{FT-Model+SAE} $\tau > 0.85$ logistic-regression detector on the COLING dataset \cite{wang-etal-2025-genai}, ranked by mean logit contribution (Logit $\Delta$) along with their auto-interpretation score (\S\ref{sec:autointerpretation}).}
\label{tab:top_features_coling_ftmodelsae}
\end{table*}

\begin{table*}[t]
\centering
\small
\setlength{\tabcolsep}{4pt}
\renewcommand{\arraystretch}{1.2}
\begin{tabular}{@{}lp{12.2cm}rr@{}}
\toprule
\multicolumn{4}{@{}c}{\textbf{Most Predictive Features: \textsc{Base-SAE} $\tau > 0.85$ --- General Domain Text (COLING)}}\\
\midrule
\textbf{Feature} & \textbf{Concept Description} \textit{(auto-interp: gemini-3.1-flash-lite)} & \textbf{Corr.} & \textbf{Logit $\Delta$}\\
\midrule
\multicolumn{4}{@{}l}{\textbf{AI Features}}\\
13\_1157  & code syntax punctuation like brackets braces and delimiters (\emph{e.g.,} \texttt{\}}, \texttt{)}, \texttt{]}) & 0.863 & $+0.604$ \\
17\_941   & html input tag type attribute syntax (\emph{e.g.,} ``type'', \texttt{type=}, \texttt{type="}) & 0.859 & $+0.558$ \\
13\_1449  & conclusion or concluding remarks headers (\emph{e.g.,} ``Conclusion'', ``conclusion'', ``conclusion.'') & 0.864 & $+0.543$ \\
22\_6940  & software license and copyright disclaimers in source code headers (\emph{e.g.,} ``licensed under'', ``WITHOUT ANY WARRANTY'', ``Apache 2.0'') & 0.903 & $+0.522$ \\
13\_16123 & words related to software development or technical documentation topics (\emph{e.g.,} ``Node'', ``browser'', ``functions'') & 0.871 & $+0.500$ \\
13\_16277 & repetitive sequences of yes or technical/descriptive text strings (\emph{e.g.,} ``Yes'', ``management'', ``the'') & 0.854 & $+0.478$ \\
17\_606   & front end web development framework keywords and comma separated lists of features (\emph{e.g.,} ``Vue'', ``components'', ``routing'') & 0.862 & $+0.395$ \\
22\_7339  & php and sql code syntax symbols (\emph{e.g.,} \texttt{<?php}, \texttt{?>}, \texttt{:}) & 0.878 & $+0.364$ \\
22\_5450  & words and phrases in celebrity or entertainment news segments (\emph{e.g.,} ``Love Island'', ``Kanye'', ``basketball'') & 0.868 & $+0.329$ \\
7\_3855   & words related to organizational strategies and social improvement initiatives (\emph{e.g.,} ``inclusive'', ``improving'', ``solutions'') & 0.876 & $+0.305$ \\
22\_13083 & introductory phrases in how-to guides and articles (\emph{e.g.,} ``In this article'', ``here's how'', ``following these'') & 0.860 & $+0.287$ \\
17\_3506  & numbers within square brackets indicating academic citations (\emph{e.g.,} \texttt{[1]}, \texttt{[10]}, \texttt{[1920]}) & 0.870 & $+0.286$ \\
13\_11434 & two consecutive colons used in text formatting or data categorization (\emph{e.g.,} \texttt{::}, \texttt{..}, \texttt{0}) & 0.854 & $+0.269$ \\
17\_3111  & code and math syntax formatting tokens including braces brackets backslashes and punctuation (\emph{e.g.,} \texttt{\}}, \texttt{!}, \texttt{\textbackslash{}n}, \texttt{\textbackslash{}\textbackslash{}}, \texttt{>{}>}) & 0.875 & $+0.256$ \\
22\_14956 & numbered list items in a guide or tutorial (\emph{e.g.,} \texttt{1.}, \texttt{2.}, \texttt{5.}) & 0.875 & $+0.219$ \\
\midrule
\multicolumn{4}{@{}l}{\textbf{Human Features}}\\
7\_9548   & chinese punctuation and specific formatting characters like ideographic commas and periods & 0.856 & $-1.171$ \\
13\_12631 & random or scrambled alphanumeric character strings and non-standard punctuation often found in corrupted text or obfuscated data (\emph{e.g.,} ``4u!).'', ``jl22'', `` e a - tim e e s;; a a i re'') & 0.876 & $-0.605$ \\
17\_8723  & code syntax comments or structural markers (\emph{e.g.,} \texttt{/*}, \texttt{*/}, \texttt{<{}<}) & 0.863 & $-0.557$ \\
22\_2286  & markdown headers and navigational labels in wiki style articles (\emph{e.g.,} ``See also'', ``References'', ``Gallery'') & 0.855 & $-0.483$ \\
22\_1640  & url placeholder tokens (\emph{e.g.,} ``URL\_'', ``URL\_0'', ``URL\_1'') & 0.855 & $-0.434$ \\
17\_741   & mathematical notation symbols and labels in scientific or technical contexts (\emph{e.g.,} \texttt{->}, \texttt{)}, ``with'', ``B3'') & 0.858 & $-0.430$ \\
22\_6877  & math and logic symbols inside latex expressions or technical parentheses (\emph{e.g.,} \texttt{>{}>}, ``infty'', \texttt{)}, \texttt{\}}) & 0.863 & $-0.358$ \\
13\_7218  & markdown header references and related newline or parenthetical formatting (\emph{e.g.,} ``References'', \texttt{\textbackslash{}n}, \texttt{(}) & 0.877 & $-0.252$ \\
17\_11950 & multiple newline characters representing large paragraph breaks (\emph{e.g.,} \texttt{\textbackslash{}n\textbackslash{}n\textbackslash{}n}, \texttt{\textbackslash{}n\textbackslash{}n\textbackslash{}n\textbackslash{}n}, \texttt{\textbackslash{}n\textbackslash{}n\textbackslash{}n\textbackslash{}n\textbackslash{}n\textbackslash{}n\textbackslash{}n\textbackslash{}n\textbackslash{}n}) & 0.893 & $-0.234$ \\
13\_4706  & conversational meta-commentary phrases like off topic or similar topic (\emph{e.g.,} ``it makes you happy'', ``another similar topic'', ``off topic'') & 0.864 & $-0.229$ \\
7\_1600   & mentions of tv tropes or specific trope related wiki formatting & 0.877 & $-0.220$ \\
13\_12992 & insulting or dismissive colloquial language directed at a person or their ideas (\emph{e.g.,} ``shitload'', ``to your ignorance'', ``wretched place'') & 0.864 & $-0.219$ \\
7\_5065   & url placeholder or link references & 0.883 & $-0.212$ \\
13\_15434 & forward slashes occurring in markdown or syntax contexts (\emph{e.g.,} \texttt{/}, \texttt{/>}) & 0.855 & $-0.194$ \\
17\_5250  & words indicating service quality or restaurant experiences including staff interactions and feelings about service (\emph{e.g.,} ``service'', ``technician'', ``servers'') & 0.856 & $-0.170$ \\
17\_10096 & french verbs or words containing substrings of common french conjugation or inflection endings (\emph{e.g.,} ``interag'', ``parl'', ``achie'') & 0.894 & $-0.170$ \\
7\_11769  & repeated capitalized or emphatic adverbs used for emphasis (\emph{e.g.,} ``many'', ``EVERY'', ``really, REALLY'') & 0.866 & $-0.169$ \\
17\_1279  & words or phrases commonly found in customer reviews or informal first-person narratives (\emph{e.g.,} ``chicken'', ``unable to find'', ``worth it'') & 0.872 & $-0.169$ \\
\bottomrule
\end{tabular}
\caption{Top features of the \textsc{Base-SAE} $\tau > 0.85$ logistic-regression detector on the COLING dataset \cite{wang-etal-2025-genai}, ranked by mean logit contribution (Logit $\Delta$) along with their auto-interpretation simulation score (\S\ref{sec:autointerpretation}).}
\label{tab:top_features_coling_basesae}
\end{table*}

\begin{table*}[t]
\centering
\small
\setlength{\tabcolsep}{4pt}
\renewcommand{\arraystretch}{1.2}
\begin{tabular}{@{}lp{12.2cm}rr@{}}
\toprule
\multicolumn{4}{@{}c}{\textbf{Most Predictive Features: \textsc{Base-SAE} (GPT-4o) $\tau > 0.85$ --- AI-written Paper Reviews}}\\
\midrule
\textbf{Feature} & \textbf{Concept Description} \textit{(auto-interp: gemini-3.1-flash-lite)} & \textbf{Corr.} & \textbf{Logit $\Delta$}\\
\midrule
\multicolumn{4}{@{}l}{\textbf{AI Features (GPT-4o)}}\\
17\_13662 & single quotation marks surrounding titles or specific terms (\emph{e.g.,} \texttt{'}, \texttt{‘}, \texttt{’}) & 0.853 & $+0.525$ \\
17\_10356 & modal verb might appearing in various contexts (\emph{e.g.,} ``might'', ``might be'', ``might enhance'') & 0.866 & $+0.286$ \\
7\_8231   & words indicating evidence or confirmation such as supported and weaknesses & 0.851 & $+0.278$ \\
7\_15224  & words related to offer such as offered offers or names containing ofer & 0.857 & $+0.272$ \\
7\_13630  & word beyond in academic or formal inquiry contexts (\emph{e.g.,} ``beyond'', ``exceed'') & 0.866 & $+0.248$ \\
17\_1592  & words related to asserting or stating an academic or scientific position like claim or claims & 0.860 & $+0.225$ \\
7\_7055   & word across or synonyms for spatial or relational distribution (\emph{e.g.,} ``across'', ``spanning'', ``on'') & 0.886 & $+0.208$ \\
22\_11041 & the word across especially when followed by text or the (\emph{e.g.,} ``across'', ``across the'', ``across other'') & 0.866 & $+0.190$ \\
17\_12216 & potential and potentially used in academic writing when discussing future possibilities or limitations & 0.879 & $+0.187$ \\
22\_5088  & legal and policy terms regarding discrimination and fair classification (\emph{e.g.,} ``disparate impact'', ``equal protection laws'', ``disparate treatment'') & 0.855 & $+0.185$ \\
22\_2955  & societal ethical and human-related concepts in academic or AI policy critiques & 0.850 & $+0.184$ \\
13\_12598 & creativity and innovation related terminology (\emph{e.g.,} ``creativity'', ``innovative'', ``creatively'') & 0.861 & $+0.175$ \\
13\_885   & additionally or additional in academic paper reviews & 0.866 & $+0.156$ \\
7\_6871   & words related to professional status or conduct (\emph{e.g.,} ``professional'', ``unprofessional'', ``profession'') & 0.864 & $+0.139$ \\
7\_8621   & word complex or complexity often in academic or technical critiques (\emph{e.g.,} ``complexity'', ``complex'', ``complex concepts'') & 0.871 & $+0.134$ \\
17\_14380 & word impact or variations like impacts or impacting used in technical or research contexts & 0.866 & $+0.126$ \\
7\_16063  & names of people especially male given names like robert or bob & 0.866 & $+0.114$ \\
13\_12197 & modal verb could in academic reviews & 0.879 & $+0.112$ \\
\midrule
\multicolumn{4}{@{}l}{\textbf{Human Features}}\\
17\_9362  & commas in parenthetical or interrupting phrases (\emph{e.g.,} \texttt{,}, ``, well,'', ``, from'') & 0.857 & $-0.734$ \\
13\_5785  & first person singular pronoun i when used as a sentence subject or standalone pronoun & 0.860 & $-0.621$ \\
22\_3564  & numbers within academic citations and reference formatting (\emph{e.g.,} \texttt{13}, \texttt{300}, \texttt{1990}) & 0.851 & $-0.549$ \\
17\_15979 & academic phrasing identifying logical necessity or conditional statements like must be a (\emph{e.g.,} ``must be a'', ``and'', ``then'') & 0.898 & $-0.386$ \\
22\_14819 & mathematical subscripts variables and indices in equations (\emph{e.g.,} ``y\_i'', ``x\_s'', ``i+1'') & 0.871 & $-0.309$ \\
13\_4202  & conversational filler phrases like i mean and like how used to clarify or elaborate & 0.900 & $-0.289$ \\
17\_8975  & latex math mode delimiters and block math tags (\emph{e.g.,} \texttt{\$\$}, \texttt{\textbackslash{}n\textbackslash{}n}, \texttt{<{}<}) & 0.860 & $-0.277$ \\
17\_13674 & words related to acknowledgements or acknowledging in academic peer review contexts (\emph{e.g.,} ``acknowledgments'', ``acknowledge'', ``Acknowledgements:'') & 0.870 & $-0.275$ \\
17\_15087 & figures and captions in academic papers (\emph{e.g.,} ``Fig.'', ``Figure'', ``in'') & 0.859 & $-0.258$ \\
13\_14227 & first person references in phrases like as i mentioned in or as i stated in (\emph{e.g.,} ``as I mentioned in'', ``as I stated in'', ``as in'') & 0.887 & $-0.250$ \\
17\_10886 & temporal indicators like recently ago or adverbial time markers (\emph{e.g.,} ``Recently,'', ``ago'', ``recently, a'') & 0.853 & $-0.246$ \\
22\_12092 & polite requests starting sentences or parenthetical phrases (\emph{e.g.,} ``Please'', ``please'') & 0.866 & $-0.241$ \\
13\_1179  & meta-references or directional markers in academic review feedback like see note or see above & 0.860 & $-0.230$ \\
13\_5876  & formatting indicators like parentheses brackets colons and markdown markers used in technical writing or paper reviews (\emph{e.g.,} \texttt{(}, \texttt{:}, \texttt{]}) & 0.868 & $-0.201$ \\
17\_9491  & bibtex punctuation and whitespace patterns specifically commas and line breaks in bibliography entries (\emph{e.g.,} \texttt{\},}, \texttt{\textbackslash{}n    }, \texttt{\},\textbackslash{}n}) & 0.857 & $-0.199$ \\
22\_7318  & activates on the word at when used in prepositional phrases or idioms (\emph{e.g.,} ``at'', ``At'', ``at least'') & 0.851 & $-0.199$ \\
22\_5561  & medical or insurance related terms and parenthetical technical notations (\emph{e.g.,} ``ICD'', ``for'', ``Medicare \& Medicaid Services'') & 0.858 & $-0.196$ \\
13\_3795  & academic paper citation brackets containing reference numbers (\emph{e.g.,} \texttt{[7,21]}, \texttt{[12,24]}, \texttt{[43]}) & 0.879 & $-0.193$ \\
\bottomrule
\end{tabular}
\caption{Top features of the \textsc{Base-SAE} (GPT-4o) $\tau > 0.85$ logistic-regression detector on the PRDB dataset \cite{yu2026paperreviewedllmbenchmarking}, ranked by mean logit contribution (Logit $\Delta$) along with their auto-interpretation score (\S\ref{sec:autointerpretation}).}
\label{tab:top_features_prdb_basesae_gpt4o}
\end{table*}

\begin{table*}[t]
\centering
\small
\setlength{\tabcolsep}{4pt}
\renewcommand{\arraystretch}{1.2}
\begin{tabular}{@{}lp{12.2cm}rr@{}}
\toprule
\multicolumn{4}{@{}c}{\textbf{Most Predictive Features: \textsc{Base-SAE} (Llama) $\tau > 0.85$ --- AI-written Paper Reviews}}\\
\midrule
\textbf{Feature} & \textbf{Concept Description} \textit{(auto-interp: gemini-3.1-flash-lite)} & \textbf{Corr.} & \textbf{Logit $\Delta$}\\
\midrule
\multicolumn{4}{@{}l}{\textbf{AI Features (Llama 3.1 70B)}}\\
17\_8515  & punctuation and special characters in url paths or academic citations (\emph{e.g.,} \texttt{.}, ``/paper'', \texttt{?}, \texttt{=}, \texttt{\&}) & 0.885 & $+0.806$ \\
7\_6754   & word also when used as a transition in academic writing & 0.865 & $+0.428$ \\
7\_156    & word field occurring in academic or technical research contexts (\emph{e.g.,} ``field'', ``fields'', ``feld'') & 0.866 & $+0.296$ \\
13\_9022  & underscore character used as a word separator or within code variable names (\emph{e.g.,} \texttt{\_}, ``best\_split'', ``x\_0'') & 0.855 & $+0.253$ \\
7\_9197   & words related to academic analysis and evaluation like analysis analyzes and analytically & 0.871 & $+0.247$ \\
7\_5580   & quantifier word several or number (\emph{e.g.,} ``several'', ``a number of'') & 0.886 & $+0.226$ \\
17\_6178  & first person phrase i would preceding questions to authors (\emph{e.g.,} ``I would'', ``I would like'', ``I would like to'') & 0.886 & $+0.219$ \\
17\_12216 & potential and potentially used in academic writing when discussing future possibilities or limitations & 0.879 & $+0.205$ \\
17\_3739  & words relating to providing or provision such as provides providers and previous (\emph{e.g.,} ``provides'', ``providers'', ``previs'') & 0.871 & $+0.199$ \\
22\_11863 & forms of the verb provide or related verbs like available (\emph{e.g.,} ``provide'', ``providing'', ``available'') & 0.862 & $+0.195$ \\
13\_15486 & polite closing remarks in academic reviews (\emph{e.g.,} ``wish you luck'', ``best for future decisions'', ``Nice work'') & 0.897 & $+0.144$ \\
7\_11400  & academic phrases directing readers to further information (\emph{e.g.,} ``details on the'', ``more on this'', ``details about this'') & 0.860 & $+0.141$ \\
7\_10420  & transitions signaling the end of a document or section like conclusion finally and lastly & 0.853 & $+0.133$ \\
22\_1809  & words related to business entities like companies (\emph{e.g.,} ``companies'', ``Company'', ``company'') & 0.866 & $+0.132$ \\
22\_16303 & words describing high intensity or severity such as severe extreme exhaustive and intensive & 0.860 & $+0.130$ \\
7\_371    & introductory transition words signaling the start of a point or list (\emph{e.g.,} ``first'', ``firstly'', ``first of all'') & 0.885 & $+0.127$ \\
17\_3069  & comma separated parenthetical phrases like say or eg (\emph{e.g.,} ``, say'', ``, e.g.'', ``, for example'') & 0.859 & $+0.117$ \\
13\_885   & additionally or additional in academic paper reviews & 0.866 & $+0.115$ \\
\midrule
\multicolumn{4}{@{}l}{\textbf{Human Features}}\\
17\_9362  & commas in parenthetical or interrupting phrases (\emph{e.g.,} \texttt{,}, ``, well,'', ``, from'') & 0.857 & $-0.728$ \\
17\_8975  & latex math mode delimiters and block math tags (\emph{e.g.,} \texttt{\$\$}, \texttt{\textbackslash{}n\textbackslash{}n}, \texttt{<{}<}) & 0.860 & $-0.534$ \\
17\_15979 & academic phrasing identifying logical necessity or conditional statements like must be a (\emph{e.g.,} ``must be a'', ``and'', ``then'') & 0.898 & $-0.509$ \\
7\_10672  & words indicating hesitation or uncertainty like seems seemingly and appear & 0.853 & $-0.441$ \\
22\_12092 & polite requests starting sentences or parenthetical phrases (\emph{e.g.,} ``Please'', ``please'') & 0.866 & $-0.398$ \\
17\_97    & phrase seems to me that and related hedges (\emph{e.g.,} ``seems to'', ``seems simple'', ``appear to'') & 0.862 & $-0.356$ \\
22\_3564  & numbers within academic citations and reference formatting (\emph{e.g.,} \texttt{13}, \texttt{300}, \texttt{1990}) & 0.851 & $-0.354$ \\
17\_12942 & wikipedia urls and related links (\emph{e.g.,} ``wikipedia.org'', ``https://en.wikipedia'', ``wikipedia'') & 0.856 & $-0.347$ \\
22\_182   & the adverb very used as an intensifier & 0.866 & $-0.322$ \\
17\_10886 & temporal indicators like recently ago or adverbial time markers (\emph{e.g.,} ``Recently,'', ``ago'', ``recently, a'') & 0.853 & $-0.303$ \\
7\_15113  & vietnamese surnames such as nguyen (\emph{e.g.,} ``Nguyen'', ``Quang'', ``Minh'') & 0.856 & $-0.298$ \\
13\_5876  & formatting indicators like parentheses brackets colons and markdown markers used in technical writing or paper reviews (\emph{e.g.,} \texttt{(}, \texttt{:}, \texttt{]}) & 0.868 & $-0.295$ \\
13\_1179  & meta-references or directional markers in academic review feedback like see note or see above & 0.860 & $-0.271$ \\
22\_7318  & activates on the word at when used in prepositional phrases or idioms (\emph{e.g.,} ``at'', ``At'', ``at least'') & 0.851 & $-0.261$ \\
13\_15120 & self-deprecating expressions of confusion or personal limitation in academic or technical peer reviews (\emph{e.g.,} ``my part'', ``dumbo'', ``my personal bias'') & 0.888 & $-0.243$ \\
13\_4202  & conversational filler phrases like i mean and like how used to clarify or elaborate & 0.900 & $-0.202$ \\
13\_13586 & modal verb should followed by a verb or punctuation (\emph{e.g.,} ``should'', ``should stabilize'', ``should not'') & 0.861 & $-0.189$ \\
17\_13674 & words related to acknowledgements or acknowledging in academic peer review contexts (\emph{e.g.,} ``acknowledgments'', ``acknowledge'', ``Acknowledgements:'') & 0.870 & $-0.183$ \\
\bottomrule
\end{tabular}
\caption{Top features of the \textsc{Base-SAE} (Llama) $\tau > 0.85$ logistic-regression detector on the PRDB dataset \cite{yu2026paperreviewedllmbenchmarking}, ranked by mean logit contribution (Logit $\Delta$) along with their auto-interpretation score (\S\ref{sec:autointerpretation}).}
\label{tab:top_features_prdb_basesae_llama}
\end{table*}

\begin{table*}[t]
\centering
\small
\setlength{\tabcolsep}{4pt}
\renewcommand{\arraystretch}{1.2}
\begin{tabular}{@{}lp{12.2cm}rr@{}}
\toprule
\multicolumn{4}{@{}c}{\textbf{Most Predictive Features: \textsc{FT-Model+SAE} (GPT-4o) $\tau > 0.85$ --- AI-written Paper Reviews}}\\
\midrule
\textbf{Feature} & \textbf{Concept Description} \textit{(auto-interp: gemini-3.1-flash-lite)} & \textbf{Corr.} & \textbf{Logit $\Delta$}\\
\midrule
\multicolumn{4}{@{}l}{\textbf{AI Features (GPT-4o)}}\\
13\_15950 & academic paper review questions asking for more information details or clarification (\emph{e.g.,} ``provide additional'', ``provide insights'', ``provide a mathematical framework'') & 0.866 & $+1.002$ \\
17\_8459  & academic review feedback or peer review decisions (\emph{e.g.,} ``acceptance'', ``reject'', ``accepting'') & 0.870 & $+0.814$ \\
22\_4800  & double colons following bolded markdown section headers (\emph{e.g.,} \texttt{**Significance:}, \texttt{**Quality:}, \texttt{**Questions:}) & 0.860 & $+0.672$ \\
17\_12216 & activates on the word potential & 0.884 & $+0.515$ \\
7\_10900  & the word particularly when used as an adverb to specify or emphasize a context & 0.859 & $+0.438$ \\
17\_11188 & academic research terminology related to study design and methods (\emph{e.g.,} ``methodological'', ``methodology'', ``methodologies'') & 0.859 & $+0.388$ \\
7\_4872   & activates on the word potentially often used to hedge academic critique or discuss future implications & 0.860 & $+0.363$ \\
13\_15079 & societal impact in research paper sections & 0.892 & $+0.357$ \\
22\_16295 & the word beyond in academic or technical contexts & 0.859 & $+0.320$ \\
7\_5712   & forms of the word enhance such as enhances enhancing and enhancements & 0.866 & $+0.307$ \\
7\_10354  & word potential & 0.869 & $+0.306$ \\
7\_6591   & descriptive terms for academic writing difficulty like dense and terse (\emph{e.g.,} ``dense'', ``terse'', ``ed'') & 0.876 & $+0.302$ \\
17\_7088  & word elaborate appearing in academic review questions (\emph{e.g.,} ``elaborate'', ``elaborate on'') & 0.878 & $+0.269$ \\
22\_8361  & word insight or insights in academic or research contexts & 0.918 & $+0.237$ \\
17\_5639  & forms of the verb offer including offering and offers in academic paper review contexts (\emph{e.g.,} ``offering'', ``offers'', ``provides'') & 0.866 & $+0.237$ \\
7\_11041  & advancement or advancements in research papers (\emph{e.g.,} ``advancement'', ``advancements'', ``manuscript'') & 0.870 & $+0.228$ \\
7\_10956  & academic peer review feedback regarding paper acceptance or rejection thresholds and status (\emph{e.g.,} ``acceptance threshold'', ``acceptance'', ``threshold'') & 0.873 & $+0.208$ \\
7\_7170   & academic references to existing literature in research papers & 0.855 & $+0.161$ \\
\midrule
\multicolumn{4}{@{}l}{\textbf{Human Features}}\\
13\_5785  & first person singular pronoun i at start of sentences or after punctuation & 0.866 & $-0.919$ \\
17\_13291 & forms of the verb seem (\emph{e.g.,} ``seems'', ``seem'', ``seemed'') & 0.859 & $-0.639$ \\
22\_8699  & academic review section headers like clarity and clarify (\emph{e.g.,} ``Clarity'', ``clarify'', ``more'') & 0.851 & $-0.574$ \\
7\_6962   & word for at the start of a sentence & 0.859 & $-0.427$ \\
7\_13654  & math and variable notation indices and subscripts (\emph{e.g.,} ``i'', ``j'', \texttt{0}) & 0.857 & $-0.392$ \\
7\_7454   & the adverb very in various forms and cases & 0.853 & $-0.341$ \\
13\_5925  & modal verb might (\emph{e.g.,} ``might'', ``may'') & 0.866 & $-0.334$ \\
7\_23     & academic document figure or fig references (\emph{e.g.,} ``figure'', ``fig'', ``figures'') & 0.860 & $-0.306$ \\
22\_13623 & polite imperative requests used in academic review feedback (\emph{e.g.,} ``Please'', ``Thanks'', ``Strong'') & 0.854 & $-0.297$ \\
13\_12569 & words related to academic paper authors and roles like writer author or auth & 0.866 & $-0.294$ \\
7\_12294  & activates on the digit zero when appearing in numerical strings or exponents (\emph{e.g.,} \texttt{0}, \texttt{00}, \texttt{10}) & 0.894 & $-0.269$ \\
22\_7172  & word then in sequences of logical or procedural progression & 0.863 & $-0.236$ \\
22\_4547  & word only when used to denote uniqueness or exclusivity & 0.873 & $-0.199$ \\
13\_667   & polite request markers (\emph{e.g.,} ``kindly'', ``please'', ``ple'') & 0.852 & $-0.196$ \\
7\_9429   & polite imperative requests and directives (\emph{e.g.,} ``Kindly'', ``please'', ``Please'') & 0.890 & $-0.186$ \\
17\_109   & the subordinating conjunction since & 0.866 & $-0.184$ \\
13\_8964  & academic citation abbreviation et in et al (\emph{e.g.,} ``et'', ``Et'', ``etc'') & 0.873 & $-0.163$ \\
13\_2760  & words related to tables and tabular data (\emph{e.g.,} ``table'', ``tabular'', ``interpretable'') & 0.854 & $-0.159$ \\
\bottomrule
\end{tabular}
\caption{Top features of the \textsc{FT-Model+SAE} (GPT-4o) $\tau > 0.85$ logistic-regression detector on the PRDB dataset \cite{yu2026paperreviewedllmbenchmarking}, ranked by mean logit contribution (Logit $\Delta$) along with their auto-interpretation simulation score (\S\ref{sec:autointerpretation}).}
\label{tab:top_features_prdb_ftmodelsae_gpt4o}
\end{table*}

\begin{table*}[t]
\centering
\small
\setlength{\tabcolsep}{4pt}
\renewcommand{\arraystretch}{1.2}
\begin{tabular}{@{}lp{12.2cm}rr@{}}
\toprule
\multicolumn{4}{@{}c}{\textbf{Most Predictive Features: \textsc{FT-Model+SAE} (Llama) $\tau > 0.85$ --- AI-written Paper Reviews}}\\
\midrule
\textbf{Feature} & \textbf{Concept Description} \textit{(auto-interp: gemini-3.1-flash-lite)} & \textbf{Corr.} & \textbf{Logit $\Delta$}\\
\midrule
\multicolumn{4}{@{}l}{\textbf{AI Features (Llama 3.1 70B)}}\\
17\_4373  & the word thorough & 0.892 & $+0.993$ \\
13\_12061 & numbered list items in parentheses within question sections of research papers (\emph{e.g.,} \texttt{(1)}, \texttt{(2)}, \texttt{1.}) & 0.858 & $+0.898$ \\
7\_5460   & the word one when it appears at the start of a sentence or bullet point & 0.851 & $+0.745$ \\
13\_164   & opening square bracket in list or string literals (\emph{e.g.,} \texttt{['}, \texttt{[}, \texttt{<{}<}) & 0.866 & $+0.610$ \\
7\_4720   & word written used to describe paper quality (\emph{e.g.,} ``written'', ``Well-written'') & 0.898 & $+0.592$ \\
22\_8578  & word firstly when used as a transition in academic paper reviews & 0.900 & $+0.553$ \\
13\_5375  & word firstly when used as a transition in academic or research paper critique sections (\emph{e.g.,} ``Firstly'', ``Secondly'') & 0.857 & $+0.519$ \\
7\_156    & academic field noun when used in singular or plural form (\emph{e.g.,} ``field'', ``fields'') & 0.852 & $+0.459$ \\
22\_7140  & activates on the word potential & 0.883 & $+0.446$ \\
7\_2635   & academic paper review phrases like has several strengths or has several weaknesses & 0.875 & $+0.446$ \\
13\_13570 & academic paper review strengths or methodology descriptions (\emph{e.g.,} ``Strengths'', ``framework'', ``approach'') & 0.861 & $+0.375$ \\
13\_4401  & word limitations within academic paper reviews & 0.884 & $+0.291$ \\
7\_5799   & word details and its variations including detailed & 0.871 & $+0.278$ \\
7\_11355  & forms of the word consider in research paper critiques (\emph{e.g.,} ``consider'', ``considerations'') & 0.866 & $+0.242$ \\
7\_7545   & word organized in academic paper reviews & 0.886 & $+0.223$ \\
13\_8059  & forms of the verb provide in academic paper reviews (\emph{e.g.,} ``provide'', ``provided'', ``provides'') & 0.870 & $+0.205$ \\
7\_1622   & word clear and its variations clearer & 0.866 & $+0.191$ \\
7\_547    & words related to showing or proving something such as demonstrate demonstrations and demonstrate (\emph{e.g.,} ``demonstrate'', ``demonstrations'', ``demonstrated'') & 0.867 & $+0.175$ \\
\midrule
\multicolumn{4}{@{}l}{\textbf{Human Features}}\\
7\_1616   & the word see used as a directive or reference to look elsewhere in academic reviews & 0.867 & $-0.836$ \\
7\_10672  & word seem or seems when used to express an opinion or observation about a paper or argument & 0.866 & $-0.808$ \\
22\_773   & closing parenthesis followed by a period or similar closing punctuation (\emph{e.g.,} \texttt{).}, \texttt{))}, \texttt{).}) & 0.853 & $-0.745$ \\
13\_667   & polite requests in formal feedback or review contexts (\emph{e.g.,} ``Please'', ``please'', ``Pls'') & 0.858 & $-0.706$ \\
22\_10620 & modal verb could in academic critique or feedback (\emph{e.g.,} ``could'', ``could be'', ``could have'') & 0.854 & $-0.705$ \\
22\_2854  & first person singular objective and reflexive pronouns (\emph{e.g.,} ``my'', ``myself'', ``me'') & 0.860 & $-0.629$ \\
7\_10682  & the digit 3 when used in a numbered list or section reference (\emph{e.g.,} \texttt{3.}, \texttt{(3)}, \texttt{3)}) & 0.853 & $-0.569$ \\
7\_1446   & word since appearing at the beginning of a sentence or clause & 0.856 & $-0.487$ \\
17\_16165 & modal verb will appearing in academic or technical discussion (\emph{e.g.,} ``will'', ``would'', ``a'') & 0.893 & $-0.379$ \\
13\_899   & past tense auxiliary verbs like were had and was & 0.869 & $-0.368$ \\
7\_9714   & first person singular verb am & 0.871 & $-0.318$ \\
13\_129   & forms of the verb give (\emph{e.g.,} ``given'', ``gives'', ``send'') & 0.866 & $-0.298$ \\
7\_550    & activates on the word works & 0.866 & $-0.256$ \\
7\_12294  & activates on the digit zero (\emph{e.g.,} \texttt{0}) & 0.894 & $-0.242$ \\
22\_14497 & modal verb should used in academic or formal critique & 0.867 & $-0.230$ \\
17\_8092  & the word should when used in academic or technical critiques or recommendations & 0.866 & $-0.227$ \\
7\_13919  & horizontal tab characters used for formatting or indentation in structured documents (\emph{e.g.,} \texttt{\textbackslash{}t}) & 0.851 & $-0.190$ \\
7\_1135   & numbers spelled as words like five and digits like 5 in academic papers & 0.864 & $-0.184$ \\
\bottomrule
\end{tabular}
\caption{Top features of the \textsc{FT-Model+SAE} (Llama) $\tau > 0.85$ logistic-regression detector on the PRDB dataset \cite{yu2026paperreviewedllmbenchmarking}, ranked by mean logit contribution (Logit $\Delta$) along with their auto-interpretation score (\S\ref{sec:autointerpretation}).}
\label{tab:top_features_prdb_ftmodelsae_llama}
\end{table*}

\begin{table*}[t]
\centering
\small
\setlength{\tabcolsep}{4pt}
\renewcommand{\arraystretch}{1.2}
\begin{tabular}{@{}lp{12.2cm}rr@{}}
\toprule
\multicolumn{4}{@{}c}{\textbf{Most Predictive Features: \textsc{FT-SAE} (GPT-4o) $\tau > 0.85$ --- AI-written Paper Reviews}}\\
\midrule
\textbf{Feature} & \textbf{Concept Description} \textit{(auto-interp: gemini-3.1-flash-lite)} & \textbf{Corr.} & \textbf{Logit $\Delta$}\\
\midrule
\multicolumn{4}{@{}l}{\textbf{AI Features (GPT-4o)}}\\
13\_1632  & modal verbs beginning questions in academic paper reviews (\emph{e.g.,} ``Can'', ``Could'', ``can'') & 0.871 & $+0.370$ \\
17\_7486  & word environments appearing in technical literature regarding system deployment or model applicability (\emph{e.g.,} ``environments'', ``domains'', ``architectures'') & 0.869 & $+0.345$ \\
22\_14556 & methodology and methodologies in academic research contexts (\emph{e.g.,} ``methodology'', ``methodologies'', ``methodological'') & 0.903 & $+0.258$ \\
13\_3382  & activates on the word clarity when used in academic or technical peer review critiques & 0.859 & $+0.240$ \\
22\_5116  & innovative variations like innovative innovations or invasive & 0.862 & $+0.237$ \\
13\_9782  & academic paper verbs introducing research contributions (\emph{e.g.,} ``presents'', ``aims to address'', ``investigates'') & 0.873 & $+0.184$ \\
13\_11446 & academic paper titles within quotes (\emph{e.g.,} ``paper'', ``entitled'', ``summary'') & 0.855 & $+0.171$ \\
7\_5773   & word particular and variations like particularly in academic writing context & 0.861 & $+0.154$ \\
7\_7818   & academic phrases predicting future influence or impact on researchers (\emph{e.g.,} ``researchers'', ``influence'', ``pave a path for'') & 0.871 & $+0.141$ \\
17\_2301  & words related to requests for improved clarity or explanation in academic reviews (\emph{e.g.,} ``explained'', ``intuitive explanations'', ``elucidation'', ``aids'', ``clearer'') & 0.871 & $+0.140$ \\
17\_9805  & text concerning accessibility or difficulty of academic papers for specific readerships (\emph{e.g.,} ``for readers'', ``challenging for'', ``limiting accessibility'') & 0.887 & $+0.135$ \\
17\_4789  & word beyond in academic contexts discussing research scope or limitations & 0.851 & $+0.135$ \\
17\_3502  & word across in academic papers (\emph{e.g.,} ``across'', ``across the'', ``across multiple'') & 0.866 & $+0.131$ \\
7\_11596  & broadly breadth broader variations of the word broad & 0.866 & $+0.131$ \\
7\_6123   & potential and related variations (\emph{e.g.,} ``potential'', ``potentially'', ``potentials'') & 0.869 & $+0.126$ \\
22\_3522  & academic peer review requests for additional information (\emph{e.g.,} ``provide more details'', ``provide more'') & 0.876 & $+0.122$ \\
17\_10051 & multi-word academic terminology specifically within titles or descriptive naming conventions (\emph{e.g.,} ``Self-Play'', ``Multi-Modal'', ``Multi-level Multimodal'') & 0.906 & $+0.120$ \\
22\_5057  & academic writing style describing paper accessibility or technical complexity (\emph{e.g.,} ``technical'', ``formalism'', ``proofs'') & 0.873 & $+0.117$ \\
\midrule
\multicolumn{4}{@{}l}{\textbf{Human Features}}\\
13\_10014 & figures and tables in technical documents (\emph{e.g.,} ``figure'', ``Figure.'', ``Table.'') & 0.867 & $-0.358$ \\
13\_782   & words or phrases directing the reader to another section of the text or to previously mentioned content (\emph{e.g.,} ``See'', ``refer to'', ``find'') & 0.875 & $-0.340$ \\
22\_7509  & readability of academic papers specifically phrases describing ease or difficulty of comprehension (\emph{e.g.,} ``easy to'', ``hard to'', ``to'') & 0.860 & $-0.338$ \\
13\_13225 & mathematical notation and variables in technical papers (\emph{e.g.,} ``G'', ``X\_t'', ``Theta'', ``expectation'', ``r’'') & 0.866 & $-0.281$ \\
22\_15079 & academic citation abbreviations like et al or etc & 0.856 & $-0.280$ \\
17\_12604 & refers to external text sections or citations using phrases like to the or see (\emph{e.g.,} ``to the'', ``to the QUESTIONS'', ``see'') & 0.868 & $-0.271$ \\
7\_11173  & initials or abbreviations like N.A. or N.A. in academic paper headings (\emph{e.g.,} ``N.A.'', ``NA.'', ``None.'') & 0.851 & $-0.269$ \\
17\_394   & first-person opinion markers like i believe or i would say (\emph{e.g.,} ``it'', ``i believe'', ``i would say'') & 0.858 & $-0.263$ \\
17\_16010 & words suggesting constructive feedback or desired improvements in academic or research review contexts (\emph{e.g.,} ``useful'', ``nice if'', ``good'') & 0.852 & $-0.229$ \\
13\_8790  & first person singular pronouns referring to the author or reviewer (\emph{e.g.,} ``me'', ``I'', ``my'') & 0.866 & $-0.210$ \\
17\_8327  & academic citation formatting and metadata specifically author names and punctuation in citations (\emph{e.g.,} ``zende'', ``feng'', ``ato'') & 0.859 & $-0.187$ \\
17\_5078  & subjective opinion phrases such as believe that think that and might that & 0.865 & $-0.179$ \\
13\_1362  & word only used as a restrictive conditional or exclusion modifier (\emph{e.g.,} ``only'', ``only if'') & 0.854 & $-0.172$ \\
13\_13629 & academic peer review feedback focusing on mathematical notation equation parts and specific paper reference structure (\emph{e.g.,} ``equation'', ``parentheses'', ``computed'') & 0.869 & $-0.165$ \\
17\_8644  & words related to expectations or anticipating outcomes or results (\emph{e.g.,} ``expect'', ``expecting'', ``expect the'') & 0.853 & $-0.163$ \\
\bottomrule
\end{tabular}
\caption{Top features of the \textsc{FT-SAE} (GPT-4o) $\tau > 0.85$ logistic-regression detector on the PRDB dataset \cite{yu2026paperreviewedllmbenchmarking}, ranked by mean logit contribution (Logit $\Delta$) along with their auto-interpretation score (\S\ref{sec:autointerpretation}).}
\label{tab:top_features_prdb_ftsae_gpt4o}
\end{table*}

\begin{table*}[t]
\centering
\small
\setlength{\tabcolsep}{4pt}
\renewcommand{\arraystretch}{1.2}
\begin{tabular}{@{}lp{12.2cm}rr@{}}
\toprule
\multicolumn{4}{@{}c}{\textbf{Most Predictive Features: \textsc{FT-SAE} (Llama) $\tau > 0.85$ --- AI-written Peer Reviews}}\\
\midrule
\textbf{Feature} & \textbf{Concept Description} \textit{(auto-interp: gemini-3.1-flash-lite)} & \textbf{Corr.} & \textbf{Logit $\Delta$}\\
\midrule
\multicolumn{4}{@{}l}{\textbf{AI Features (Llama 3.1 70B)}}\\
17\_5027  & academic review phrases like provide a clear or provide a nuanced (\emph{e.g.,} ``provide a clear'', ``provide a concise'', ``provide a nuanced'') & 0.893 & $+0.583$ \\
13\_13076 & introductory phrases for questions in academic paper reviews (\emph{e.g.,} ``include:'', ``potential questions for'', ``is'') & 0.880 & $+0.497$ \\
7\_13926  & thorough analysis of phrase (\emph{e.g.,} ``thorough analysis of'', ``analysis of'', ``evaluation of'') & 0.886 & $+0.464$ \\
13\_3534  & initial transition words like firstly and introductory adverbial clauses (\emph{e.g.,} ``Firstly'', ``First'', ``The'') & 0.875 & $+0.452$ \\
13\_16216 & words related to thorough or thoroughness (\emph{e.g.,} ``thorough'', ``thoroughly'', ``through'') & 0.867 & $+0.361$ \\
17\_12556 & negation phrases involving may or not (\emph{e.g.,} ``which may not'', ``may not be'', ``may not always'') & 0.857 & $+0.287$ \\
7\_1881   & words related to the word thorough and its variations (\emph{e.g.,} ``thorough'', ``thoroughly'', ``thorough and'') & 0.858 & $+0.279$ \\
22\_1479  & compares performance or methodology to other things (\emph{e.g.,} ``compare to'', ``compare to other'') & 0.870 & $+0.237$ \\
13\_2903  & the phrase be the in academic critique & 0.853 & $+0.228$ \\
13\_7866  & academic review phrases evaluating research papers (\emph{e.g.,} ``this is a solid'', ``the paper contributes'', ``the paper presents a significant contribution to'') & 0.878 & $+0.213$ \\
7\_13578  & modal verbs of negation like may not or might not (\emph{e.g.,} ``may not'', ``might not'', ``may not be'') & 0.881 & $+0.210$ \\
13\_12882 & academic review language focusing on paper contributions (\emph{e.g.,} ``contribution to the field'', ``approach to a'', ``novel'') & 0.860 & $+0.192$ \\
22\_12836 & academic review phrases like paper is well written (\emph{e.g.,} ``paper is well'', ``well-written'', ``the paper'') & 0.865 & $+0.179$ \\
7\_3884   & transitions used to introduce new points or arguments like additionally and furthermore & 0.850 & $+0.176$ \\
13\_16340 & the copula is in the context of academic paper weakness sections (\emph{e.g.,} ``is'', ``is that'', ``is the'') & 0.853 & $+0.160$ \\
13\_4619  & academic peer review feedback praising or criticizing the clarity and explanation of research papers (\emph{e.g.,} ``clear and concise'', ``explanation'', ``effectiveness'') & 0.852 & $+0.155$ \\
22\_6633  & the word of in the phrase state-of-the-art & 0.871 & $+0.152$ \\
\midrule
\multicolumn{4}{@{}l}{\textbf{Human Features}}\\
7\_14726  & numbered list items in academic paper reviews (\emph{e.g.,} \texttt{1.}, \texttt{1)}, \texttt{2.}) & 0.893 & $-0.457$ \\
7\_10672  & the verb seem or appears when used in academic or technical critique & 0.871 & $-0.383$ \\
7\_8431   & contractions and auxiliary verb phrases like ve, s, and t (\emph{e.g.,} ``'ve'', ``'s'', ``'t'') & 0.857 & $-0.329$ \\
13\_10014 & figures and tables in technical documents (\emph{e.g.,} ``figure'', ``Figure.'', ``Table.'') & 0.867 & $-0.327$ \\
17\_394   & first-person opinion markers like i believe or i would say (\emph{e.g.,} ``it'', ``i believe'', ``i would say'') & 0.858 & $-0.322$ \\
13\_782   & words or phrases directing the reader to another section of the text or to previously mentioned content (\emph{e.g.,} ``See'', ``refer to'', ``find'') & 0.875 & $-0.305$ \\
17\_15040 & forms of the word seem including seem seemingly seems & 0.866 & $-0.294$ \\
22\_15079 & academic citation abbreviations like et al or etc & 0.856 & $-0.285$ \\
13\_10878 & interrogative phrases and questions in technical reviews or paper discussions (\emph{e.g.,} ``What'', ``Why'', ``what does'') & 0.859 & $-0.282$ \\
7\_12258  & the phrase summary of the review typically within bold markdown (\emph{e.g.,} ``Summary of the Review'', \texttt{**Summary of the Review**}, ``Summary of'') & 0.864 & $-0.280$ \\
13\_13225 & mathematical notation and variables in technical papers (\emph{e.g.,} ``G'', ``X\_t'', ``Theta'', ``expectation'', ``r’'') & 0.866 & $-0.261$ \\
7\_11173  & initials or abbreviations like N.A. or N.A. in academic paper headings (\emph{e.g.,} ``N.A.'', ``NA.'', ``None.'') & 0.851 & $-0.256$ \\
17\_12604 & refers to external text sections or citations using phrases like to the or see (\emph{e.g.,} ``to the'', ``to the QUESTIONS'', ``see'') & 0.868 & $-0.222$ \\
17\_1388  & first person subjective expressions or comparisons of ideas to previous work (\emph{e.g.,} ``bears a resemblance'', ``me'', ``reminds me'') & 0.851 & $-0.220$ \\
13\_1362  & word only used as a restrictive conditional or exclusion modifier (\emph{e.g.,} ``only'', ``only if'') & 0.854 & $-0.205$ \\
17\_8644  & words related to expectations or anticipating outcomes or results (\emph{e.g.,} ``expect'', ``expecting'', ``expect the'') & 0.853 & $-0.197$ \\
\bottomrule
\end{tabular}
\caption{Top features of the \textsc{FT-SAE} (Llama) $\tau > 0.85$ logistic-regression detector on the PRDB dataset \cite{yu2026paperreviewedllmbenchmarking}, ranked by mean logit contribution (Logit $\Delta$) along with their auto-interpretation simulation score (\S\ref{sec:autointerpretation}).}
\label{tab:top_features_prdb_ftsae_llama}
\end{table*}

\end{document}